\documentclass[runningheads]{llncs}

\usepackage[mobile]{eccv}

\usepackage{eccvabbrv}

\usepackage{graphicx}
\usepackage{booktabs}
\usepackage{bm}
\usepackage{cite}

\usepackage[dvipsnames]{xcolor}

\usepackage{xcolor,colortbl}
\usepackage{arydshln} 
\usepackage{array} % Required for custom column specification

\definecolor{Gray}{gray}{0.3}
\definecolor{LightCyan}{rgb}{0.88,1,1}

\newcolumntype{a}{:c|}
\newcolumntype{d}{:c}

\definecolor{tabfirst}{rgb}{0.4, 0.65, 0.3} % dark green
\definecolor{tabsecond}{rgb}{0.7, 0.8, 0.65} % light green

\usepackage{multirow}
\usepackage{hyperref}

\DeclareMathOperator*{\argmin}{arg\,min}
\usepackage{orcidlink}

\begin{document}

\title{Global Structure-from-Motion Revisited} 


\author{Linfei Pan$^1$\and
Dániel Baráth$^1$ \and
Marc Pollefeys$^{1,2}$ \and
Johannes L.~Sch\"{o}nberger$^2$
\\
$^1$ ETH Zurich \quad\quad $^2$ Microsoft
}

\authorrunning{L. Pan et al.}

\institute{}
\maketitle

% \vspace{-5px}
\begin{abstract}
    Recovering 3D structure and camera motion from images has been a long-standing focus of computer vision research and is known as Structure-from-Motion (SfM). Solutions to this problem are categorized into incremental and global approaches. Until now, the most popular systems follow the incremental paradigm due to its superior accuracy and robustness, while global approaches are drastically more scalable and efficient. With this work, we revisit the problem of global SfM and propose GLOMAP as a new general-purpose system that outperforms the state of the art in global SfM. In terms of accuracy and robustness, we achieve results on-par or superior to COLMAP, the most widely used incremental SfM, while being orders of magnitude faster. We share our system as an open-source implementation at \url{https://github.com/colmap/glomap}.
  % \keywords{Structure-from-Motion \and Global SfM \and Translation Averaging \and Camera Motion Estimation}
\end{abstract}

\section{Introduction}

\begin{figure}[t]
    \centering
    % \resizebox{\textwidth}{!}{
    % \begin{tabular}{c c 
    % } 
    % \includegraphics[height=0.25\textwidth]{figure/teaser_reconstruction5.png}
    % & \includegraphics[height=0.25\textwidth]{figure/LIN_map.png}\\
    % (a) Example reconstructions
    % & (b) LaMAR~\cite{sarlin2022lamar} \textit{LIN} reconstructions 
    % \end{tabular}
    % }
    
    \subfloat[Example reconstructions]{%
    \includegraphics[height=0.27\textwidth]{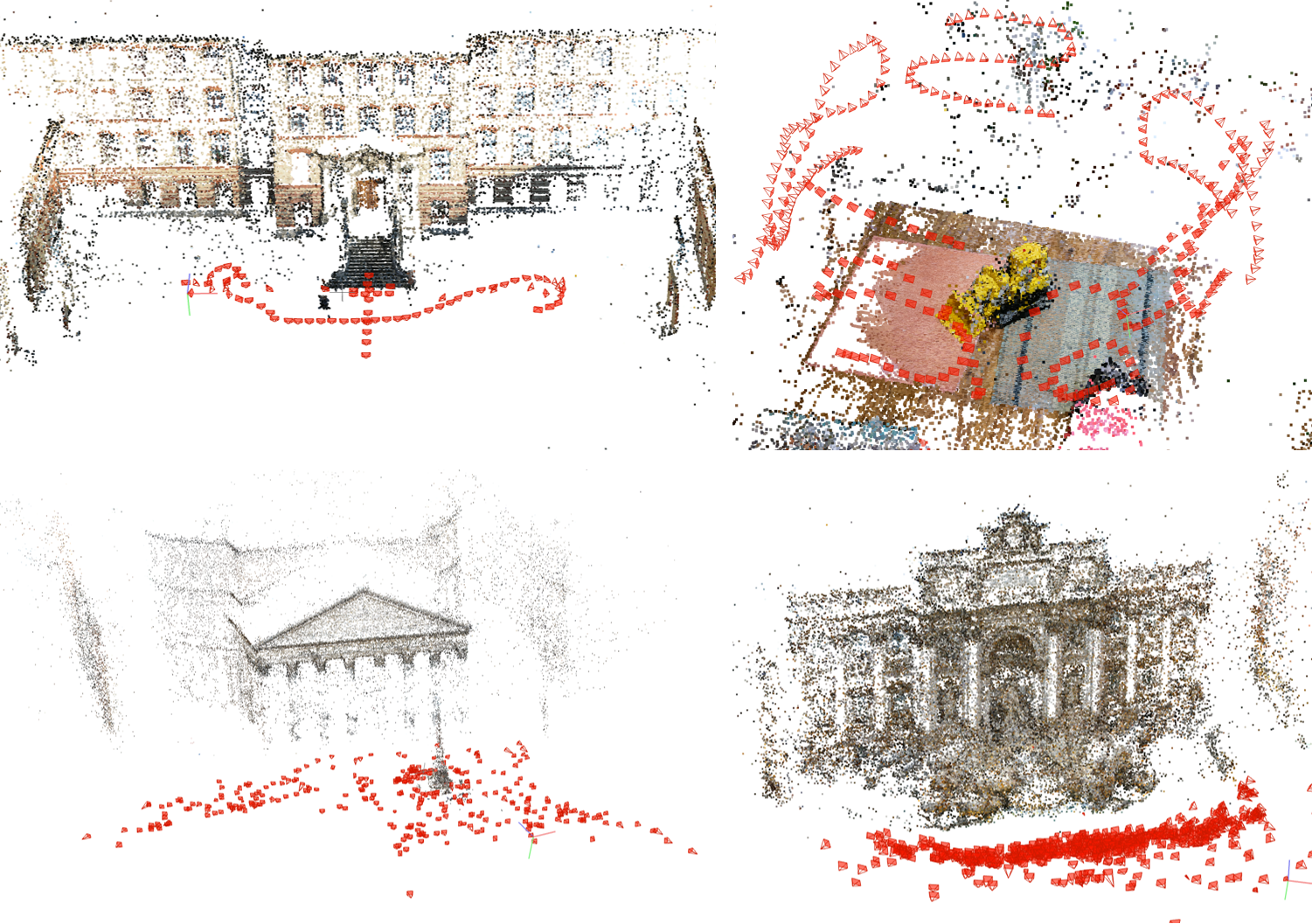}
    \label{fig:teaser_reconstructions}
    }
    \subfloat[LaMAR~\cite{sarlin2022lamar} \textit{LIN} reconstructions] {
    \includegraphics[height=0.27\textwidth]{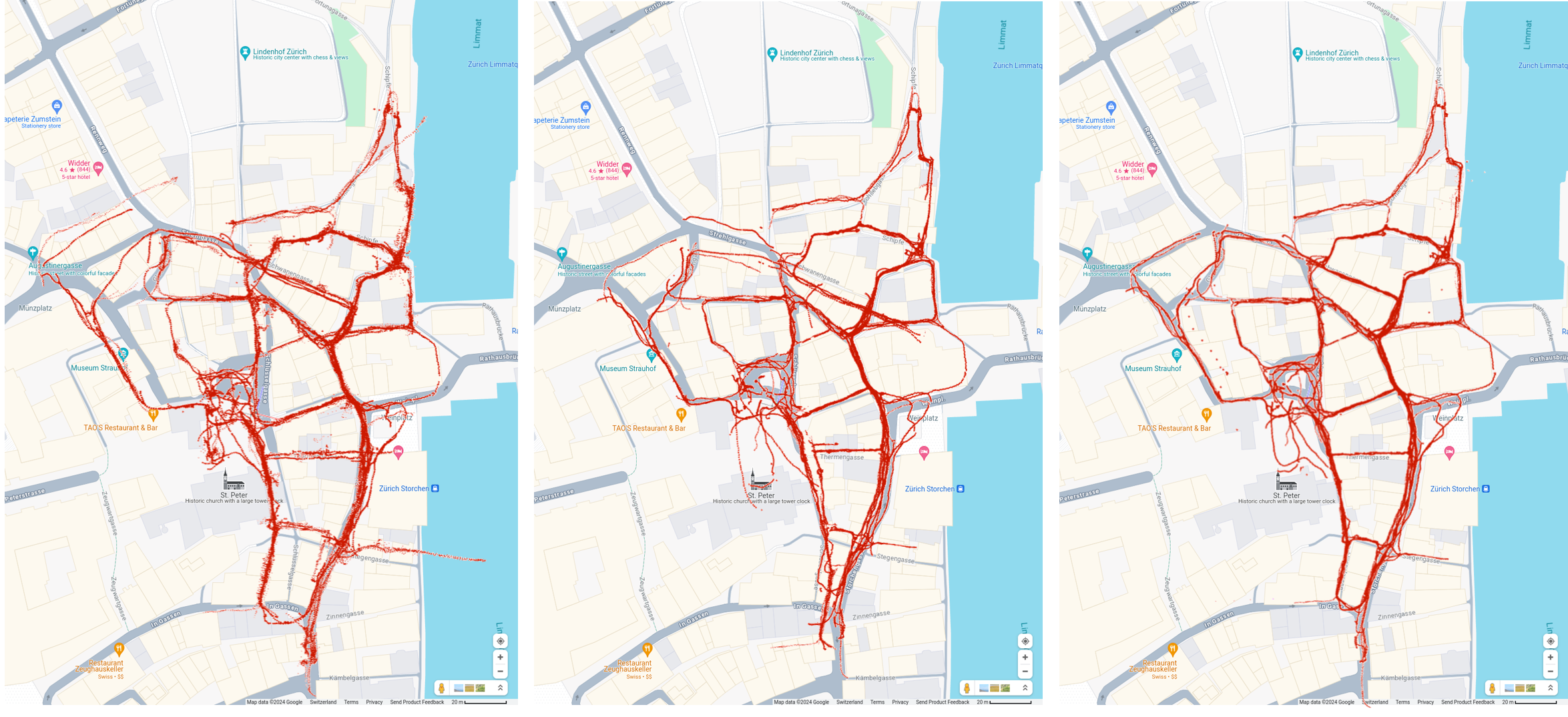}
    \label{fig:teaser_lamar}
    }
    \caption{
    Proposed GLOMAP produces satisfying reconstructions on various datasets. For (b), from left to right are estimated by Theia~\cite{theia-manual}, COLMAP~\cite{schoenberger2016sfm}, GLOMAP. While baseline models fail to produce reliable estimations, GLOMAP achieves high accuracy.
    % deviates heavily from real world, GLOMAP successful estimate accurate camera poses. 
    }
    % \vspace{-5px}
\label{fig:reconstructions}
\end{figure}

Recovering 3D structure and camera motion from a collection of images remains a fundamental problem in computer vision that is highly relevant for a variety of downstream tasks, such as novel-view-synthesis~\cite{mildenhall2021nerf,kerbl2023gaussian} or cloud-based mapping and localization~\cite{kipman2019azure,google2019google}.
The literature commonly refers to this problem as Structure-from-Motion (SfM)~\cite{ullman1979interpretation} and, over the years, two main paradigms for solving it have emerged: \textit{incremental} and \textit{global} approaches.
Both of them start with image-based feature extraction and matching followed by two-view geometry estimation to construct the initial view graph of the input images.
Incremental methods then seed the reconstruction from two views and sequentially expand it by registering additional camera images and associated 3D structure. This sequential process interleaves absolute camera pose estimation, triangulation, and bundle adjustment, which, despite achieving high accuracy and robustness, limits scalability due to the costly repeated bundle adjustments.
In contrast, global methods recover the camera geometry for all input images at once in separate rotation and translation averaging steps by jointly considering all two-view geometries in the view graph.
Typically, the globally estimated camera geometry is then used as an initialization for triangulation of the 3D structure before a final global bundle adjustment step.
While state-of-the-art incremental approaches are considered more accurate and robust, the reconstruction process of global approaches is more scalable and, in practice, orders of magnitude faster.
% Although these methods are generally considered less accurate than incremental ones, they have the potential for lightweight, efficient, and scalable reconstruction.
In this paper, we revisit the problem of global SfM and propose a comprehensive system achieving a similar level of accuracy and robustness as state-of-the-art incremental SfM (\eg Fig.~\ref{fig:teaser_reconstructions}) while maintaining the efficiency and scalability of global approaches.

% \begin{figure}[t]
%     \centering
%     \includegraphics[height=0.18\textwidth]{figure/reconstruction_bonsai.png}
%     % \includegraphics[height=0.185\textwidth]{figure/reconstruction_brandenburg.png}
%     \includegraphics[height=0.18\textwidth]{figure/reconstruction_pantheon.png}
%     \includegraphics[height=0.18\textwidth]{figure/reconstruction_statue.png}
%     \includegraphics[height=0.18\textwidth]{figure/reconstruction_courtyard.png}
%     \includegraphics[height=0.18\textwidth]{figure/reconstruction_relief2.png}
%     \includegraphics[height=0.18\textwidth]{figure/reconstruction_terrain.png}
%     \includegraphics[height=0.18\textwidth]{figure/reconstruction_trevi_fountain.png}
%     \caption{
%     Example reconstructions from the proposed GLOMAP on various datasets.
%     }
%     \label{fig:reconstructions}
%     \vspace{-20px}
% \end{figure}

The main reason for the accuracy and robustness gap between incremental and global SfM lies in the global translation averaging step.
Translation averaging describes the problem of estimating global camera positions from the set of relative poses in the view graph with the camera orientations recovered before by rotation averaging.
This process faces three major challenges in practice. 
The first being scale ambiguity: relative translation from estimated two-view geometry can only be determined up to scale~\cite{hartley2003multiple}. 
As such, to accurately estimate global camera positions, triplets of relative directions are required. 
However, when these triplets form skewed triangles, the estimated scales are especially prone to noise in the observations~\cite{manam2024sensitivity}. 
Second, accurately decomposing relative two-view geometry into rotation and translation components requires prior knowledge of accurate camera intrinsics.
Without this information, the estimated translation direction is often subject to large errors.
The third challenge arises for nearly co-linear motion that leads to a degenerate reconstruction problem.
Such motion patterns are common, especially in sequential datasets. 
These issues collectively contribute to the instability of camera position estimation, severely affecting the overall accuracy and robustness of existing global SfM systems.
% With the ill-posed problem formulation and the noisy input, the translation averaging is sensitive and not robust.
Motivated by the difficulties in translation averaging, significant research efforts have been dedicated to this problem.
Many of the recent approaches~\cite{cui2015global,cui2017hsfm,arie2012global,wilson2014robust, cui2015linear,holynski2020reducing, cai2021pose} share a common characteristic with incremental SfM as they incorporate image points into the problem formulation. 
Building on this insight, we propose a global SfM system that directly combines the estimation of camera positions and 3D structure in a single global positioning step.

The main contribution of this work is the introduction of a general-purpose global SfM system, termed GLOMAP.
The core difference to previous global SfM systems lies in the step of global positioning.
Instead of first performing ill-posed translation averaging followed by global triangulation, our proposed method performs joint camera and point position estimation.
GLOMAP achieves a similar level of robustness and accuracy as state-of-the-art incremental SfM systems~\cite{schoenberger2016sfm} while maintaining the efficiency of global SfM pipelines.
Unlike most previous global SfM systems, ours can deal with unknown camera intrinsics (\eg, as found in internet photos) and robustly handles sequential image data (\eg, handheld videos or self-driving car scenarios). We share our system as an open-source implementation at \url{https://github.com/colmap/glomap}.

\section{Review of Global Structure-from-Motion}

Global SfM pipelines generally consist of three main steps: correspondence search, camera pose estimation, and joint camera and structure refinement. The next sections provide a detailed review of state-of-the-art algorithms and frameworks.

\subsection{Correspondence Search}

Both incremental and global SfM begin with salient image feature extraction from the input images $\mathcal{I} = \{I_1, \cdots, I_N\}$.
Traditionally, feature points~\cite{lowe2004distinctive,detone2018superpoint} are detected and then described with compact signatures derived from the local context around the detection.
This is followed by the search for feature correspondences between pairs of images $(I_i, I_j)$, which starts by efficiently identifying subsets of images~\cite{arandjelovic2016netvlad} with overlapping fields of view and subsequently matching them in a more costly procedure~\cite{lowe2004distinctive,sarlin2020superglue}.
The matching is usually first done purely based on compact visual signatures producing a relatively large fraction of outliers initially.
These are then verified by robustly~\cite{barath2020magsac++} recovering the two-view geometry for overlapping pairs.
Based on the geometric configuration of the cameras, this yields either a homography $\bm{H}_{ij}$ for planar scenes with general motion and pure camera rotation with general scenes, or a fundamental matrix $\bm{F}_{ij}$ (uncalibrated) and essential matrix $\bm{E}_{ij}$ (calibrated) for general scenes and general motion.
When the camera intrinsics are approximately known, these can be decomposed~\cite{hartley2003multiple} into relative rotation $\bm{R}_{ij} \in \text{SO}(3)$ and translation $\bm{t}_{ij} \in \mathbb{R}^3$.

The computed two-view geometries with associated inlier correspondences define the view graph $\mathcal{G}$ that serves as the input to the global reconstruction steps.
In our pipeline, we rely on COLMAP's~\cite{schoenberger2016sfm} correspondence search implementation~\cite{schoenberger2018thesis} with RootSIFT features and scalable bag-of-words image retrieval~\cite{schoenberger2016vote} to find candidate overlapping pairs for brute-force feature matching.

\subsection{Global Camera Pose Estimation}

Global camera pose estimation is the key step distinguishing global from incremental SfM.
Instead of sequentially registering cameras with repeated triangulation and bundle adjustment, global SfM seeks to estimate all the camera poses $\bm{P}_i = (\bm{R}_i, \bm{c}_i) \in \text{SE}(3)$ at once using the view graph $\mathcal{G}$ as input.
To make the problem tractable, it is typically decomposed into separate rotation and translation averaging steps~\cite{theia-manual,moulon2016openmvg} with some works also refining the view graph before~\cite{sweeney2015optimizing}, or directly estimating camera poses from the view-graph of two-view geometries \cite{kasten2019algebraic, kasten2019gpsfm}.
% with some works also attempting to solve it with similarity averaging~\cite{cui2015global}.
The main challenge lies in dealing with noise and outliers in the view graph by careful modeling and solving of the optimization problems.
% \todo{CHECK THIS CITATION! A BIT STRANGE}.
% Global w
% To estimate the camera pose, rotations and translation are generally estimated separately, b

% As rotation 
% \vspace{0mm}
\noindent\textbf{Rotation Averaging}, sometimes also referred to as rotation synchronization, has been studied for several decades~\cite{hartley2013rotation, martinec2007robust} and is related to pose graph optimization (PGO) algorithms~\cite{carlone2012linear, carlone2018convex}.
It is typically formulated as a non-linear optimization, penalizing the deviation of the global rotation from estimated relative poses.
Specifically, absolute rotations $\bm{R}_i$ and relative rotations $\bm{R}_{ij}$ should ideally satisfy the constraint $\bm{R}_{ij} = \bm{R}_j\bm{R}_i^\top$.
%
% \begin{equation}
%     \bm{R}_{ij} = \bm{R}_j\bm{R}_i^\top.
%     \label{eq:r_ij}
% \end{equation}
%
However, in practice, this does not hold exactly due to noise and outliers.
Thus, the problem is generally modeled with a robust least-metric objective and optimized as:
\begin{equation}
    \argmin_{\bm{R}} \sum_{i, j} \rho\left(d(\bm{R}_j^\top \bm{R}_{ij} \bm{R}_i, \bm{I})^p\right) .
    \label{eq:R_err} 
\end{equation}
Hartley~\etal~\cite{hartley2013rotation} provide a comprehensive overview of various choices of robustifiers $\rho$ (\eg, Huber), rotation parameterizations (\eg, quaternion or axis-angle), and distance metrics $d$ (\eg, chordal distance or geodesic distance).
% Chordal distance is defined to be the Frobenius norm of the two rotation matrix difference $\|\bm{R}- \bm{R}'\|_F$, while geodesic distance represents the angle difference between two rotations, which can be calculated as $\arccos\left(\frac{tr( {\bm{R}} \bm{R'}^\top) - 1}{2}\right)$.

Based on these principles, a multitude of methods have been proposed.
Govindu~\cite{govindu2001combining} linearizes the problem via quaternions, while Martinec and Pajdla~\cite{martinec2007robust} relax the problem by omitting certain constraints on rotation matrices.
Eriksson~\etal~\cite{eriksson2018rotation} leverage strong duality.
The tractability condition of the problem is examined by Wilson~\etal~\cite{wilson2016rotations}.
Approaches utilizing semidefinite programming-based (SDP) relaxations~\cite{arie2012global,fredriksson2012simultaneous} ensure optimality guarantees by minimizing chordal distances~\cite{hartley2013rotation}.
Dellaert~\etal~\cite{dellaert2020shonan} sequentially elevate the problem into higher-dimensional rotations within $\text{SO}(n)$ to circumvent local minima where standard numerical optimization techniques might fail~\cite{levenberg1944method,marquardt1963algorithm}.
Various robust loss functions have been explored to handle outliers~\cite{hartley2011l1,chatterjee2013efficient,chatterjee2017robust,sidhartha2021all,zhang2023revisiting}.
Recently, learning-based methodologies have emerged.
NeuRoRa~\cite{purkait2020neurora}, MSP~\cite{yang2021end}, and PoGO-Net~\cite{li2021pogo} leverage Graph Neural Networks to eliminate outliers and to estimate absolute camera poses.
DMF-synch~\cite{tejus2023rotation} relies on matrix factorization techniques for pose extraction.
In this work, we use our own implementation of Chatterjee~\etal~\cite{chatterjee2013efficient} as a scalable approach that provides accurate results in presence of noisy as well as outlier-contaminated input rotations.

% due to its robustness and efficiency.
% Again, it can be replaced by any other rotation averaging scheme.

% \subsection{Translation Averaging}
% \todo{
% \begin{itemize}
%     \item pure translation direction based: 1DSFM, $L_\infty$, LUD (Revised / shapefit), BATA, Jiang, 
%     \item with point tracks (1DSFM, BATA)
% \end{itemize}
% }
% \vspace{0mm}
\noindent\textbf{Translation Averaging.}
After rotation averaging, the rotations $\bm{R}_i$ can be factored out from the camera poses.
What remains to be determined are the camera positions $\bm{c}_i$.
Translation averaging describes the problem of estimating global camera positions that are maximally consistent with the pairwise relative translations $\bm{t}_{ij}$ based on the constraint $\bm{t}_{ij} = \frac{\bm{c}_j - \bm{c}_i}{\Vert \bm{c}_j - \bm{c}_i \Vert}$.
However, due to noise and outliers as well as the unknown scale of the relative translations, the task is especially challenging.
In principle, the camera pose can be uniquely determined if the view graph has the property of \textit{parallel rigidity}. 
Parallel rigidity, also known as bearing rigidity, has been researched in different fields of computer vision~\cite{arrigoni2015computing,ozyesil2015stable}, robotics~\cite{kennedy2012identifying}, decision and control~\cite{zhao2016localizability} as well as computer-aided design~\cite{servatius1999constraining}.
Arrigoni~\etal\cite{arrigoni2018bearing} offers a unified review of this topic.

% The key challenge of translation averaging is on the lack of scale for the relative translation.
Different translation averaging methods have been proposed over the past years.
The pioneering work by Govindu~\cite{govindu2001combining} minimizes the cross-product between the relative camera locations and the observed directions.
Jiang~\etal~\cite{jiang2013global} linearizes the problem in units of triplets.
Wilson~\etal~\cite{wilson2014robust} optimizes the difference of directions directly and designs a dedicated outlier filtering mechanism.
Ozyesil~\etal~\cite{ozyesil2015robust} proposes a convex relaxation to the original problem and solves the Least Unsquared Deviations (LUD) problem with an $L_1$ loss for robustness.
% Goldstein~\etal\cite{goldstein2016shapefit} pre
Zhuang~\etal\cite{zhuang2018baseline} realizes the sensitivity of the LUD method in terms of camera baseline and proposes the Bilinear Angle-based Translation Averaging (BATA) error for optimization.
While significant improvements have been made in these works, translation averaging generally only works reliably when the view graph is well connected.
The problem is also inherently ill-posed and sensitive to noisy measurements when cameras are subject to or close to co-linear motion.
Furthermore, extraction of relative translation from two-view geometry is only possible with known camera intrinsics.
When such information is inaccurate, the extracted translations are not reliable.
Inspired by the observation that point tracks generally help in translation averaging~\cite{cui2015global,cui2017hsfm, arie2012global, wilson2014robust, cui2015linear,holynski2020reducing}, in our proposed system, we \textit{skip} the step of translation averaging.
Instead, we directly perform a joint estimation of the camera and point positions.
We refer to this step as global positioning with details introduced in Section~\ref{sec:global_positioning}.

% We refer to this s
% Wilson and Snavely \cite{wilson2014robust} proposes to use the discrepancy between the translation direction as cost function. 
% They additionally proposed a delicately designed outlier filter mechanism based on projection of translation direction into 1D subspace.
% Olsson and Envist \cite{olsson2011stable}, followed by Moloun, et. al \cite{moulon2013global} employs $L_\infty$ for robust 3D Points and camera postion estimation.
% Ozyesil \etal~\cite{ozyesil2015robust} proposes a convex relaxation for estimating the camera position.
% Zhuang \etal~\cite{zhuang2018baseline} propose a baseline desensitized cost function for translation averaging. 
% Above methods generally regard camera intrinsics are known beforehand, however, such an assumption does not always hold true.
% Sweeney \etal ~\cite{sweeney2015optimizing} proposes to optimize the fundamental matrices on view-graph for improving the view consistence, and enforcing the essential matrix constraint to improve the camera intrinsics.
% Yet, as the fundamental matrix are noisy and error-prone, it requires dense image pairs to achieve robustness.
% \todo{THIS PART IS TOO SHORT NOW}

% \vspace{0mm}
\noindent\textbf{Structure for Camera Pose Estimation.}
% \todo{
% \begin{itemize}
%     \item arie2012global, 1dsfm, bata, cui2015linear
% \end{itemize}
% }
% Many works have realized the usefulness of points in determining the camera position.
Several works have explored incorporating 3D structure into camera position estimation.
Ariel~\etal~\cite{arie2012global} directly use the correspondences in two-view geometry for estimating global translation.
Wilson~\etal~\cite{wilson2014robust} discovered that 3D points can be treated in a similar manner as the camera centers and, thus, can be easily incorporated into the optimization.
Cui~\etal~\cite{cui2015linear} extends \cite{jiang2013global} by including point tracks into the optimization problem with linear relations.
To reduce scale drifting, Holynski~\etal~\cite{holynski2020reducing} integrates line and plane features into the optimization problem.
Manam~\etal~\cite{manam2022correspondence} incorporates the correspondences by reweighing the relative translation in the optimization.
LiGT~\cite{cai2021pose} proposes a ``pose only'' method for solving the camera positions with a linear global translation constraint imposed by points.
The common theme of these works is that incorporating constraints on the 3D scene structure aids the robustness and accuracy of camera position estimation, which we take as an inspiration for our work.
% Those work jointly agree on the benefits brought by the 3D points.
% Though BATA~\cite{zhuang2018baseline} argues it is not neccesary to include point tracks

% Instead of directly adding the points into the problem formulation, 

\subsection{Global Structure and Pose Refinement}

After recovering the cameras, the global 3D structure can be obtained via triangulation.
Together with the camera extrinsics and intrinsics, the 3D structure is then typically refined using global bundle adjustment.

% \vspace{0mm}
\noindent\textbf{Global Triangulation.}
Given two-view matches, transitive correspondences can be leveraged for boosting completeness and accuracy~\cite{schoenberger2016sfm}.
Moulon~\etal~\cite{moulon2012unordered} presents an efficient way for concatenating tracks.
Triangulating multi-view points has a long research history~\cite{hartley1997triangulation, lu2007fast}.
Common practices for such a task are the direct linear transformation (DLT) and midpoint methods \cite{hartley1997triangulation, abdel2015direct, hartley2003multiple}.
% The gold standard for such a task is the direct linear transformation (DLT) and midpoint methods \cite{hartley1997triangulation, abdel2015direct, hartley2003multiple}.
% $L_2$ based and midpoint based method are the common choice for the triangulation.
Recently, LOST~\cite{henry2023absolute} was proposed as an uncertainty-based triangulation.
Yet, the above triangulation mechanisms often break in the presence of arbitrary levels of outliers.
In this regard, Sch{\"o}nberger~\etal~\cite{schoenberger2016sfm} proposes a RANSAC-based triangulation scheme, seeking to establish multiple point tracks in the presence of mismatches.
Instead, our approach directly estimates 3D points by a single, joint global optimization method together with the camera positions (see Section~\ref{sec:global_positioning}).

% \vspace{0mm}
\noindent\textbf{Global Bundle Adjustment}
is essential in obtaining accurate final 3D structure $\bm{X}_k \in \mathbb{R}^3$, camera extrinsics $\bm{P}_i$ and camera intrinsics $\pi_i$. 
It is formulated as a joint robust optimization by minimizing reprojection errors as
\begin{equation}
    \argmin_{\pi , \bm{P}, \bm{X} } \sum_{i, k} \rho \left(\|\pi_i(\bm{P}_i, \bm{X}_k) - x_{ik}\|_2\right).
\end{equation}
Please refer to Triggs~\etal~\cite{bundleadjustment2000triggs} for a comprehensive review of bundle adjustment.

\subsection{Hybrid Structure-from-Motion}

To combine the robustness of incremental and efficiency of global SfM, previous works have formulated hybrid systems.
HSfM~\cite{cui2017hsfm} proposes to incrementally estimate camera positions with rotations.
Liu~\etal~\cite{liu2023efficient} proposes a graph partitioning method by first dividing the whole set of images into overlapping clusters.
Within each cluster, camera poses are estimated via a global SfM method.
However, such methods are still not applicable when camera intrinsics are inaccurate according to their formulation.
Our method overcomes this limitation by different modeling of the objective in the global positioning step.

\subsection{Frameworks for Structure-from-Motion}

Multiple open-source SfM pipelines are available.
The incremental SfM paradigm is currently the most widely used due to its robustness and accuracy in real-world scenarios. 
Bundler~\cite{snavely2006photo} and VisualSfM~\cite{wu2013towards} are systems dating back a decade ago.
Building upon these, Sch{\"o}nberger~\etal~\cite{schoenberger2016sfm} developed COLMAP, a general-purpose SfM and multi-view stereo~\cite{schoenberger2016mvs} system. COLMAP is versatile and has demonstrated robust performance across many datasets, making it the standard tool for image-based 3D reconstruction in recent years.

Several open-source pipelines are available for global SfM as well. 
OpenMVG~\cite{moulon2016openmvg} stands out as a prominent framework in this category. 
Starting with geometrically verified matches, it estimates the relative pose using a contrario RANSAC~\cite{moisan2012automatic}.
Following this, OpenMVG assesses rotation consistency through adjusted cycle length weighting to eliminate outlier edges and solves for global rotation using the remaining edges with a sparse eigenvalue solver. 
Global translations are refined through the trifocal tensor and then subjected to translation averaging using the $L_\infty$ method. 
Finally, OpenMVG performs global triangulation via per-point optimization and a global bundle adjustment.

Theia~\cite{theia-manual} is another well-established global SfM pipeline. 
It adopts a similar approach to OpenMVG by initially estimating global rotations via averaging and then estimating camera positions through translation averaging. 
For rotation averaging, Theia employs a robust scheme from \cite{chatterjee2013efficient}. 
For translation averaging, it defaults to using the LUD method \cite{ozyesil2015robust}. 
The pipeline concludes with global triangulation and bundle adjustment, similar to OpenMVG.

Several learning-based pipelines are available.
PixSfM~\cite{lindenberger2021pixel} proposes a joint refinement mechanism over features and structure to achieve sub-pixel accurate reconstruction and can be combined with our system.
VGGSfM~\cite{wang2023vggsfm} proposes an end-to-end learning framework for the SfM task, and Zhuang~\etal~\cite{zhang2024cameras} proposes to operate on pixel-wise correspondences to regress camera position directly.
% However, the scalability of these two methods is limited to 10s of images.
However, these two methods are limited to handling tens of images.

In this paper, we propose a new end-to-end global SfM pipeline and release it to the community as an open-source contribution to facilitate downstream applications and further research.

\begin{figure}[t]
    \centering
     \includegraphics[width=\textwidth]{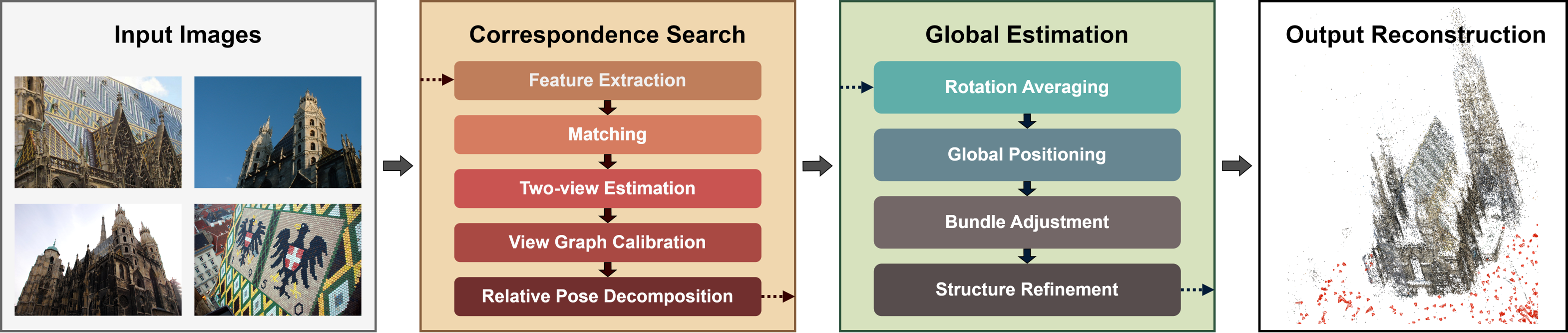}\\
    % \vspace{-5px}
    \caption{ 
    Pipeline of proposed GLOMAP system, a global Structure-from-Motion framework, that distinguishes itself from other global methods by merging the translation averaging and triangulation phase into a single global positioning step.
    }
    \label{fig:pipeline}
    % \vspace{-15px}
\end{figure}

% \vspace{-5px}
\section{Technical Contributions}
% \section{Proposed Pipeline}

This section presents our key technical contributions to improve upon the state of the art in global SfM (\cf~Fig.~\ref{fig:pipeline}) and close the gap to incremental SfM in terms of robustness and accuracy.

\subsection{Feature Track Construction}
% For the joint global camera position and structure estimation, 
Feature tracks must be carefully constructed as to achieve accurate reconstruction.
% enable robust convergence of the optimization.
% To keep global positioning runtimes low, we first sub-sample the edges in the view graph to have around 50 edges per node by randomly keeping edges with probability $\frac{50\sum_k \deg(I_k)}{N \deg(I_i) \deg(I_j)}$.
We start by only considering inlier feature correspondences produced by two-view geometry verification.
In this step, we distinguish between the initial classification of the two-view-geometry~\cite{schoenberger2016sfm}: if homography $\bm{H}$ best describes the two-view geometry, we use $\bm{H}$ for the verification of inliers.
% \TODO{cite COLMAP}
The same principle is applied to essential matrix $\bm{E}$ and fundamental matrix $\bm{F}$.
We further filter outliers by performing a cheirality test~\cite{hartley1993cheirality,werner2001cheirality}.
Matches that are close to any of the epipoles or have small triangulation angles are also removed to avoid singularities due to large uncertainties.
After pairwise filtering of all view graph edges, we form feature tracks by concatenating all remaining matches.

% % we use Homography if the pair is uncalibrated, we use fundamental matrix $\bm{F}$ for testing inliers, and if i

% In this step, we use the pre-determined type of two-view-geometry for establishing 
% To decrease the number of outliers in the tracks, we perform filtering in the two-view matches.

% To initialize the binda
% \subsection{Track Selection}
\subsection{Global Positioning of Cameras and Points \label{sec:global_positioning}} 
% Joint Triangulation and Camera Center Estimation

\begin{figure}[t]
    \centering
     \includegraphics[width=\textwidth]{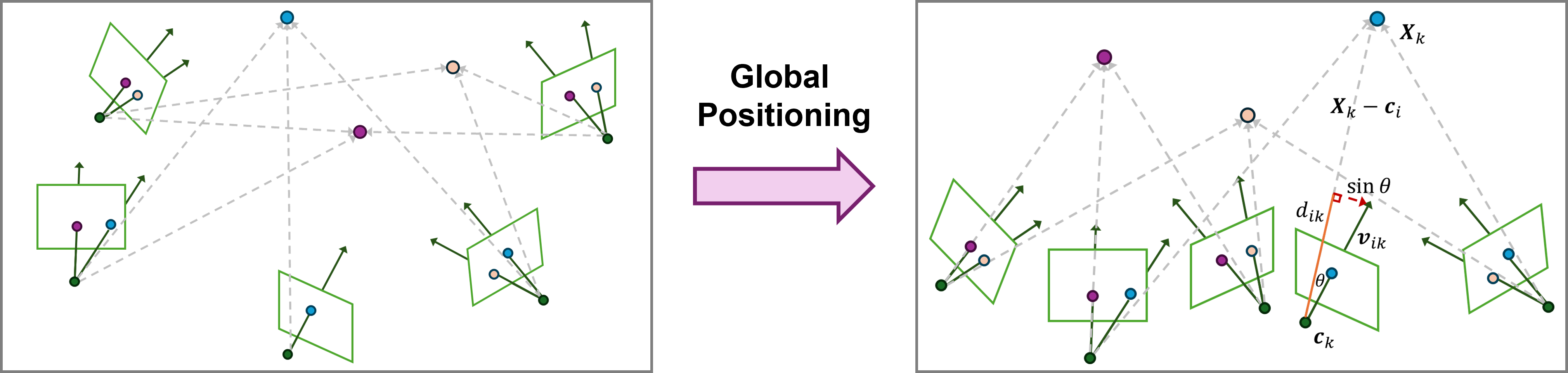}\\
    % \vspace{-5px}
    \caption{ 
    \textbf{Global Positioning.}
    The left figure visualizes the initial configuration, depicting randomly initialized cameras and points. 
    Black arrows, traversing through colored circles on the image planes, denote the measurements. 
    Dashed lines represent the actual image rays, which are subject to optimization by adjusting the positions of the  points and the cameras while their orientations remain constant. 
    The right figure displays the outcome following the minimization of angles between the measurements (solid lines) and the image rays from the 3D points (dashed lines).
    }
    \label{fig:global_positioning}
    % \vspace{-10px}
\end{figure}

This step aims to jointly recover point and camera positions (see Fig.~\ref{fig:global_positioning}).
% With camera rotation fixed and tracks initialized, the next step is to estimate camera center and point location.
Instead of performing translation averaging followed by global triangulation, we directly perform a joint global triangulation and camera position estimation.
Different from most previous works, our objective function is initialization-free and consistently converges to a good solution in practice.
In standard incremental and global SfM systems, feature tracks are verified and optimized by reprojection errors to ensure reliable and accurate triangulations.
However, the reprojection error across multiple views is highly non-convex, thus requiring careful initialization.
Moreover, the error is unbounded, so it is not robust to outliers.
To overcome these challenges, we build upon the objective function proposed by~\cite{zhuang2018baseline} and use normalized direction differences as an error measure.
The original formulation was proposed in terms of the relative translations, whereas, in our formulation, we discard the relative translation constraints and only include camera ray constraints.
Concretely, our problem is modeled and optimized as:
\begin{equation}
\begin{split}
    % &\argmin_{\bm{X}, \bm{c}} \sum_{i,k} \rho\left(\bm{v}_{ik} - d_{ij} (\bm{X}_k - \bm{c}_i)\right)^2,\\
    % &\text{subject to} \quad d_{ij} \geq 0,
    \argmin_{\bm{X}, \bm{c}, d} \sum_{i,k} \rho\left( \| \bm{v}_{ik} - d_{ik} (\bm{X}_k - \bm{c}_i) \|_2 \right),\quad
    \text{subject to} \quad d_{ik} \geq 0,
\end{split}
\label{eq:bata}
\end{equation}
where the $\bm{v}_{ik}$ is the globally rotated camera ray observing point $\bm{X}_k$ from camera $\bm{c}_i$, while $d_{ik}$ is a normalizing factor.
% $\bm{v}_{ik}$ is normalized to have unit length.
We use Huber~\cite{huber1992robust} as a robustifier $\rho$ and Levenberg–Marquardt~\cite{levenberg1944method} from Ceres~\cite{ceres_solver} as the optimizer.
% We fix the gauge freedom by keeping the one camera constant.
All point and camera variables are initialized by a uniform random distribution in the range $[-1, 1]$ while the normalization factors are initialized as $d_{ik} = 1$.
We down-weight terms involving cameras with unknown intrinsics by factor 2 to reduce their influence.
% This designs is of the same form as BATA \cite{zhuang2018baseline}, however, we discard the camera to camera constraints, and directly formulate it on tracks only.

Compared to reprojection errors, this has several advantages.
The first is robustness.
% While reprojection errors are unbounded, the above is equivalent to
While reprojection errors are unbounded, the above is equivalent to
%
% $\sin(\theta) (\theta\in[0,\frac{\pi}{2}), 1 (\theta\in[\frac{\pi}{2}, \pi])$
% $\begin{array}
%      & \sin{\theta} & \theta\\
%      & 1\\
% \end{array}
% % \sin{\theta}
% % \right,
% $
\begin{equation}
\left\{
\begin{split}
     & \sin{\theta} & \text{if}\quad \theta\in[0, \pi / 2),\\
     & 1 & \text{if}\quad \theta \in [\pi / 2, \pi],\\
\end{split}
\right.
\end{equation}
where $\theta$ is the angle between $\bm{v}_{ik}$ and $\bm{X}_k - \bm{c}_i$ for optimal $d_{ik}$ \cite{zhuang2018baseline}. 
Thus, the error is strictly bounded to the range $[0,1]$.
As such, outliers do not heavily bias the result.
Secondly, the objective function, as we experimentally show, converges reliably with random initialization due to its bilinear form~\cite{zhuang2018baseline}.
% Second, according to the authors of the original paper~\cite{}
% Second is the convexity of the objective function.
% When the robust loss is chosen to be a convex function, \eg Huber Loss \cite{huber1992robust}, the objective function is convex in the region where $d_{ik}$ are positive, which suggests that the optimization can be done in an efficient and theoretically guaranteed manner.

Compared with classical translation averaging, discarding the relative translation terms in the optimization has two key advantages.
First, the applicability of our method on datasets with inaccurate or unknown camera intrinsics as well as degenerate cameras not following the expected pinhole model (\eg, when dealing with arbitrary internet photos).
This is because the knowledge of accurate intrinsics is required to solve for relative translation.
When they deviate from the expected value, the estimated two-view translations suffer from large errors.
Since translation averaging is inherently ill-posed due to unknown scale, recovering camera positions from noisy and outlier-contaminated observations is challenging, especially as relative translation errors exacerbate with longer baselines.
Our proposed pipeline, instead, relies on careful filtering of two-view geometry and the error is defined \wrt the camera rays.
Hence, poor camera intrinsics only bias the estimation of individual cameras instead of also biasing other overlapping cameras.
% , and it can be balanced out as the image rays from this camera undergo the same distortion.
Second, the applicability of global SfM in co-linear motion scenarios, which is a known degenerate case for translation averaging.
Compared to pairwise relative translations, feature tracks constrain multiple overlapping cameras.
As such, our proposed pipeline can deal more reliably in common forward or sideward motion scenarios (see Section~\ref{sec:ablation_pt}).

% Above two properties indicate that the optimization is well-conditioned, and empirical results show the success of the method.

% These two properties makes the problem well-conditioned, and can be 

% 1DSfM \cite{xx} and BATA \cite{xx} shows how camera points can be be treated the same way as cameras in the translation averaging problem.
% In
% 1DSfM \cite{xx} shows how camera points can be be treated the same way as cameras and BATA \cite{xx} change the cost from from direction different, to
% The optimization follows a similar 

% \todo{EQUATIONS HERE}
% 1DSfM \cite{xx} has shown that camera points can be treated the same way as cameras in the original translation averaging problem.
% BATA \cite{xx} following a similar 

\subsection{Global Bundle Adjustment}
% The global positioning step provides a robust estimation for cameras and points.
% However, the accuracy is limited, especially when camera intrinsics are not known in advance.
% As a further refinement, we perform two rounds of global bundle adjustment using Levenberg-Marquardt and the Huber loss as a robustifier, optimizing jointly over camera extrinsics, intrinsics, and points.
% Before constructing the first bundle adjustment problem, we apply a pre-filtering of 3D point observations based on the angular error while allowing a larger error for uncalibrated cameras.
% Then, before constructing the second bundle adjustment problem, we filter tracks based on reprojection errors in image space.

The global positioning step provides a robust estimation for cameras and points.
However, the accuracy is limited, especially when camera intrinsics are not known in advance.
As a further refinement, we perform several rounds of global bundle adjustment using Levenberg-Marquardt and the Huber loss as a robustifier.
Within each round, camera rotations are first fixed, then jointly optimized with intrinsics and points.
% optimizing jointly over camera extrinsics, intrinsics, and points.
% We split each global bundle adjustment into two stages.
% In the first stage, we keep camera rotation fixed, while in the second stage, we optimize jointly over camera extrinsics, intrinsics, and points.
Such a design is particularly important for reconstructing sequential data.
Before constructing the first bundle adjustment problem, we apply a pre-filtering of 3D point observations based on the angular error while allowing a larger error for uncalibrated cameras.
Afterward, we filter tracks based on reprojection errors in image space.
Iterations are halted when the ratio of filtered tracks falls below 0.1$\%$.
% Empirically, we found 3 rounds are sufficient, and we halt the process early if the ratio of filtered tracks falls below 0.1$\%$.
% Then, before constructing the second bundle adjustment problem, we filter tracks based on reprojection errors in image space.

\subsection{Camera Clustering}

For images collected from the internet, non-overlapping images can be wrongly matched together.
Consequently, different reconstructions collapse into a single one.
To overcome this issue, we post-process the reconstruction by performing clustering of cameras.
First, the covisibility graph $\mathcal{G}$ is constructed by counting the number of visible points for each image pair.
Pairs with fewer than 5 counts are discarded as the relative pose cannot be reliably determined below this number, and the median of the remaining pairs is used to set inlier threshold $\tau$.
Then, we find well-constrained clusters of cameras by finding strongly connected components in $\mathcal{G}$.
Such components are defined by only connecting pairs with more than $\tau$ counts.
% edges that pass the angular filtering of globally estimated rotations and have a large number of inlier feature correspondences.
Afterwards, we carefully attempt to merge two strong components, if there are at least two edges with more than $0.75\tau$ counts.
% least 75\% of inlier correspondences of the initial threshold.
We recursively repeat this procedure until no more clusters can be merged.
Each connected component is output as a separate reconstruction.
% For all our experiments, we then only reconstruct the largest connected component, but others could also be independently reconstructed in general.

\subsection{Proposed Pipeline}
The pipeline of the proposed method is summarized in Fig.~\ref{fig:pipeline}.
It consists of two major components: \textit{correspondence search} and \textit{global estimation}.
For correspondence search, it starts with feature extractions and matching.
Two-view geometry, including fundamental matrix, essential matrix, and homography, are estimated from the matches.
Geometrically infeasible matches are excluded.
% Given a collection of input images, features are first extracted and matched.
Then, view graph calibration is performed similar to Sweeney~\etal~\cite{sweeney2015optimizing} on geometrically verified image pairs.
With the updated camera intrinsics, relative camera poses are estimated.
As for global estimation, global rotations are estimated via averaging~\cite{chatterjee2013efficient} and inconsistent relative poses are filtered by thresholding the angular distance between $\bm{R}_{ij}$ and $\bm{R}_j \bm{R}_i^\top$.
Then, the positions of cameras and points are jointly estimated via global positioning, followed by global bundle adjustment.
Optionally, the accuracy of the reconstruction can be further boosted with structure refinement.
Within this step, points are retriangulated with the estimated camera pose, and rounds of global bundle adjustment are performed.
Camera clustering can also be applied to achieve coherent reconstructions.

\begin{table}[t]
    \centering
    \caption{Result on ETH3D SLAM~\cite{Schops_2019_CVPR} dataset. Each row represents the average results on scenes with the same prefix. The proposed system outperforms global SfM baselines by a large margin, and also achieves better results than COLMAP~\cite{schoenberger2016sfm}.}
    % \vspace{-5px}
    \resizebox{\textwidth}{!}{
    \begin{tabular}{l l c c c a l c c c a l c c c a
    l c c c d
    } \toprule 
        % && \multicolumn{4}{c}{OpenMVG} && \multicolumn{4}{c}{Theia} && \multicolumn{4}{c}{Ours} 
        % && \multicolumn{4}{c}{COLMAP}
        && \multicolumn{4}{c}{Recall @ 0.1m} && \multicolumn{4}{c}{AUC @ 0.1m} && \multicolumn{4}{c}{AUC @ 0.5m} 
        && \multicolumn{4}{c}{Time (s)}
        \\
        \cmidrule{3-6} \cmidrule{8-11} \cmidrule{13-16} \cmidrule{18-21}
        && \tiny{OpenMVG} & \tiny{~~Theia~~} & \tiny{GLOMAP} & \tiny{COLMAP} && \tiny{OpenMVG} & \tiny{~~Theia~~} & \tiny{GLOMAP} & \tiny{COLMAP} && \tiny{OpenMVG} & \tiny{~~Theia~~} & \tiny{GLOMAP} & \tiny{COLMAP} && \tiny{OpenMVG} & \tiny{~~Theia~~} & \tiny{GLOMAP} & \tiny{COLMAP} \\  \midrule
cables && \cellcolor{tabsecond}76.8 & \cellcolor{tabfirst}88.0 & \cellcolor{tabfirst}88.0 & 68.0 && 59.1 & \cellcolor{tabsecond}60.2 & \cellcolor{tabfirst}76.2 & 56.5 && 77.4 & \cellcolor{tabsecond}82.4 & \cellcolor{tabfirst}85.6 & 70.4 && \cellcolor{tabfirst}103.1 & 339.5 & \cellcolor{tabsecond}195.6 & 2553.4 \\
camera && 2.2 & 26.6 & \cellcolor{tabsecond}32.6 & \cellcolor{tabfirst}34.8 && 2.0 & 12.9 & \cellcolor{tabsecond}21.9 & \cellcolor{tabfirst}24.7 && 2.2 & 26.2 & \cellcolor{tabsecond}30.8 & \cellcolor{tabfirst}33.2 && \cellcolor{tabfirst}1.5 & \cellcolor{tabsecond}5.3 & 10.0 & 196.2 \\
ceiling && 6.4 & 22.8 & \cellcolor{tabfirst}28.7 & \cellcolor{tabsecond}28.6 && 2.9 & \cellcolor{tabsecond}17.0 & \cellcolor{tabfirst}22.3 & 15.3 && 8.2 & 21.7 & \cellcolor{tabfirst}27.4 & \cellcolor{tabsecond}26.3 && \cellcolor{tabsecond}78.3 & \cellcolor{tabfirst}52.6 & 111.0 & 1057.7 \\
desk && 28.0 & \cellcolor{tabsecond}32.2 & \cellcolor{tabfirst}32.3 & 24.4 && 16.0 & \cellcolor{tabfirst}29.2 & \cellcolor{tabsecond}28.5 & 21.1 && \cellcolor{tabsecond}28.6 & \cellcolor{tabfirst}31.6 & \cellcolor{tabfirst}31.6 & 23.7 && 376.2 & \cellcolor{tabsecond}195.3 & \cellcolor{tabfirst}150.0 & 1115.1 \\
einstein && 32.9 & \cellcolor{tabsecond}47.8 & \cellcolor{tabfirst}48.5 & 33.6 && 22.7 & \cellcolor{tabsecond}32.1 & \cellcolor{tabfirst}36.5 & 25.4 && 34.2 & \cellcolor{tabsecond}46.7 & \cellcolor{tabfirst}47.9 & 36.7 && 150.3 & \cellcolor{tabfirst}70.5 & \cellcolor{tabsecond}142.1 & 1230.8 \\
kidnap && \cellcolor{tabsecond}73.1 & \cellcolor{tabfirst}73.3 & \cellcolor{tabfirst}73.3 & \cellcolor{tabfirst}73.3 && 63.4 & 62.3 & \cellcolor{tabfirst}70.3 & \cellcolor{tabsecond}68.5 && 71.2 & 71.1 & \cellcolor{tabfirst}72.7 & \cellcolor{tabsecond}72.3 && \cellcolor{tabfirst}114.4 & 356.7 & \cellcolor{tabsecond}144.3 & 731.2 \\
large && 35.4 & \cellcolor{tabsecond}48.6 & \cellcolor{tabfirst}49.0 & 44.5 && 18.1 & \cellcolor{tabsecond}37.8 & \cellcolor{tabfirst}45.8 & 20.7 && 33.7 & \cellcolor{tabsecond}46.6 & \cellcolor{tabfirst}48.4 & 43.4 && 91.9 & \cellcolor{tabfirst}60.2 & \cellcolor{tabsecond}77.6 & 983.8 \\
mannequin && 49.1 & \cellcolor{tabsecond}62.2 & \cellcolor{tabfirst}67.4 & 59.3 && 36.7 & 46.5 & \cellcolor{tabfirst}61.4 & \cellcolor{tabsecond}52.8 && 49.1 & \cellcolor{tabsecond}59.7 & \cellcolor{tabfirst}66.4 & 58.4 && \cellcolor{tabsecond}33.5 & \cellcolor{tabfirst}29.7 & 44.3 & 301.2 \\
motion && \cellcolor{tabsecond}18.8 & 16.9 & \cellcolor{tabfirst}39.8 & 17.7 && 11.0 & 11.9 & \cellcolor{tabfirst}22.5 & \cellcolor{tabsecond}12.9 && \cellcolor{tabsecond}23.3 & 19.2 & \cellcolor{tabfirst}45.9 & 19.7 && 859.7 & \cellcolor{tabfirst}109.0 & \cellcolor{tabsecond}788.9 & 9995.1 \\
planar && \cellcolor{tabsecond}30.6 & \cellcolor{tabfirst}100.0 & \cellcolor{tabfirst}100.0 & \cellcolor{tabfirst}100.0 && 12.5 & 97.8 & \cellcolor{tabfirst}98.7 & \cellcolor{tabsecond}98.3 && 38.0 & \cellcolor{tabsecond}99.6 & \cellcolor{tabfirst}99.7 & \cellcolor{tabfirst}99.7 && \cellcolor{tabsecond}313.8 & \cellcolor{tabfirst}167.5 & 533.3 & 2349.7 \\
plant && 77.0 & 89.1 & \cellcolor{tabfirst}93.3 & \cellcolor{tabsecond}92.8 && 62.9 & 75.3 & \cellcolor{tabsecond}82.0 & \cellcolor{tabfirst}82.3 && 77.1 & 88.2 & \cellcolor{tabfirst}93.4 & \cellcolor{tabsecond}92.5 && \cellcolor{tabfirst}21.7 & 35.7 & \cellcolor{tabsecond}28.7 & 202.7 \\
reflective && 12.6 & 16.1 & \cellcolor{tabsecond}22.0 & \cellcolor{tabfirst}26.2 && 6.7 & 9.0 & \cellcolor{tabfirst}12.1 & \cellcolor{tabsecond}9.2 && 16.5 & 23.0 & \cellcolor{tabsecond}31.3 & \cellcolor{tabfirst}33.5 && 721.3 & \cellcolor{tabfirst}118.3 & \cellcolor{tabsecond}434.4 & 6573.9 \\
repetitive && 26.3 & \cellcolor{tabsecond}28.5 & \cellcolor{tabfirst}32.7 & \cellcolor{tabsecond}28.5 && 23.9 & 15.2 & \cellcolor{tabfirst}29.2 & \cellcolor{tabsecond}27.2 && 25.8 & 27.0 & \cellcolor{tabfirst}32.0 & \cellcolor{tabsecond}28.3 && \cellcolor{tabfirst}63.2 & 136.8 & \cellcolor{tabsecond}74.5 & 561.1 \\
sfm && 83.2 & \cellcolor{tabsecond}94.1 & \cellcolor{tabfirst}97.0 & 55.9 && \cellcolor{tabsecond}57.9 & 53.7 & \cellcolor{tabfirst}79.6 & 35.7 && 80.0 & \cellcolor{tabsecond}88.7 & \cellcolor{tabfirst}95.2 & 55.8 && \cellcolor{tabfirst}91.7 & \cellcolor{tabsecond}170.5 & 239.7 & 469.6 \\
sofa && 11.2 & 22.0 & \cellcolor{tabsecond}23.9 & \cellcolor{tabfirst}32.2 && 5.5 & 13.1 & \cellcolor{tabsecond}22.1 & \cellcolor{tabfirst}28.6 && 10.3 & 21.3 & \cellcolor{tabsecond}23.5 & \cellcolor{tabfirst}31.6 && \cellcolor{tabsecond}9.1 & \cellcolor{tabfirst}8.9 & 10.1 & 157.3 \\
table && 79.1 & 93.7 & \cellcolor{tabsecond}94.3 & \cellcolor{tabfirst}99.9 && 68.2 & 69.2 & \cellcolor{tabsecond}84.3 & \cellcolor{tabfirst}95.6 && 76.9 & 89.2 & \cellcolor{tabsecond}92.3 & \cellcolor{tabfirst}99.1 && \cellcolor{tabsecond}182.0 & \cellcolor{tabfirst}97.8 & 221.5 & 2777.7 \\
vicon && 64.6 & \cellcolor{tabsecond}84.2 & \cellcolor{tabfirst}97.0 & 81.1 && 20.4 & \cellcolor{tabsecond}57.1 & \cellcolor{tabfirst}80.5 & 38.9 && 71.7 & \cellcolor{tabsecond}87.6 & \cellcolor{tabfirst}93.7 & 75.0 && \cellcolor{tabsecond}50.9 & 88.2 & \cellcolor{tabfirst}46.2 & 474.8 \\
\midrule
\textit{Average} && 48.2 & \cellcolor{tabsecond}62.8 & \cellcolor{tabfirst}66.4 & 57.9 && 34.9 & 46.0 & \cellcolor{tabfirst}57.0 & \cellcolor{tabsecond}47.6 && 48.6 & \cellcolor{tabsecond}61.1 & \cellcolor{tabfirst}65.7 & 57.9 && \cellcolor{tabsecond}120.8 & \cellcolor{tabfirst}91.8 & 133.5 & 1115.4 \\
      \bottomrule
    \end{tabular}
    }
    \label{tbl:eth3d_slam}
    % \vspace{-3px}
\end{table}
\section{Experiments}
To demonstrate the performance of our proposed \textit{GLOMAP} system, we conduct extensive experiments on various datasets, ranging from calibrated to uncalibrated and from unordered to sequential scenarios.
More specifically, we compare against the state-of-the-art frameworks (OpenMVG~\cite{moulon2016openmvg}, Theia~\cite{theia-manual}, COLMAP~\cite{schoenberger2016sfm}) on the ETH3D~\cite{schoeps2017cvpr,Schops_2019_CVPR}, LaMAR~\cite{sarlin2022lamar}, Image Matching Challenge 2023 (IMC 2023) ~\cite{IMC2023}, and MIP360~\cite{barron2022mip} datasets.
Furthermore, we present ablations to study the behavior of different components of our proposed system.

\begin{table}[t]
    \centering
    \caption{Results on ETH3D MVS (rig)~\cite{schoeps2017cvpr}. Proposed GLOMAP largely outperforms other SfM systems by a large margin while maintaining the efficiency of global SfM.}
    % \vspace{-5px}
    \resizebox{\textwidth}{!}{
    \begin{tabular}{l l c c c a l c c c a l c c c a
    l c c c d
    } \toprule 
        % && \multicolumn{4}{c}{OpenMVG} && \multicolumn{4}{c}{Theia} && \multicolumn{4}{c}{Ours} 
        % && \multicolumn{4}{c}{COLMAP}
        && \multicolumn{4}{c}{AUC @ $1^\circ$} && \multicolumn{4}{c}{AUC @ $3^\circ$} && \multicolumn{4}{c}{AUC @ $5^\circ$} 
        && \multicolumn{4}{c}{Time (s)}
        \\
        \cmidrule{3-6} \cmidrule{8-11} \cmidrule{13-16} \cmidrule{18-21}
        && \tiny{OpenMVG} & \tiny{~~~Theia~~~} & \tiny{GLOMAP} & \tiny{COLMAP} && \tiny{OpenMVG} & \tiny{~~~Theia~~~} & \tiny{GLOMAP} & \tiny{COLMAP} && \tiny{OpenMVG} & \tiny{~~~Theia~~~} & \tiny{GLOMAP} & \tiny{COLMAP} && \tiny{OpenMVG} & \tiny{~~~Theia~~~} & \tiny{GLOMAP} & \tiny{COLMAP} \\  \midrule
delivery\_area && 0.0 & 47.8 & \cellcolor{tabfirst}75.9 & \cellcolor{tabsecond}66.6 && 0.1 & 81.0 & \cellcolor{tabfirst}91.2 & \cellcolor{tabsecond}87.5 && 0.3 & 88.2 & \cellcolor{tabfirst}94.6 & \cellcolor{tabsecond}92.2 && \cellcolor{tabfirst}235.7 & \cellcolor{tabsecond}259.5 & 519.9 & 1745.9 \\
electro && 0.2 & 25.9 & \cellcolor{tabfirst}47.5 & \cellcolor{tabsecond}38.4 && 1.8 & 61.6 & \cellcolor{tabfirst}72.9 & \cellcolor{tabsecond}65.2 && 2.9 & 73.5 & \cellcolor{tabfirst}81.7 & \cellcolor{tabsecond}75.2 && \cellcolor{tabfirst}99.9 & \cellcolor{tabsecond}151.8 & 429.2 & 3283.1 \\
forest && 0.0 & 65.6 & \cellcolor{tabsecond}74.1 & \cellcolor{tabfirst}74.7 && 0.0 & 87.1 & \cellcolor{tabsecond}90.0 & \cellcolor{tabfirst}90.3 && 0.1 & 91.8 & \cellcolor{tabsecond}93.5 & \cellcolor{tabfirst}93.7 && \cellcolor{tabfirst}306.6 & \cellcolor{tabsecond}563.2 & 1658.4 & 7571.6 \\
playground && 0.0 & \cellcolor{tabsecond}23.1 & \cellcolor{tabfirst}40.0 & 0.0 && 0.1 & \cellcolor{tabsecond}62.1 & \cellcolor{tabfirst}72.5 & 0.0 && 0.2 & \cellcolor{tabsecond}74.1 & \cellcolor{tabfirst}81.0 & 0.0 && \cellcolor{tabfirst}121.5 & 598.6 & 1008.3 & \cellcolor{tabsecond}350.2 \\
terrains && 0.7 & 39.6 & \cellcolor{tabfirst}50.2 & \cellcolor{tabsecond}48.0 && 2.6 & 71.2 & \cellcolor{tabfirst}78.1 & \cellcolor{tabsecond}76.8 && 3.1 & 79.6 & \cellcolor{tabfirst}85.3 & \cellcolor{tabsecond}84.3 && \cellcolor{tabfirst}62.6 & \cellcolor{tabsecond}179.4 & 353.0 & 1333.9 \\
\midrule
\textit{Average} && 0.2 & 40.4 & \cellcolor{tabfirst}57.6 & \cellcolor{tabsecond}45.5 && 0.9 & \cellcolor{tabsecond}72.6 & \cellcolor{tabfirst}80.9 & 64.0 && 1.3 & \cellcolor{tabsecond}81.4 & \cellcolor{tabfirst}87.2 & 69.1 && \cellcolor{tabfirst}165.3 & \cellcolor{tabsecond}350.5 & 793.8 & 2857.0 \\
      \bottomrule
    \end{tabular}
    }
    % \caption{Results on ETH3D MVS~\cite{schoeps2017cvpr}, rig sequences. The proposed GLOMAP largely outperforms other SfM systems by a large margin.}
    \label{tbl:eth3d_rig}
    % \vspace{-10px}
\end{table}

\noindent
\textbf{Metrics.}
For all evaluations, we adopt two standard metrics.
For unordered image data, we report the AUC (Area Under the recall Curve) scores calculated from the maximum of relative rotation and translation error between every image pair, similar to \cite{Schops_2019_CVPR,IMC2023,he2023detector}.
Such an error formulation considers the deviation between every possible camera pair.
For sequential image data, we report the AUC scores calculated from the camera position error after globally aligning the reconstruction to the ground truth using a robust RANSAC scheme~\cite{schoenberger2016sfm}.
When images are taken in sequences, especially in the case when cameras are nearly co-linear, the relative error does not capture the scale drift well. 
% Additionally, a slight change in camera position may result in a large error.
Thus, we directly focus on the camera positions.
For a fair comparison, we use the same feature matches as input to all methods and thus also exclude correspondence search from the reported runtimes.
We also tried using OpenMVG's and Theia's correspondence search implementations but consistently obtained better results using COLMAP.
We employ Kapture~\cite{kapture} for importing verified matches to OpenMVG.
We use fixed settings for GLOMAP and the default recommended settings for OpenMVG and Theia.

\begin{table}[t]
    \centering
    \caption{Results on ETH3D MVS (DSLR)~\cite{schoeps2017cvpr}. On this dataset, the proposed method outperforms other global SfM baselines and is comparable to COLMAP~\cite{schoenberger2016sfm}.}
    % \vspace{-5px}
    \resizebox{\textwidth}{!}{
    \begin{tabular}{l l c c c a l c c c a l c c c a l c c c d
    } \toprule 
        % && \multicolumn{4}{c}{OpenMVG} && \multicolumn{4}{c}{Theia} && \multicolumn{4}{c}{Ours} 
        % && \multicolumn{4}{c}{COLMAP}
        && \multicolumn{4}{c}{AUC @ $1^\circ$} && \multicolumn{4}{c}{AUC @ $3^\circ$} && \multicolumn{4}{c}{AUC @ $5^\circ$} 
        && \multicolumn{4}{c}{Time (s)}
        \\
        \cmidrule{3-6} \cmidrule{8-11} \cmidrule{13-16} \cmidrule{18-21}
        && \tiny{OpenMVG} & \tiny{ ~~Theia~~} & \tiny{GLOMAP} & \tiny{COLMAP} && \tiny{OpenMVG} & \tiny{ ~~Theia~~} & \tiny{GLOMAP} & \tiny{COLMAP} && \tiny{OpenMVG} & \tiny{ ~~Theia~~} & \tiny{GLOMAP} & \tiny{COLMAP} && \tiny{OpenMVG} & \tiny{ ~~Theia~~} & \tiny{GLOMAP} & \tiny{COLMAP} \\  \midrule
courtyard && 68.2 & \cellcolor{tabfirst}91.2 & \cellcolor{tabsecond}87.8 & 87.3 && 81.7 & \cellcolor{tabfirst}97.0 & \cellcolor{tabsecond}95.9 & 95.8 && 85.9 & \cellcolor{tabfirst}98.2 & \cellcolor{tabsecond}97.6 & 97.5 && \cellcolor{tabfirst}11.2 & \cellcolor{tabsecond}11.9 & 24.5 & 39.2 \\
delivery\_area && 91.4 & 92.0 & \cellcolor{tabfirst}92.8 & \cellcolor{tabsecond}92.3 && 97.1 & 97.3 & \cellcolor{tabfirst}97.6 & \cellcolor{tabsecond}97.4 && 98.3 & 98.4 & \cellcolor{tabfirst}98.6 & \cellcolor{tabsecond}98.5 && 235.7 & \cellcolor{tabfirst}3.8 & \cellcolor{tabsecond}9.2 & 26.4 \\
electro && \cellcolor{tabsecond}72.6 & 49.8 & \cellcolor{tabfirst}79.3 & 70.0 && \cellcolor{tabsecond}81.8 & 54.9 & \cellcolor{tabfirst}86.7 & 75.7 && \cellcolor{tabsecond}83.9 & 57.9 & \cellcolor{tabfirst}88.5 & 77.0 && 99.9 & \cellcolor{tabfirst}2.8 & \cellcolor{tabsecond}7.5 & 23.9 \\
facade && 89.4 & 88.1 & \cellcolor{tabfirst}91.1 & \cellcolor{tabsecond}90.3 && 96.4 & 94.3 & \cellcolor{tabfirst}97.0 & \cellcolor{tabsecond}96.7 && 97.8 & 95.5 & \cellcolor{tabfirst}98.2 & \cellcolor{tabsecond}98.0 && \cellcolor{tabfirst}35.1 & \cellcolor{tabsecond}47.4 & 91.4 & 113.5 \\
kicker && 82.0 & 71.0 & \cellcolor{tabsecond}86.3 & \cellcolor{tabfirst}86.6 && 89.6 & 77.6 & \cellcolor{tabfirst}94.3 & \cellcolor{tabsecond}91.3 && 91.2 & 79.2 & \cellcolor{tabfirst}96.5 & \cellcolor{tabsecond}92.2 && \cellcolor{tabfirst}1.7 & \cellcolor{tabsecond}3.5 & 6.4 & 16.2 \\
meadow && 10.5 & 17.3 & \cellcolor{tabfirst}74.7 & \cellcolor{tabsecond}61.6 && 13.8 & 21.5 & \cellcolor{tabfirst}90.4 & \cellcolor{tabsecond}77.8 && 14.7 & 23.4 & \cellcolor{tabfirst}94.2 & \cellcolor{tabsecond}81.5 && \cellcolor{tabfirst}0.3 & \cellcolor{tabsecond}1.8 & 2.1 & 5.1 \\
office && 19.3 & 27.6 & \cellcolor{tabfirst}59.6 & \cellcolor{tabsecond}45.2 && 24.1 & 32.8 & \cellcolor{tabfirst}81.8 & \cellcolor{tabsecond}56.5 && 25.6 & 35.5 & \cellcolor{tabfirst}88.6 & \cellcolor{tabsecond}59.9 && \cellcolor{tabfirst}0.4 & \cellcolor{tabsecond}0.7 & 1.3 & 19.1 \\
pipes && 29.4 & 41.0 & \cellcolor{tabfirst}89.8 & \cellcolor{tabsecond}86.3 && 36.3 & 53.5 & \cellcolor{tabfirst}96.6 & \cellcolor{tabsecond}95.4 && 41.1 & 56.7 & \cellcolor{tabfirst}98.0 & \cellcolor{tabsecond}97.3 && \cellcolor{tabfirst}0.3 & \cellcolor{tabsecond}0.6 & 0.8 & 3.4 \\
playground && 49.4 & 72.3 & \cellcolor{tabfirst}91.2 & \cellcolor{tabsecond}90.6 && 55.6 & 77.8 & \cellcolor{tabfirst}97.0 & \cellcolor{tabsecond}96.8 && 56.9 & 78.9 & \cellcolor{tabfirst}98.2 & \cellcolor{tabsecond}98.1 && 121.5 & \cellcolor{tabfirst}3.0 & \cellcolor{tabsecond}7.7 & 29.8 \\
relief && 89.8 & 66.9 & \cellcolor{tabfirst}93.7 & \cellcolor{tabsecond}93.3 && 96.6 & 73.0 & \cellcolor{tabfirst}97.9 & \cellcolor{tabsecond}97.8 && \cellcolor{tabsecond}97.9 & 76.4 & \cellcolor{tabfirst}98.7 & \cellcolor{tabfirst}98.7 && \cellcolor{tabfirst}3.7 & \cellcolor{tabsecond}4.7 & 12.7 & 34.1 \\
relief\_2 && 11.9 & 94.1 & \cellcolor{tabfirst}95.0 & \cellcolor{tabsecond}94.9 && 12.4 & 98.0 & \cellcolor{tabfirst}98.3 & \cellcolor{tabsecond}98.2 && 12.5 & 98.8 & \cellcolor{tabfirst}99.0 & \cellcolor{tabsecond}98.9 && \cellcolor{tabfirst}0.7 & \cellcolor{tabsecond}3.2 & 5.4 & 24.1 \\
terrace && 91.9 & \cellcolor{tabfirst}94.0 & \cellcolor{tabsecond}92.8 & 92.3 && 97.3 & \cellcolor{tabfirst}98.0 & \cellcolor{tabsecond}97.6 & 97.4 && 98.4 & \cellcolor{tabfirst}98.8 & \cellcolor{tabsecond}98.6 & 98.5 && \cellcolor{tabfirst}1.1 & \cellcolor{tabsecond}1.6 & 4.0 & 10.9 \\
terrains && 70.9 & \cellcolor{tabfirst}86.4 & \cellcolor{tabsecond}82.2 & 81.9 && 89.3 & \cellcolor{tabfirst}95.3 & \cellcolor{tabsecond}93.7 & 93.6 && 93.4 & \cellcolor{tabfirst}97.1 & \cellcolor{tabsecond}96.2 & 96.1 && 62.6 & \cellcolor{tabfirst}2.1 & \cellcolor{tabsecond}6.8 & 21.7 \\
botanical\_garden && 26.0 & \cellcolor{tabsecond}51.0 & \cellcolor{tabfirst}87.2 & 5.1 && 30.1 & \cellcolor{tabsecond}74.5 & \cellcolor{tabfirst}95.6 & 5.3 && 30.9 & \cellcolor{tabsecond}82.4 & \cellcolor{tabfirst}97.3 & 5.4 && \cellcolor{tabfirst}1.0 & \cellcolor{tabfirst}1.0 & \cellcolor{tabsecond}4.2 & 9.0 \\
boulders && 89.6 & 89.8 & \cellcolor{tabfirst}91.0 & \cellcolor{tabsecond}90.6 && 96.4 & 96.5 & \cellcolor{tabfirst}97.0 & \cellcolor{tabsecond}96.9 && 97.9 & 97.9 & \cellcolor{tabfirst}98.2 & \cellcolor{tabsecond}98.1 && \cellcolor{tabsecond}4.1 & \cellcolor{tabfirst}2.6 & 8.3 & 18.5 \\
bridge && 44.9 & 89.8 & \cellcolor{tabfirst}91.8 & \cellcolor{tabsecond}91.6 && 47.9 & \cellcolor{tabsecond}95.3 & \cellcolor{tabfirst}97.2 & \cellcolor{tabfirst}97.2 && 48.5 & \cellcolor{tabsecond}96.4 & \cellcolor{tabfirst}98.3 & \cellcolor{tabfirst}98.3 && \cellcolor{tabsecond}35.1 & \cellcolor{tabfirst}31.3 & 87.2 & 91.9 \\
door && 93.8 & 93.8 & \cellcolor{tabsecond}95.1 & \cellcolor{tabfirst}96.9 && 97.9 & 97.9 & \cellcolor{tabsecond}98.4 & \cellcolor{tabfirst}99.0 && 98.8 & 98.8 & \cellcolor{tabsecond}99.0 & \cellcolor{tabfirst}99.4 && \cellcolor{tabsecond}1.2 & \cellcolor{tabfirst}1.0 & 1.8 & 4.2 \\
exhibition\_hall && 11.0 & \cellcolor{tabsecond}84.4 & 25.5 & \cellcolor{tabfirst}85.4 && 15.2 & \cellcolor{tabsecond}93.5 & 29.7 & \cellcolor{tabfirst}94.3 && 16.3 & \cellcolor{tabsecond}96.0 & 30.7 & \cellcolor{tabfirst}96.5 && \cellcolor{tabfirst}20.6 & \cellcolor{tabsecond}49.8 & 68.3 & 72.0 \\
lecture\_room && 80.2 & 69.4 & \cellcolor{tabfirst}83.3 & \cellcolor{tabsecond}82.9 && \cellcolor{tabsecond}92.7 & 79.6 & \cellcolor{tabfirst}94.1 & \cellcolor{tabfirst}94.1 && 95.6 & 82.6 & \cellcolor{tabfirst}96.5 & \cellcolor{tabsecond}96.4 && \cellcolor{tabsecond}3.3 & \cellcolor{tabfirst}2.4 & 7.4 & 10.9 \\
living\_room && \cellcolor{tabsecond}88.0 & 84.8 & \cellcolor{tabsecond}88.0 & \cellcolor{tabfirst}88.3 && \cellcolor{tabfirst}95.8 & \cellcolor{tabsecond}92.6 & \cellcolor{tabfirst}95.8 & \cellcolor{tabfirst}95.8 && \cellcolor{tabsecond}97.4 & 94.3 & \cellcolor{tabsecond}97.4 & \cellcolor{tabfirst}97.5 && \cellcolor{tabfirst}9.4 & \cellcolor{tabsecond}10.5 & 22.5 & 49.1 \\
lounge && \cellcolor{tabfirst}34.1 & \cellcolor{tabfirst}34.1 & \cellcolor{tabsecond}34.0 & 33.9 && \cellcolor{tabfirst}35.4 & \cellcolor{tabfirst}35.4 & \cellcolor{tabsecond}35.3 & \cellcolor{tabsecond}35.3 && \cellcolor{tabfirst}35.6 & \cellcolor{tabfirst}35.6 & \cellcolor{tabfirst}35.6 & \cellcolor{tabfirst}35.6 && \cellcolor{tabfirst}0.4 & \cellcolor{tabsecond}1.0 & 1.2 & 1.8 \\
observatory && 58.4 & \cellcolor{tabfirst}65.8 & \cellcolor{tabsecond}65.4 & 64.4 && 83.7 & \cellcolor{tabfirst}87.1 & \cellcolor{tabsecond}87.0 & 86.5 && 89.9 & \cellcolor{tabfirst}92.2 & \cellcolor{tabsecond}92.1 & 91.8 && \cellcolor{tabsecond}5.8 & \cellcolor{tabfirst}4.8 & 13.2 & 24.4 \\
old\_computer && 23.3 & 49.8 & \cellcolor{tabsecond}53.8 & \cellcolor{tabfirst}78.7 && 41.5 & 59.8 & \cellcolor{tabsecond}61.4 & \cellcolor{tabfirst}90.2 && 48.4 & 62.0 & \cellcolor{tabsecond}63.0 & \cellcolor{tabfirst}92.7 && \cellcolor{tabsecond}5.9 & \cellcolor{tabfirst}4.1 & 8.9 & 19.6 \\
statue && 96.4 & \cellcolor{tabfirst}98.8 & \cellcolor{tabfirst}98.8 & \cellcolor{tabsecond}98.7 && \cellcolor{tabsecond}98.8 & \cellcolor{tabfirst}99.6 & \cellcolor{tabfirst}99.6 & \cellcolor{tabfirst}99.6 && 99.3 & \cellcolor{tabfirst}99.8 & \cellcolor{tabfirst}99.8 & \cellcolor{tabsecond}99.7 && \cellcolor{tabfirst}0.9 & \cellcolor{tabsecond}1.6 & 5.7 & 7.4 \\
terrace\_2 && 87.9 & \cellcolor{tabsecond}90.8 & \cellcolor{tabfirst}91.0 & 90.7 && 95.7 & \cellcolor{tabfirst}96.9 & \cellcolor{tabfirst}96.9 & \cellcolor{tabsecond}96.8 && 97.4 & \cellcolor{tabsecond}98.1 & \cellcolor{tabfirst}98.2 & \cellcolor{tabsecond}98.1 && \cellcolor{tabfirst}1.0 & \cellcolor{tabsecond}1.5 & 3.3 & 10.2 \\
\midrule
\textit{Average} && 60.4 & 71.8 & \cellcolor{tabfirst}80.8 & \cellcolor{tabsecond}79.2 && 68.1 & 79.2 & \cellcolor{tabfirst}88.5 & \cellcolor{tabsecond}86.5 && 70.1 & 81.2 & \cellcolor{tabfirst}90.3 & \cellcolor{tabsecond}88.1 && 26.5 & \cellcolor{tabfirst}7.9 & \cellcolor{tabsecond}16.5 & 27.5 \\
      \bottomrule
    \end{tabular}
    }
    \label{tbl:eth3d_drls}
    % \vspace{-15px}
\end{table}

\subsection{Calibrated Image Collections}

% To test our method with known camera intrinsics, we use the ETH3D SLAM~\cite{Schops_2019_CVPR} and MVS~\cite{schoeps2017cvpr} datasets as well as LaMAR~\cite{sarlin2022lamar}.

% \vspace{0mm}
\noindent{\textbf{ETH3D SLAM}\cite{Schops_2019_CVPR}} is a challenging dataset containing sequential data with sparse features, dynamic objects, and drastic illumination changes. 
We evaluated our method on the training sequences that come with millimeter-accurate ground truth.
Ground truth is not available for test sequences and some frames, so we do not consider them.
The results are presented in Table~\ref{tbl:eth3d_slam}.
% , highlighting the recall at $0.1$m, the AUC score for camera position at $0.1$m, and the $0.5$m. 
% As ground truth is not available at every timestamp, we only report results on frames with known ground truth.
% The comparison only includes timestamps for which ground truth is available. 
Each row in the table averages the results across sequences sharing the same prefix with full results in the suppl.~material.
The results demonstrate that our proposed GLOMAP system achieves approximately 8\% higher recall and scores 9 and 8 additional points in AUC at the 0.1m and 0.5m thresholds, respectively, compared to COLMAP, which is also one order of magnitude slower. 
Against other global SfM pipelines, GLOMAP shows a 18\% and 4\% improvement in recall and around 11 points higher AUC at 0.1m, confirming its robustness.

% \vspace{0mm}
\noindent{\textbf{ETH3D MVS (rig)} \cite{schoeps2017cvpr}}
% contains 5 scenes where each scene contains about 1000 images taken by 4 cameras.
contains, per scene, about 1000 multi-rig exposures with each 4 images.
The dataset contains both outdoor and indoor scenes with millimeter-accurate ground truth for 5 training sequences.
% Each timestamp has 4 images acquired by a rig with fixed relative poses.
We do not fix the pose of the rig for any of the methods.
Results on this dataset can be found in Table~\ref{tbl:eth3d_rig}.
Ours successfully reconstructs all scenes.
% Ours successfully reconstructs all scenes, with recall at 0.1m being 99\% or higher.
In contrast, OpenMVG performs poorly on all scenes while COLMAP fails for one, and Theia performs consistently worse than ours.
On the sequences where COLMAP succeeds, ours achieves similar or higher accuracy.
% While Theia succeeds runs on all sequences, the proposed method obtains about 4.5 percent higher recall at 0.1m, and about 20 more points for AUC at 0.1m.
Our runtime is a little slower than global SfM baselines and about 3.5 times faster than COLMAP.

% Theia achieve relative good performance, y

% \begin{itemize}
%     \item Low resolution image
%     \item sequence taken by a rig
% \end{itemize}

% \begin{itemize}
%     \item High resolution images
%     \item unordered map with low overlapping field of view
% \end{itemize}

% \vspace{0mm}
\noindent{\textbf{ETH3D MVS (DSLR)} \cite{schoeps2017cvpr}}
features an unordered collection of high-resolution images of outdoor and indoor scenes with millimeter-accurate ground truth for both training and testing sequences and results reported in Table~\ref{tbl:eth3d_drls}.
Consistent with other ETH3D datasets, ours outperforms OpenMVG and Theia while achieving similar accuracy as COLMAP.\
For \textit{exihibition\_hall}, GLOMAP performs inaccurately because of rotational symmetry of the scene, causing rotation averaging to collapse.
Due to the small scale of the scenes, all methods achieve comparable runtimes.

\begin{table}[t]
    \centering
    \caption{Results on LaMAR~\cite{sarlin2022lamar} datasets. The proposed method largely outperforms other baselines as well as COLMAP~\cite{schoenberger2016sfm}. For LIN, structure refinement is not performed for GLOMAP due to memory limitation (marked as $^*$).}
    % \vspace{-5px}
    \resizebox{\textwidth}{!}{
    \begin{tabular}{l l c c c a l c c c a l c c c a l c c c d
    } \toprule 
        % && \multicolumn{4}{c}{OpenMVG} && \multicolumn{4}{c}{Theia} && \multicolumn{4}{c}{Ours} 
        % && \multicolumn{4}{c}{COLMAP}
        % && \multicolumn{3}{c}{Thres 5m} && \multicolumn{3}{c}{AUC @ 5} && \multicolumn{3}{c}{AUC @ 10} 
        && \multicolumn{4}{c}{Recall @ 1m} && \multicolumn{4}{c}{AUC @ 1m} && \multicolumn{4}{c}{AUC @ 5m} 
        && \multicolumn{4}{c}{Time (s)}
        \\
        \cmidrule{3-6} \cmidrule{8-11} \cmidrule{13-16} \cmidrule{18-21}
        && \tiny{OpenMVG} & \tiny{~~Theia~~} & \tiny{GLOMAP} & \tiny{COLMAP} &&  \tiny{OpenMVG} & \tiny{~~Theia~~} & \tiny{GLOMAP} & \tiny{COLMAP} && \tiny{OpenMVG} & \tiny{~~Theia~~} & \tiny{GLOMAP} & \tiny{COLMAP} && \tiny{OpenMVG} & \tiny{~~Theia~~} & \tiny{GLOMAP} & \tiny{COLMAP} \\  \midrule
% CAB && - & 3.2 & \cellcolor{tabsecond}11.5 & \cellcolor{tabfirst}13.0 && - & 1.3 & \cellcolor{tabfirst}5.9 & \cellcolor{tabsecond}5.8 && - & 9.2 & \cellcolor{tabsecond}16.5 & \cellcolor{tabfirst}19.2 && - & \cellcolor{tabfirst}1953.5 & \cellcolor{tabsecond}3696.4 & 194033.6 \\
% HGE && - & 3.4 & \cellcolor{tabsecond}33.9 & \cellcolor{tabfirst}38.9 && - & 1.3 & \cellcolor{tabfirst}18.6 & \cellcolor{tabsecond}18.0 && - & 9.6 & \cellcolor{tabsecond}37.0 & \cellcolor{tabfirst}46.9 && - & \cellcolor{tabfirst}1829.7 & \cellcolor{tabsecond}3974.0 & 249771.1 \\
% LIN && - & 8.6 & \cellcolor{tabfirst}83.4 & \cellcolor{tabsecond}44.2 && - & 3.3 & \cellcolor{tabfirst}38.7 & \cellcolor{tabsecond}17.7 && - & 24.1 & \cellcolor{tabfirst}83.5 & \cellcolor{tabsecond}52.3 && - & \cellcolor{tabfirst}2783.3 & \cellcolor{tabsecond}6994.4 & 620176.4 \\
% \midrule
% \textit{Average} && - & 5.1 & \cellcolor{tabfirst}42.9 & \cellcolor{tabsecond}32.0 && - & 2.0 & \cellcolor{tabfirst}21.1 & \cellcolor{tabsecond}13.8 && - & 14.3 & \cellcolor{tabfirst}45.6 & \cellcolor{tabsecond}39.4 && - & \cellcolor{tabfirst}2188.8 & \cellcolor{tabsecond}4888.3 & 354660.4 \\
CAB && - & 6.0 & \cellcolor{tabsecond}11.6 & \cellcolor{tabfirst}13.0 && - & 3.2 & \cellcolor{tabsecond}4.7 & \cellcolor{tabfirst}5.8 && - & 9.3 & \cellcolor{tabsecond}16.9 & \cellcolor{tabfirst}19.2 && - & \cellcolor{tabfirst}1345.6 & \cellcolor{tabsecond}6162.2 & 194033.6 \\
HGE && - & 8.3 & \cellcolor{tabfirst}48.4 & \cellcolor{tabsecond}38.9 && - & 2.9 & \cellcolor{tabfirst}22.2 & \cellcolor{tabsecond}18.0 && - & 9.4 & \cellcolor{tabfirst}50.3 & \cellcolor{tabsecond}46.9 && - & \cellcolor{tabfirst}1182.4 & \cellcolor{tabsecond}12587.2 & 249771.1 \\
LIN$^*$ && - & 18.8 & \cellcolor{tabfirst}87.3 & \cellcolor{tabsecond}44.2 && - & 7.0 & \cellcolor{tabfirst}46.7 & \cellcolor{tabsecond}17.7 && - & 38.5 & \cellcolor{tabfirst}85.6 & \cellcolor{tabsecond}52.3 && - & \cellcolor{tabfirst}2097.9 & \cellcolor{tabsecond}18466.6 & 620176.4 \\
\midrule
\textit{Average} && - & 11.0 & \cellcolor{tabfirst}49.1 & \cellcolor{tabsecond}32.0 && - & 4.4 & \cellcolor{tabfirst}24.5 & \cellcolor{tabsecond}13.8 && - & 19.1 & \cellcolor{tabfirst}50.9 & \cellcolor{tabsecond}39.4 && - & \cellcolor{tabfirst}1542.0 & \cellcolor{tabsecond}12405.3 & 354660.4 \\
      \bottomrule
    \end{tabular}
    }
    \label{tbl:lamar}
    % \vspace{-10px}
\end{table}
% \vspace{0mm}
\noindent{\textbf{LaMAR}~\cite{sarlin2022lamar}} is a large-scale indoor and outdoor benchmark with each scene containing several tens of thousands of images captured by a variety of AR devices and smartphones.
% We test all methods on the largest scene \textit{LIN} which contains 37677 images. \todo{HGE/CAB}
For this dataset, we use the retrieval pipeline from the benchmark~\cite{sarlin2022lamar} to establish matches.
% The mat
The results on this dataset can be found in Table.~\ref{tbl:lamar}, and the qualitative result can be found in Figure~\ref{fig:teaser_lamar}.
% Except for the \textit{LIN} scene, all methods including COLMAP perform poorly, especially upon visual inspection, on this extremely challenging benchmark due to many forward motion trajectories, drastic day-night illumination changes, and many symmetries across floors/rooms and repetitive facades.
% OpenMVG~\cite{moulon2016openmvg} fails completely, Theia~\cite{theia-manual} performs a little better, while GLOMAP has the smallest gap to COLMAP and is two orders of magnitude faster.
GLOMAP achieves significantly more accurate reconstruction on \textit{HGE} and \textit{LIN} compared to all other baselines, including COLMAP~\cite{schoenberger2016sfm} while being orders of magnitude faster than COLMAP.
On \textit{CAB}, all methods, including COLMAP, perform poorly, especially upon visual inspection, on this extremely challenging benchmark due to many forward motion trajectories, drastic day-night illumination changes, and many symmetries across floors/rooms and repetitive facades.

\subsection{Uncalibrated Images Collections}

% To test the performance on uncalibrated cameras, we test on the training set of Image Matching Challenge 2023 (IMC)~\cite{IMC2023} and datasets from MIP360~\cite{barron2022mip}.
% Images in these two datasets are not calibrated

% \begin{itemize}
%     \item Unordered set of images
%     \item Images with varying intrinsics and resolution
%     \item Ground truth from a reconstruction with more images built with COLMAP
%     \item slightly biased towards COLMAP results
% \end{itemize}

% \vspace{0mm}
\noindent{\textbf{IMC 2023} \cite{IMC2023}}
contains unordered image collections over complex scenes.
Images are collected from various sources and often lack prior camera intrinsics.
The ground truth of the dataset is built by COLMAP~\cite{schoenberger2016sfm} with held out imagery.
% thus exhibiting a bias towards performing well for COLMAP versus other methods.
% contains 3 categories of images, \textit{haiper}, \textit{urban}, \textit{phototourism},
As the accuracy of this dataset is not very high, we follow the same scheme as He~\etal~\cite{he2023detector} to report the AUC scores at $3^\circ, 5^\circ, 10^\circ$.
The results on training sets can be found in Table~\ref{tbl:imc}.
% On this dataset, the average AUC scores of the proposed method at $3^\circ$ and $5^\circ$ is 1.6 -  3.4 times larger than other global SfM baselines, and the proposed method obtains 25 - 46 points higher in AUC score at $5^\circ$.
On this dataset, the average AUC scores of the proposed method at $3^\circ$, $5^\circ$ and $10^\circ$ is several times higher than other global SfM baselines.
The runtime is similar to other global SfM pipelines.
% Compared to COLMAP~\cite{schoenberger2016sfm}, the proposed method is about 10, 7, and 4 points lower in AUC scores at $3^\circ, 5^\circ$, and $10^\circ$ respectively, but is about 16 times faster.
Compared to COLMAP~\cite{schoenberger2016sfm}, the proposed method is about 4 points higher in AUC scores at $3^\circ, 5^\circ$, and $10^\circ$, and is about 8 times faster.

\begin{table}[t]
    \centering
    \caption{Results on IMC 2023~\cite{IMC2023}. Our GLOMAP method comes close to COLMAP generated ground truth while outperforming global SfM by a large margin.}
    % \vspace{-5px}
    \resizebox{\textwidth}{!}{
    \begin{tabular}{l l c c c a l c c c a l c c c a l c c c d
    } \toprule 
        && \multicolumn{4}{c}{AUC @ $3^\circ$} && \multicolumn{4}{c}{AUC @ $5^\circ$} && \multicolumn{4}{c}{AUC @ $10^\circ$} 
        && \multicolumn{4}{c}{Time (s)}
        \\
        \cmidrule{3-6} \cmidrule{8-11} \cmidrule{13-16} \cmidrule{18-21}
        && \tiny{OpenMVG} & \tiny{~~~Theia~~~} & \tiny{GLOMAP} & \tiny{COLMAP} && \tiny{OpenMVG} & \tiny{~~~Theia~~~} & \tiny{GLOMAP} & \tiny{COLMAP} && \tiny{OpenMVG} & \tiny{~~~Theia~~~} & \tiny{GLOMAP} & \tiny{COLMAP} && \tiny{OpenMVG} & \tiny{~~~Theia~~~} & \tiny{GLOMAP} & \tiny{COLMAP} \\  \midrule
bike && - & 0.0 & \cellcolor{tabsecond}35.0 & \cellcolor{tabfirst}77.9 && - & 0.0 & \cellcolor{tabsecond}38.9 & \cellcolor{tabfirst}86.7 && - & 0.0 & \cellcolor{tabsecond}41.9 & \cellcolor{tabfirst}93.4 && - & \cellcolor{tabsecond}1.4 & 1.5 & \cellcolor{tabfirst}1.1 \\
chairs && - & 0.0 & \cellcolor{tabfirst}82.6 & \cellcolor{tabsecond}0.8 && - & 0.0 & \cellcolor{tabfirst}89.6 & \cellcolor{tabsecond}0.8 && - & 0.0 & \cellcolor{tabfirst}94.8 & \cellcolor{tabsecond}0.8 && - & 1.7 & \cellcolor{tabsecond}1.4 & \cellcolor{tabfirst}0.6 \\
fountain && 22.1 & 57.1 & \cellcolor{tabsecond}91.2 & \cellcolor{tabfirst}91.3 && 24.2 & 61.6 & \cellcolor{tabsecond}94.7 & \cellcolor{tabfirst}94.8 && 25.7 & \cellcolor{tabsecond}64.9 & \cellcolor{tabfirst}97.4 & \cellcolor{tabfirst}97.4 && \cellcolor{tabfirst}0.8 & \cellcolor{tabsecond}1.4 & 3.4 & 5.9 \\
cyprus && 1.6 & 17.3 & \cellcolor{tabfirst}67.1 & \cellcolor{tabsecond}45.7 && 1.7 & 21.8 & \cellcolor{tabfirst}73.8 & \cellcolor{tabsecond}48.9 && 1.7 & 28.0 & \cellcolor{tabfirst}80.5 & \cellcolor{tabsecond}51.4 && \cellcolor{tabfirst}0.3 & \cellcolor{tabsecond}1.3 & 2.6 & 11.1 \\
dioscuri && 0.4 & 1.7 & \cellcolor{tabfirst}59.4 & \cellcolor{tabsecond}58.7 && 0.4 & 2.5 & \cellcolor{tabfirst}61.9 & \cellcolor{tabsecond}61.4 && 0.5 & 4.5 & \cellcolor{tabfirst}64.4 & \cellcolor{tabsecond}63.9 && \cellcolor{tabsecond}41.9 & \cellcolor{tabfirst}24.7 & 115.6 & 156.3 \\
wall && 57.1 & 84.6 & \cellcolor{tabfirst}95.3 & \cellcolor{tabsecond}88.6 && 73.9 & 88.9 & \cellcolor{tabfirst}97.2 & \cellcolor{tabsecond}93.2 && 87.0 & 92.1 & \cellcolor{tabfirst}98.6 & \cellcolor{tabsecond}96.6 && \cellcolor{tabfirst}23.8 & \cellcolor{tabsecond}28.1 & 77.8 & 63.5 \\
kyiv-puppet-theater && 0.7 & \cellcolor{tabsecond}1.0 & \cellcolor{tabfirst}10.0 & 0.3 && 0.7 & \cellcolor{tabsecond}1.1 & \cellcolor{tabfirst}12.0 & 0.3 && 0.9 & \cellcolor{tabsecond}1.5 & \cellcolor{tabfirst}19.5 & 0.3 && \cellcolor{tabfirst}0.3 & \cellcolor{tabsecond}2.5 & 2.6 & \cellcolor{tabfirst}0.3 \\
brandenburg\_gate && 21.4 & 42.3 & \cellcolor{tabsecond}68.7 & \cellcolor{tabfirst}70.1 && 35.0 & 52.8 & \cellcolor{tabsecond}75.2 & \cellcolor{tabfirst}77.2 && 53.6 & 65.2 & \cellcolor{tabsecond}81.5 & \cellcolor{tabfirst}83.9 && 1171.8 & \cellcolor{tabfirst}199.9 & \cellcolor{tabsecond}368.4 & 1472.1 \\
british\_museum && 18.5 & 34.9 & \cellcolor{tabfirst}62.0 & \cellcolor{tabsecond}61.6 && 32.0 & \cellcolor{tabsecond}47.4 & \cellcolor{tabfirst}72.7 & \cellcolor{tabfirst}72.7 && 51.3 & 63.8 & \cellcolor{tabsecond}83.5 & \cellcolor{tabfirst}83.9 && \cellcolor{tabfirst}78.1 & \cellcolor{tabsecond}84.6 & 117.6 & 318.0 \\
buckingham\_palace && 4.1 & 26.0 & \cellcolor{tabfirst}85.9 & \cellcolor{tabsecond}80.5 && 12.5 & 37.2 & \cellcolor{tabfirst}89.1 & \cellcolor{tabsecond}86.2 && 34.0 & 53.0 & \cellcolor{tabfirst}92.1 & \cellcolor{tabsecond}91.1 && \cellcolor{tabsecond}429.7 & \cellcolor{tabfirst}173.3 & 484.1 & 4948.2 \\
colosseum\_exterior && 37.3 & 69.0 & \cellcolor{tabsecond}80.5 & \cellcolor{tabfirst}80.7 && 52.8 & 77.1 & \cellcolor{tabsecond}85.8 & \cellcolor{tabfirst}86.1 && 69.2 & 84.7 & \cellcolor{tabsecond}90.5 & \cellcolor{tabfirst}90.8 && \cellcolor{tabsecond}542.8 & \cellcolor{tabfirst}489.2 & 767.9 & 3561.7 \\
grand\_place\_brussels && 18.3 & 34.7 & \cellcolor{tabfirst}71.8 & \cellcolor{tabsecond}68.7 && 32.4 & 50.8 & \cellcolor{tabfirst}78.5 & \cellcolor{tabsecond}76.6 && 50.5 & 67.6 & \cellcolor{tabfirst}84.6 & \cellcolor{tabsecond}84.2 && \cellcolor{tabsecond}124.2 & \cellcolor{tabfirst}87.6 & 220.7 & 520.3 \\
lincoln\_..statue && 1.0 & 30.7 & \cellcolor{tabfirst}68.4 & \cellcolor{tabsecond}67.2 && 4.4 & 36.3 & \cellcolor{tabfirst}72.5 & \cellcolor{tabsecond}71.5 && 23.8 & 41.7 & \cellcolor{tabfirst}76.0 & \cellcolor{tabsecond}75.3 && \cellcolor{tabsecond}99.9 & \cellcolor{tabfirst}46.1 & 103.9 & 362.3 \\
notre\_dame\_..facade && 32.5 & 52.3 & \cellcolor{tabsecond}67.2 & \cellcolor{tabfirst}69.1 && 43.0 & 59.1 & \cellcolor{tabsecond}70.8 & \cellcolor{tabfirst}72.5 && 55.2 & 65.5 & \cellcolor{tabsecond}74.3 & \cellcolor{tabfirst}75.5 && \cellcolor{tabsecond}2592.1 & \cellcolor{tabfirst}2393.9 & 4146.9 & 52135.4 \\
pantheon\_exterior && 49.6 & 68.3 & \cellcolor{tabsecond}77.0 & \cellcolor{tabfirst}79.4 && 62.3 & 74.9 & \cellcolor{tabsecond}81.8 & \cellcolor{tabfirst}83.5 && 74.4 & 81.2 & \cellcolor{tabsecond}86.2 & \cellcolor{tabfirst}87.2 && \cellcolor{tabsecond}285.7 & \cellcolor{tabfirst}169.3 & 488.8 & 1454.6 \\
piazza\_san\_marco && 2.6 & 48.6 & \cellcolor{tabfirst}72.7 & \cellcolor{tabsecond}58.0 && 6.4 & 62.0 & \cellcolor{tabfirst}82.3 & \cellcolor{tabsecond}71.0 && 23.7 & 75.9 & \cellcolor{tabfirst}90.6 & \cellcolor{tabsecond}83.7 && \cellcolor{tabsecond}16.7 & \cellcolor{tabfirst}12.6 & 43.4 & 74.2 \\
sacre\_coeur && 37.1 & 73.7 & \cellcolor{tabfirst}78.8 & \cellcolor{tabsecond}78.5 && 50.8 & 77.5 & \cellcolor{tabfirst}81.1 & \cellcolor{tabsecond}80.9 && 64.8 & 80.9 & \cellcolor{tabfirst}83.0 & \cellcolor{tabsecond}82.8 && \cellcolor{tabsecond}196.1 & \cellcolor{tabfirst}130.4 & 266.9 & 682.2 \\
sagrada\_familia && 30.3 & 50.7 & \cellcolor{tabsecond}53.8 & \cellcolor{tabfirst}54.3 && 39.2 & 55.5 & \cellcolor{tabsecond}58.7 & \cellcolor{tabfirst}58.9 && 48.9 & \cellcolor{tabsecond}60.0 & \cellcolor{tabfirst}62.9 & \cellcolor{tabfirst}62.9 && \cellcolor{tabfirst}42.6 & \cellcolor{tabsecond}48.6 & 140.9 & 237.0 \\
st\_pauls\_cathedral && 1.9 & 60.5 & \cellcolor{tabfirst}74.2 & \cellcolor{tabsecond}72.7 && 6.5 & 70.4 & \cellcolor{tabfirst}80.2 & \cellcolor{tabsecond}78.9 && 25.8 & 79.7 & \cellcolor{tabfirst}85.8 & \cellcolor{tabsecond}85.0 && \cellcolor{tabsecond}61.9 & \cellcolor{tabfirst}45.0 & 101.0 & 241.4 \\
st\_peters\_square && 31.1 & 56.3 & \cellcolor{tabsecond}79.5 & \cellcolor{tabfirst}83.6 && 46.0 & 66.8 & \cellcolor{tabsecond}84.2 & \cellcolor{tabfirst}87.8 && 64.0 & 77.5 & \cellcolor{tabsecond}88.8 & \cellcolor{tabfirst}91.6 && \cellcolor{tabsecond}961.0 & \cellcolor{tabfirst}621.1 & 1177.9 & 6051.9 \\
taj\_mahal && 38.6 & 58.6 & \cellcolor{tabfirst}72.1 & \cellcolor{tabsecond}68.9 && 51.0 & 65.7 & \cellcolor{tabfirst}77.3 & \cellcolor{tabsecond}75.5 && 64.9 & 73.3 & \cellcolor{tabfirst}82.4 & \cellcolor{tabsecond}81.7 && \cellcolor{tabsecond}380.1 & \cellcolor{tabfirst}379.0 & 630.7 & 4528.9 \\
trevi\_fountain && 43.4 & 64.6 & \cellcolor{tabsecond}79.0 & \cellcolor{tabfirst}80.5 && 56.2 & 72.7 & \cellcolor{tabsecond}82.8 & \cellcolor{tabfirst}84.4 && 69.3 & 80.3 & \cellcolor{tabsecond}86.4 & \cellcolor{tabfirst}87.8 && \cellcolor{tabsecond}1202.9 & \cellcolor{tabfirst}669.6 & 1676.0 & 12294.4 \\
\midrule
\textit{Average} && 20.4 & 42.4 & \cellcolor{tabfirst}69.6 & \cellcolor{tabsecond}65.3 && 28.7 & 49.2 & \cellcolor{tabfirst}74.6 & \cellcolor{tabsecond}70.4 && 40.4 & 56.4 & \cellcolor{tabfirst}79.4 & \cellcolor{tabsecond}75.1 && \cellcolor{tabsecond}412.6 & \cellcolor{tabfirst}255.1 & 497.3 & 4051.0 \\
\bottomrule 
    \end{tabular}
    }
    \label{tbl:imc}
\end{table}

% \vspace{0mm}
\noindent{\textbf{MIP360}\cite{barron2022mip}}
contains 7 object-centric scenes with high-resolution images taken by the same camera.
The provided COLMAP model is considered as (pseudo) ground truth for this dataset. 
Similarly, as the accuracy of ground truth is limited, the AUC scores at $3^\circ, 5^\circ$, and $10^\circ$ are reported.
COLMAP reconstructions are re-estimated with the same matches as other methods.
Results are summarized in Table~\ref{tbl:mip360} and our method is significantly closer to the reference model compared with other global SfM methods while rerunning COLMAP produces similar results as ours.
% The difference in the reconstruction of COLMAP in our experiments and ground truth comes from the difference in the matches.
% We obtainting different 
Ours is more than 1.5 times faster than COLMAP.

\begin{table}[t]
    \centering
    \caption{Results on MIP360~\cite{barron2022mip} datasets. The proposed method largely outperforms other baselines while obtaining similar results as COLMAP~\cite{schoenberger2016sfm}.}
    % \vspace{-5px}
    \resizebox{\textwidth}{!}{
    \begin{tabular}{l l c c c a l c c c a l c c c a l c c c d
    } \toprule 
        % && \multicolumn{4}{c}{OpenMVG} && \multicolumn{4}{c}{Theia} && \multicolumn{4}{c}{Ours} 
        % && \multicolumn{4}{c}{COLMAP}
        && \multicolumn{4}{c}{AUC @ $3^\circ$} && \multicolumn{4}{c}{AUC @ $5^\circ$} && \multicolumn{4}{c}{AUC @ $10^\circ$} 
        && \multicolumn{4}{c}{Time (s)}
        \\
        \cmidrule{3-6} \cmidrule{8-11} \cmidrule{13-16} \cmidrule{18-21}
        && \tiny{OpenMVG} & \tiny{~~~Theia~~~} & \tiny{GLOMAP} & \tiny{COLMAP} && \tiny{OpenMVG} & \tiny{~~~Theia~~~} & \tiny{GLOMAP} & \tiny{COLMAP} && \tiny{OpenMVG} & \tiny{~~~Theia~~~} & \tiny{GLOMAP} & \tiny{COLMAP} && \tiny{OpenMVG} & \tiny{~~~Theia~~~} & \tiny{GLOMAP} & \tiny{COLMAP} \\  \midrule
bicycle && \cellcolor{tabsecond}89.2 & 12.0 & \cellcolor{tabfirst}95.8 & \cellcolor{tabfirst}95.8 && \cellcolor{tabsecond}92.7 & 14.7 & \cellcolor{tabfirst}97.5 & \cellcolor{tabfirst}97.5 && \cellcolor{tabsecond}95.3 & 17.9 & \cellcolor{tabfirst}98.7 & \cellcolor{tabfirst}98.7 && 123.1 & \cellcolor{tabfirst}28.6 & \cellcolor{tabsecond}66.9 & 120.6 \\
bonsai && 9.4 & 87.0 & \cellcolor{tabfirst}98.5 & \cellcolor{tabsecond}92.6 && 27.9 & 91.8 & \cellcolor{tabfirst}99.1 & \cellcolor{tabsecond}95.6 && 61.2 & 95.6 & \cellcolor{tabfirst}99.5 & \cellcolor{tabsecond}97.8 && \cellcolor{tabfirst}176.0 & \cellcolor{tabsecond}194.5 & 467.5 & 662.5 \\
counter && 96.9 & 98.9 & \cellcolor{tabfirst}99.3 & \cellcolor{tabsecond}99.2 && 98.1 & 99.4 & \cellcolor{tabfirst}99.6 & \cellcolor{tabsecond}99.5 && 99.1 & \cellcolor{tabsecond}99.7 & \cellcolor{tabfirst}99.8 & \cellcolor{tabfirst}99.8 && \cellcolor{tabfirst}46.6 & \cellcolor{tabsecond}71.0 & 203.9 & 270.5 \\
garden && \cellcolor{tabsecond}95.0 & 36.8 & \cellcolor{tabfirst}97.3 & \cellcolor{tabfirst}97.3 && \cellcolor{tabsecond}97.0 & 37.9 & \cellcolor{tabfirst}98.4 & \cellcolor{tabfirst}98.4 && \cellcolor{tabsecond}98.5 & 38.9 & \cellcolor{tabfirst}99.2 & \cellcolor{tabfirst}99.2 && \cellcolor{tabsecond}40.2 & \cellcolor{tabfirst}39.6 & 128.9 & 291.8 \\
kitchen && 88.5 & 93.5 & \cellcolor{tabsecond}94.8 & \cellcolor{tabfirst}94.9 && 93.1 & 95.8 & \cellcolor{tabsecond}96.9 & \cellcolor{tabfirst}97.0 && 96.5 & 97.6 & \cellcolor{tabsecond}98.4 & \cellcolor{tabfirst}98.5 && \cellcolor{tabfirst}127.0 & \cellcolor{tabsecond}187.3 & 426.9 & 619.1 \\
room && 39.6 & 26.0 & \cellcolor{tabfirst}97.7 & \cellcolor{tabsecond}96.2 && 42.2 & 26.8 & \cellcolor{tabfirst}98.6 & \cellcolor{tabsecond}97.7 && 44.9 & 27.6 & \cellcolor{tabfirst}99.3 & \cellcolor{tabsecond}98.9 && \cellcolor{tabsecond}96.3 & \cellcolor{tabfirst}85.6 & 216.3 & 371.6 \\
stump && \cellcolor{tabsecond}95.3 & 7.1 & \cellcolor{tabfirst}99.1 & \cellcolor{tabfirst}99.1 && \cellcolor{tabsecond}97.1 & 7.5 & \cellcolor{tabfirst}99.5 & \cellcolor{tabfirst}99.5 && \cellcolor{tabsecond}98.5 & 8.0 & \cellcolor{tabfirst}99.7 & \cellcolor{tabfirst}99.7 && 38.7 & \cellcolor{tabfirst}10.9 & \cellcolor{tabsecond}36.6 & 83.9 \\
\midrule
\textit{Average} && 73.4 & 51.6 & \cellcolor{tabfirst}97.5 & \cellcolor{tabsecond}96.5 && 78.3 & 53.4 & \cellcolor{tabfirst}98.5 & \cellcolor{tabsecond}97.9 && 84.9 & 55.0 & \cellcolor{tabfirst}99.2 & \cellcolor{tabsecond}98.9 && \cellcolor{tabsecond}92.6 & \cellcolor{tabfirst}88.2 & 221.0 & 345.7 \\
      \bottomrule
    \end{tabular}
    }
    \label{tbl:mip360}
    % \vspace{-5px}
\end{table}

% Within a 
% \begin{itemize}
%     \item Images with shared intrinsics
%     \item
% \end{itemize}

% \vspace{-10px}
\subsection{Ablation\label{sec:ablation_pt}}

% In this section, we designed two experiments to demonstrate the effectiveness and robustness of the global positioning.
% \vspace{0mm}
% \noindent\textbf{Global Positioning Constraints.\label{sec:ablation_pt}}
To demonstrate the effectiveness of the global position strategy, we conduct experiments by replacing the component by 1) adding only relative translation constraints, denoted as (BATA, cam), and 2) adding both points as well as translation constraints (BATA, cam+pt). 
For the (BATA, cam+pt) experiment, we use a similar weighting strategy for two types of constraints as implemented in Theia~\cite{theia-manual}.
We also compare the result of replacing the global positioning by Theia's LUD~\cite{ozyesil2015robust}.
For the experiments (BATA, cam) and LUD, we perform extra global positioning with fixed cameras to obtain point positions for subsequent bundle adjustment.
We tested on both ETH3D MVS (DSLR)~\cite{schoeps2017cvpr} and IMC 2023~\cite{barron2022mip}.
Results are summarized in Table~\ref{tbl:ablation_pt}.
% with more results in the supp.~material.
We see that relative translation constraints deteriorate convergence and overall performance.
% In this regard, we conclude that it is undesirable to include relative translation constraints.
% Also, we visually summarize the objective value of the global positioning during the optimization Fig.~\ref{fig:loss_function}.
% The convergence rate of the function is sharp, thanks to the bilinear property of the function. 

% We also designed an experiment to demonstrate the robustness of the global positioning. Interested readers can refer to suppl.~material for details.

% \input{tex/fig/ablations}
% \textit{Average} && 75.68 & 83.87 & 85.81 & \textbf{3.39} && 73.37 & 81.77 & 83.82 & 8.51 && 75.14 & 83.32 & 85.33 & 6.83 && \textbf{77.84} & \textbf{85.93} & \textbf{87.79} & 12.26 \\

\begin{table}[t]
    \centering
    \caption{Ablation on global positioning constraints shows points alone perform best.}
    % \vspace{-5px}
    \resizebox{0.6\textwidth}{!}{
    \begin{tabular}{c r c c c c l c c c c
    } \toprule 
        % && \multicolumn{4}{c}{OpenMVG} && \multicolumn{4}{c}{Theia} && \multicolumn{4}{c}{Ours} 
        % && \multicolumn{4}{c}{COLMAP}
        % && \multicolumn{4}{c}{AUC @ $1^\circ$} && \multicolumn{4}{c}{AUC @ $3^\circ$} && \multicolumn{4}{c}{AUC @ $5^\circ$} 
        && \multicolumn{4}{c}{ETH3D DSLR} && \multicolumn{4}{c}{IMC 2023} 
        \\
        \cmidrule{3-6} \cmidrule{8-11} 
        && \tiny{AUC@$1^\circ$} & \tiny{AUC@$3^\circ$} & \tiny{AUC@$3^\circ$} & \tiny{Time (s)} && \tiny{AUC@$3^\circ$} & \tiny{AUC@$5^\circ$} & \tiny{AUC@$10^\circ$} & \tiny{Time (s)}  \\  \midrule
\multicolumn{2}{c}{LUD}  & 77.2 & 85.7 & 87.9 & 37.1 && 64.3 & 69.5 & 74.6 & 987.2  \\
\midrule
\multirow{3}{*}{BATA~~} & cam & 73.5 & 80.5 & 82.4 & 21.2 && 62.1 & 67.0 & 71.6 & 886.9 \\
& pt+cam & 80.1 & 87.7 & 89.5 & 17.1 && 68.6 & 73.6 & 78.3 & 541.2  \\
& pt & \textbf{80.8} & \textbf{88.5} & \textbf{90.3} & \textbf{16.5} && \textbf{69.6} & \textbf{74.6} & \textbf{79.4} & \textbf{497.3} \\
      \bottomrule
    \end{tabular}
    }
    \label{tbl:ablation_pt}
    % \vspace{-5px}
\end{table}

% ETH3D DSLR
% \textit{Average} && 77.2 & 73.5 & \cellcolor{tabsecond}80.1 & \cellcolor{tabfirst}80.8 && 85.7 & 80.5 & \cellcolor{tabsecond}87.7 & \cellcolor{tabfirst}88.5 && 87.9 & 82.4 & \cellcolor{tabsecond}89.5 & \cellcolor{tabfirst}90.3 && 37.1 & 21.2 & \cellcolor{tabsecond}17.1 & \cellcolor{tabfirst}16.5 \\

% IMC2023
% \textit{Average} && 53.3 & 62.1 & \cellcolor{tabsecond}68.6 & \cellcolor{tabfirst}69.6 && 57.7 & 67.0 & \cellcolor{tabsecond}73.6 & \cellcolor{tabfirst}74.6 && 62.1 & 71.6 & \cellcolor{tabsecond}78.3 & \cellcolor{tabfirst}79.4 && 902.5 & 886.9 & \cellcolor{tabsecond}541.2 & \cellcolor{tabfirst}497.3 \\

\subsection{Limitations}
Though generally achieving satisfying performance, there still remain some failure cases.
The major cause is a failure of rotation averaging, \eg, due to symmetric structures (see \textit{Exhibition\_Hall} in Table~\ref{tbl:eth3d_drls}).
In such a case, our method could be combined with existing approaches like Doppelganger~\cite{cai2023doppelgangers}.
Also, since we rely on traditional correspondence search, incorrectly estimated two-view geometries or the inability to match image pairs altogether (\eg, due to drastic appearance or viewpoint changes) will lead to degraded results or, in the worst case, catastrophic failures.

% \section{LImi}
% \textbf{Key }
% \textbf{Limitations}

% \section{Discussions}
% \noindent\textbf{Key to the accurate reconstruction}

% \noindent\textbf{Limitations}

% \vspace{-10px}
\section{Conclusion}

% The major reason for the 
% For dataset as \textit{Exhibition} (see Table~\ref{tbl:eth3d_drls}), the pipeline fails.
In summary, we proposed GLOMAP as a new global SfM pipeline.
Previous systems within this category have been considered more efficient but less robust than incremental approaches.
We revisited the problem and concluded that the key lies in the use of points in the optimization.
Instead of estimating camera positions via ill-posed translation averaging and separately obtaining 3D structure from point triangulation, we merge them into a single global positioning step.
Extensive experiments on various datasets show that the proposed system achieves comparable or superior results to incremental methods in terms of accuracy and robustness while being orders of magnitude faster.
The code is made available as open-source under a commercially friendly license.

\section*{Acknowledgment}
The authors thank Philipp Lindenberger for the thoughtful discussions and comments on the text.
This work was partially funded by the Hasler Stiftung Research Grant via the ETH Zurich Foundation and the ETH Zurich Career Seed Award.
Linfei Pan was supported by gift funding from Microsoft.
% The main contribution of this work is a general-purpose global SfM system that achieves comparable results to incremental methods in terms of accuracy and robustness while being orders of magnitudes faster.
% Our paper provides a survey of the state-of-the-art techniques and combines them, as a first, in a comprehensive, end-to-end reconstruction pipeline.
% The core difference to previous global SfM systems lies in the step of global positioning.
% Instead of first performing ill-posed translation averaging with followed triangulation, our proposed method performs joint camera position estimation with triangulation.
% It achieves a similar level of performance as state-of-the-art incremental SfM systems~\cite{schoenberger2016sfm}, while maintaining the efficiency of global SfM systems.
% Unlike most previous global SfM systems, ours can deal with unknown camera intrinsics (\eg, internet photos) and robustly handles sequential image data (\eg, handheld videos or self-driving car scenarios).

\bibliographystyle{splncs04}
\bibliography{main}

% %%%%%%%%%% Merge with supplemental materials %%%%%%%%%%
% \newpage
% \begin{center}
% \textbf{\large Supplemental Materials}
% \end{center}
% %%%%%%%%%% Merge with supplemental materials %%%%%%%%%%
% %%%%%%%%%% Prefix a "S" to all equations, figures, tables and reset the counter %%%%%%%%%%
% \setcounter{equation}{0}
% \setcounter{figure}{0}
% \setcounter{table}{0}
% \setcounter{section}{0}
% \makeatletter
% \renewcommand{\theequation}{S\arabic{equation}}
% \renewcommand{\thefigure}{S\arabic{figure}}
% \renewcommand{\thetable}{S\arabic{table}}
% \renewcommand{\thesection}{S\arabic{section}}
% %%%%%%%%%% Prefix a "S" to all equations, figures, tables and reset the counter %%%%%%%%%%

% \input{tex/appendix}

% \end{document}

%%%%%%%%%% Merge with supplemental materials %%%%%%%%%%
\newpage
\begin{center}
\textbf{\large Supplemental Materials}
\end{center}
%%%%%%%%%% Merge with supplemental materials %%%%%%%%%%
%%%%%%%%%% Prefix a "S" to all equations, figures, tables and reset the counter %%%%%%%%%%
\setcounter{equation}{0}
\setcounter{figure}{0}
\setcounter{table}{0}
\setcounter{section}{0}
\makeatletter
\renewcommand{\theequation}{S\arabic{equation}}
\renewcommand{\thefigure}{S\arabic{figure}}
\renewcommand{\thetable}{S\arabic{table}}
\renewcommand{\thesection}{S\arabic{section}}
%%%%%%%%%% Prefix a "S" to all equations, figures, tables and reset the counter %%%%%%%%%%

% -----------------------------------------------------
% Appendix
% -----------------------------------------------------

\section{Additional Comparisons}
In this section, we present extra experiments in comparison with HSfM~\cite{cui2017hsfm} and LiGT~\cite{cai2021pose}, which are two more camera pose estimation pipelines.

\noindent{\textbf{HSfM}}~\cite{cui2017hsfm} is a hybrid Structure-from-Motion (SfM) pipeline that estimates camera rotations with global rotation averaging and estimates camera translation with incremental SfM.
In this experiment, we compare with the algorithm implemented in the Theia~\cite{theia-manual} library.
By default, the Theia implementation does not run a full bundle adjustment (BA) at the end of the HSfM reconstruction.
For a fair comparison, we added these extra steps in the pipeline.
Otherwise, we use the default parameter setting.
%Wrwhile the pipeline is modified to have a final triangulation and full bundle adjustment.
%  and use the default parameters.

\noindent{\textbf{LiGT~\cite{cai2021pose}}} is a camera pose estimation algorithm based on linear constraints posed by points.
We use the implementation in OpenMVG~\cite{moulon2016openmvg}, provided by the authors\footnote{\url{https://github.com/openMVG/openMVG/pull/2065}}, and experiment with the same pipeline as OpenMVG experiments.

The results on the datasets providing camera intrinsics are reported in Table~\ref{tbl:sm_calibrated}.
On the ETH3D MVS rig and DSLR~\cite{schoeps2017cvpr} datasets, HSfM achieves comparable results to the proposed GLOMAP.
However, on the ETH3D SLAM~\cite{Schops_2019_CVPR} and LaMAR~\cite{sarlin2022lamar} datasets, HSfM fails, leading to significantly less accurate reconstructions than all other pipelines.
We attribute this to the sparsity and the sequential nature of these datasets. 

%The average performance of the method is worse than other methods due to more failure cases for HSfM.
%However, for ETH3D SLAM~\cite{Schops_2019_CVPR} and LaMAR~\cite{sarlin2022lamar}, due to the sparsity and the sequential nature of the data, incremental pipelines without adjusting the camera rotations can subject to heavier drifting, thus HSfM falls short on these datasets.
% For the ETH3D MVS (rig)~\cite{schoeps2017cvpr} dataset, HSfM achieves comparable results as the proposed method.
% However, for other datasets, the method fails to compete against other methods.
% We attribute this to the fact that HSfM only performs a single low-cost global bundle adjustment with camera rotation and intrinsics in the end.
% In this regard, the optimization starts from a point when it close to local minima, thus it is difficult to rectify the accumulated errors during the incremental pose estimation.
% Thus, it requires high accuracy in the rotation averaging, and initial camera intrinsics,

Results on the datasets lacking camera calibrations are summarized in Table~\ref{tbl:sm_uncalibrated}.
On the IMC 2023 and MIP360 datasets, HSfM substantially falls behind all tested methods in terms of accuracy. 
This is expected as HSfM assumes known intrinsics which are then kept fixed until the last BA step. 
On these datasets, we are only given coarse intrinsics prior, which is not sufficient for reconstruction without additional refinement steps. 

%As expected, the coarse camera intrinsics prior are not accurate enough to recover accurate camera position.
%In consequence, the performance of the method falls short in comparison with other methods.
% the camera intrinsics are not known for these 

% Qualitatively the result lo
% , and that on uncalibrated datasets are in Table~\ref{tbl:sm_uncalibrated}.
% The method does not pero

% \input{tex/tbl/eth3d_dslr_hsfm}
% \input{tex/tbl/eth3d_slam_hsfm}
% \input{tex/tbl/eth3d_rig_hsfm}
% \input{tex/tbl/mip360_hsfm}
% \input{tex/tbl/imc_hsfm}
% \input{tex/tbl/lamar_map_hsfm}

\begin{table}[h]
    \centering
    \caption{Results on datasets with known camera intrinsics.}
    % \vspace{-5px}
    \resizebox{\textwidth}{!}{
    \begin{tabular}{l r c c c c l c c c c l c c c c l c c c c
    } \toprule 
        % && \multicolumn{4}{c}{OpenMVG} && \multicolumn{4}{c}{Theia} && \multicolumn{4}{c}{Ours} 
        % && \multicolumn{4}{c}{COLMAP}
        % && \multicolumn{4}{c}{AUC @ $1^\circ$} && \multicolumn{4}{c}{AUC @ $3^\circ$} && \multicolumn{4}{c}{AUC @ $5^\circ$} 
        && \multicolumn{4}{c}{ETH3D SLAM} && \multicolumn{4}{c}{ETH3D MVS (rig)} && \multicolumn{4}{c}{ETH3D MVS (DSLR)} && \multicolumn{4}{c}{LaMAR} 
        \\
        \cmidrule{3-6} \cmidrule{8-11} \cmidrule{13-16} \cmidrule{18-21}
        && \tiny{~R@0.1m~} & \tiny{AUC@$0.1$m} & \tiny{AUC@$0.5$m} & \tiny{Time (s)} && \tiny{AUC@$1^\circ$} & \tiny{AUC@$3^\circ$} & \tiny{AUC@$5^\circ$}  & \tiny{Time (s)} && \tiny{AUC@$1^\circ$} & \tiny{AUC@$3^\circ$} & \tiny{AUC@$5^\circ$} & \tiny{Time (s)} && \tiny{~~R@1m~~} & \tiny{AUC@$1$m} & \tiny{AUC@$5$m} & \tiny{Time (s)}  \\  \midrule
OpenMVG && 48.2 & 34.9 & 48.6 & 120.8
&& 0.2 & 0.9 & 1.3 & \cellcolor{tabsecond}165.3 
&& 60.4 & 68.1 & 70.1 & 26.5
&& - & - & - & - \\ 
Theia  && \cellcolor{tabsecond}62.8 & 46.0 & \cellcolor{tabsecond}61.1 & \cellcolor{tabfirst}91.8
&& 40.4 & \cellcolor{tabsecond}72.6 & \cellcolor{tabsecond}81.4 & 350.5 
&& 71.8 & 79.2 & 81.2 & \cellcolor{tabfirst}7.9
&& 11.1 & 4.4 & 19.1 & \cellcolor{tabfirst}1542.0 \\
HSfM && 42.5 & 36.7 & 41.7 & \cellcolor{tabsecond}117.2
&& 40.5 & \cellcolor{tabsecond}72.6 & 81.3 & 745.6 
&& 65.0 & 70.5 & 71.7 & 17.2
&& 10.4 & 6.9 & 10.9 & \cellcolor{tabsecond} 8670.0 \\
LiGT && 34.0 & 27.8 & 33.4 & 210.3
&& 0.0 & 0.0 & 0.1 & \cellcolor{tabfirst}127.1
&& 50.0 & 56.2 & 57.8 & 18.9
&& - & - & - & - \\
GLOMAP && \cellcolor{tabfirst}66.4 & \cellcolor{tabfirst}57.0 & \cellcolor{tabfirst}65.7 & 133.5
&& \cellcolor{tabfirst}57.6 & \cellcolor{tabfirst}80.9 & \cellcolor{tabfirst}87.2 & 793.8
&& \cellcolor{tabfirst}80.8 & \cellcolor{tabfirst}88.5 & \cellcolor{tabfirst}90.3 & \cellcolor{tabsecond}16.5
&& \cellcolor{tabfirst}49.1 & \cellcolor{tabfirst}24.5 & \cellcolor{tabfirst}50.9 & 12405.3 \\
\hdashline
COLMAP && 57.9 & \cellcolor{tabsecond}47.6 & 57.9 & 1115.4
&& \cellcolor{tabsecond}45.5 & 64.0 & 69.1 & 2857.0
&& \cellcolor{tabsecond}79.2 & \cellcolor{tabsecond}86.5 & \cellcolor{tabsecond}88.1 & 27.5 
&& \cellcolor{tabsecond}32.0 & \cellcolor{tabsecond}13.8 & \cellcolor{tabsecond}39.4 & 354660.4 \\
\bottomrule

    \end{tabular}
    }
    \label{tbl:sm_calibrated}
    % \vspace{-15px}
\end{table}

% \begin{table}[h]
%     \centering
%     \caption{Results on uncalibrated datasets.}
%     % \vspace{-5px}
%     \resizebox{0.6\textwidth}{!}{
%     \begin{tabular}{l r c c c c l c c c c 
%     } \toprule 
%         % && \multicolumn{4}{c}{OpenMVG} && \multicolumn{4}{c}{Theia} && \multicolumn{4}{c}{Ours} 
%         % && \multicolumn{4}{c}{COLMAP}
%         % && \multicolumn{4}{c}{AUC @ $1^\circ$} && \multicolumn{4}{c}{AUC @ $3^\circ$} && \multicolumn{4}{c}{AUC @ $5^\circ$} 
%         && \multicolumn{4}{c}{IMC2023} && \multicolumn{4}{c}{MIP360}
%         \\
%         \cmidrule{3-6} \cmidrule{8-11} 
%         && \tiny{AUC@$3^\circ$} & \tiny{AUC@$5^\circ$} & \tiny{AUC@$10^\circ$} & \tiny{Time (s)} && \tiny{AUC@$3^\circ$} & \tiny{AUC@$5^\circ$} & \tiny{AUC@$10^\circ$} & \tiny{Time (s)} \\
%         \midrule
% OpenMVG && 22.0 & 31.2 & 44.1 & 412.6 && 73.4 & 78.3 & 84.9 & 100.2 \\
% Theia && 22.6 & 29.4 & 40.6 & \textbf{149.6} && 49.2 & 52.1 & 58.5 & \textbf{87.2} \\
% HSfM &&  12.3 & 17.7 & 26.8 & 219.6 && 38.5 & 43.8 & 48.9 & 149.8 \\
% GLOMAP && 55.7 & 63.3 & 71.3 & 267.9 && 93.6 & 96.1 & 98.0 & 164.5 \\
% \hdashline
% COLMAP && \textbf{65.0} & \textbf{70.2} & \textbf{74.9} & 4051.0 && \textbf{96.4}& \textbf{97.8} & \textbf{98.9} & 345.7 \\
% \bottomrule
%     \end{tabular}
%     }
%     \label{tbl:sm_uncalibrated}
%     % \vspace{-15px}
% \end{table}

\begin{table}[h]
    \centering
    \caption{Results on datasets with missing camera intrinsics.}
    % \vspace{-5px}
    \resizebox{0.6\textwidth}{!}{
    \begin{tabular}{l r c c c c l c c c c 
    } \toprule 
        % && \multicolumn{4}{c}{OpenMVG} && \multicolumn{4}{c}{Theia} && \multicolumn{4}{c}{Ours} 
        % && \multicolumn{4}{c}{COLMAP}
        % && \multicolumn{4}{c}{AUC @ $1^\circ$} && \multicolumn{4}{c}{AUC @ $3^\circ$} && \multicolumn{4}{c}{AUC @ $5^\circ$} 
        && \multicolumn{4}{c}{IMC 2023} && \multicolumn{4}{c}{MIP360}
        \\
        \cmidrule{3-6} \cmidrule{8-11} 
        && \tiny{AUC@$3^\circ$} & \tiny{AUC@$5^\circ$} & \tiny{AUC@$10^\circ$} & \tiny{Time (s)} && \tiny{AUC@$3^\circ$} & \tiny{AUC@$5^\circ$} & \tiny{AUC@$10^\circ$} & \tiny{Time (s)} \\
        \midrule
OpenMVG && 20.4 & 28.7 & 40.4 & 412.6 
&& 73.4 & 78.3 & 84.9 & \cellcolor{tabsecond}92.6 \\
Theia && 42.5 & 49.2 & 56.5 & \cellcolor{tabfirst}149.6 
&& 51.6 & 53.4 & 55.0 & \cellcolor{tabfirst}88.2 \\
HSfM &&  29.7 & 36.5 & 44.2 & \cellcolor{tabsecond}325.2
&& 37.1 & 40.7 & 44.5 & 248.9 \\
LiGT && 8.9 & 14.0 & 22.9 & 438.7
&& 43.6 & 49.2 & 57.1 & 107.1 \\
GLOMAP && \cellcolor{tabfirst}69.6 & \cellcolor{tabfirst}74.6 & \cellcolor{tabfirst}79.4 & 497.3 
&& \cellcolor{tabfirst}97.5 & \cellcolor{tabfirst}98.5 & \cellcolor{tabfirst}99.2 & 221.0  \\
\hdashline
COLMAP && \cellcolor{tabsecond}65.3 & \cellcolor{tabsecond}70.4 & \cellcolor{tabsecond}75.1 & 4051.0 
&& \cellcolor{tabsecond}96.5 & \cellcolor{tabsecond}97.9 & \cellcolor{tabsecond}98.9 & 345.7 \\
\bottomrule
    \end{tabular}
    }
    \label{tbl:sm_uncalibrated}
    % \vspace{-15px}
\end{table}

% OpenMVG, Theia, HSfM, GLOMAP, COML
% IMC2023
% \textit{Average} && 20.4 & 42.5 & 29.7 & \cellcolor{tabfirst}69.6 & \cellcolor{tabsecond}65.3 && 28.7 & 49.3 & 36.5 & \cellcolor{tabfirst}74.6 & \cellcolor{tabsecond}70.4 && 40.4 & 56.5 & 44.2 & \cellcolor{tabfirst}79.4 & \cellcolor{tabsecond}75.1 && \cellcolor{tabsecond}412.6 & \cellcolor{tabfirst}149.6 & 507.5 & 497.3 & 4051.0 \\

For more direct comparison with LiGT~\cite{cai2021pose}, additional results on Strecha~\cite{strecha2008benchmarking} dataset are summarized in Table~\ref{tbl:ligt_strecha}.

\begin{table}[h]
    \centering
    % \vspace{-20px}
    \caption{Average camera position errors (in mm) for Strecha dataset~\cite{strecha2008benchmarking}. Rows with $^*$ are taken from the paper~\cite{cai2021pose}.}
    \resizebox{0.8\columnwidth}{!}{
    \begin{tabular}{l l c c c c c c
    } \toprule 
        && Herz-Jesus-P8 & Herz-Jesus-P25 & Fountain-P11 & Entry-P10 & Castle-P19 & Castle-P30 \\\midrule
        LiGT$^*$ && 5.01 & 6.86 & 3.17 & \textbf{5.50} & 41.88 & 49.84 \\
        LiGT && \textbf{3.54} & \textbf{5.29} & 2.81 & 9.08 & \textbf{24.72} & 36.36 \\
        GLOMAP && 4.13 & 5.40 & \textbf{2.79} & 6.32 & 24.95 & \textbf{22.36} \\
      \bottomrule
    \end{tabular}
    }
    % \vspace{-10px}
    \label{tbl:ligt_strecha}
% \vspace{-10px}
\end{table}

\section{Additional Reconstruction Results}
\begin{figure}[t]
    \centering
    \includegraphics[height=0.18\textwidth]{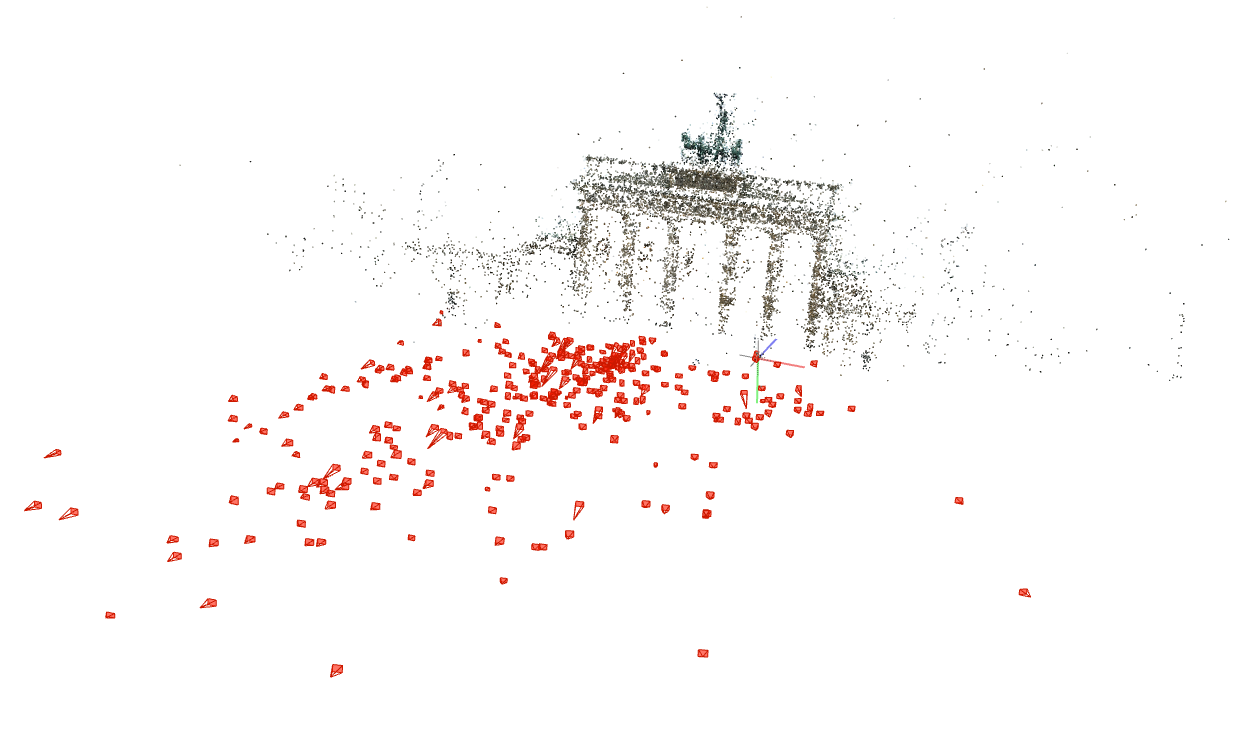}
    \includegraphics[height=0.18\textwidth]{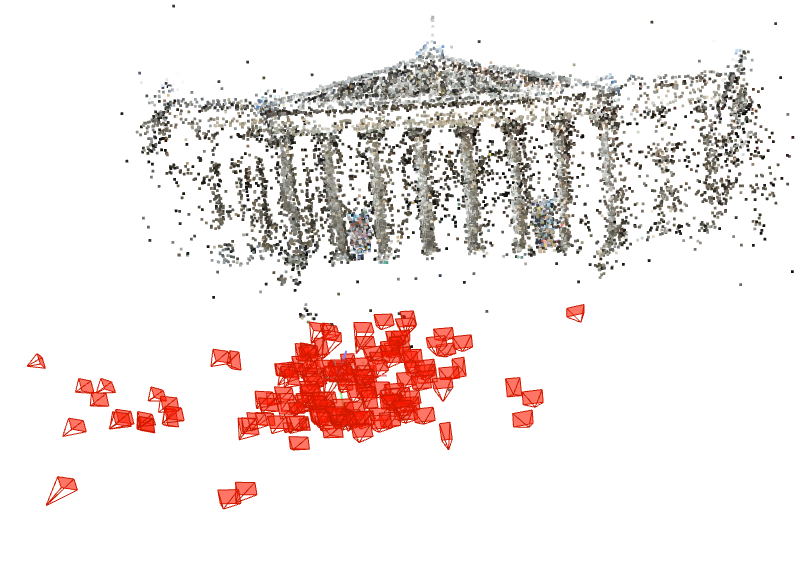}
    \includegraphics[height=0.18\textwidth]{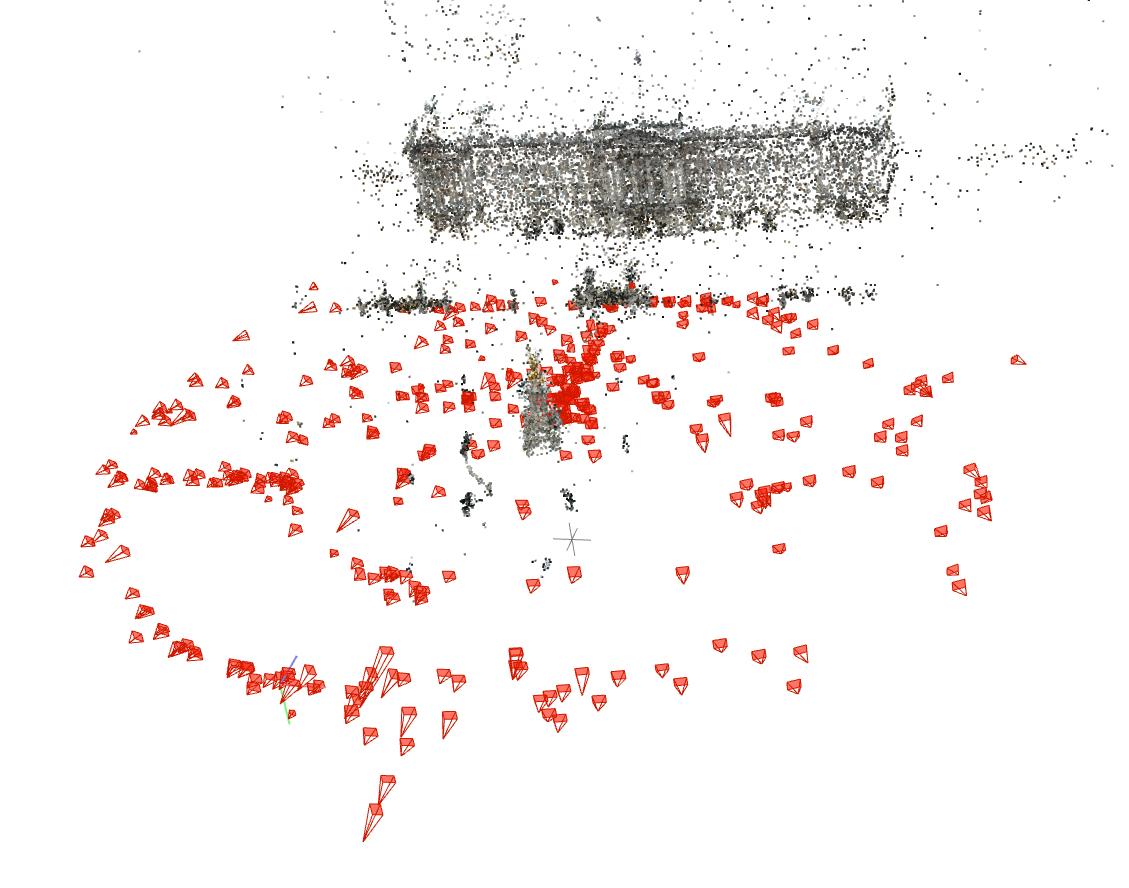}
    \includegraphics[height=0.18\textwidth]{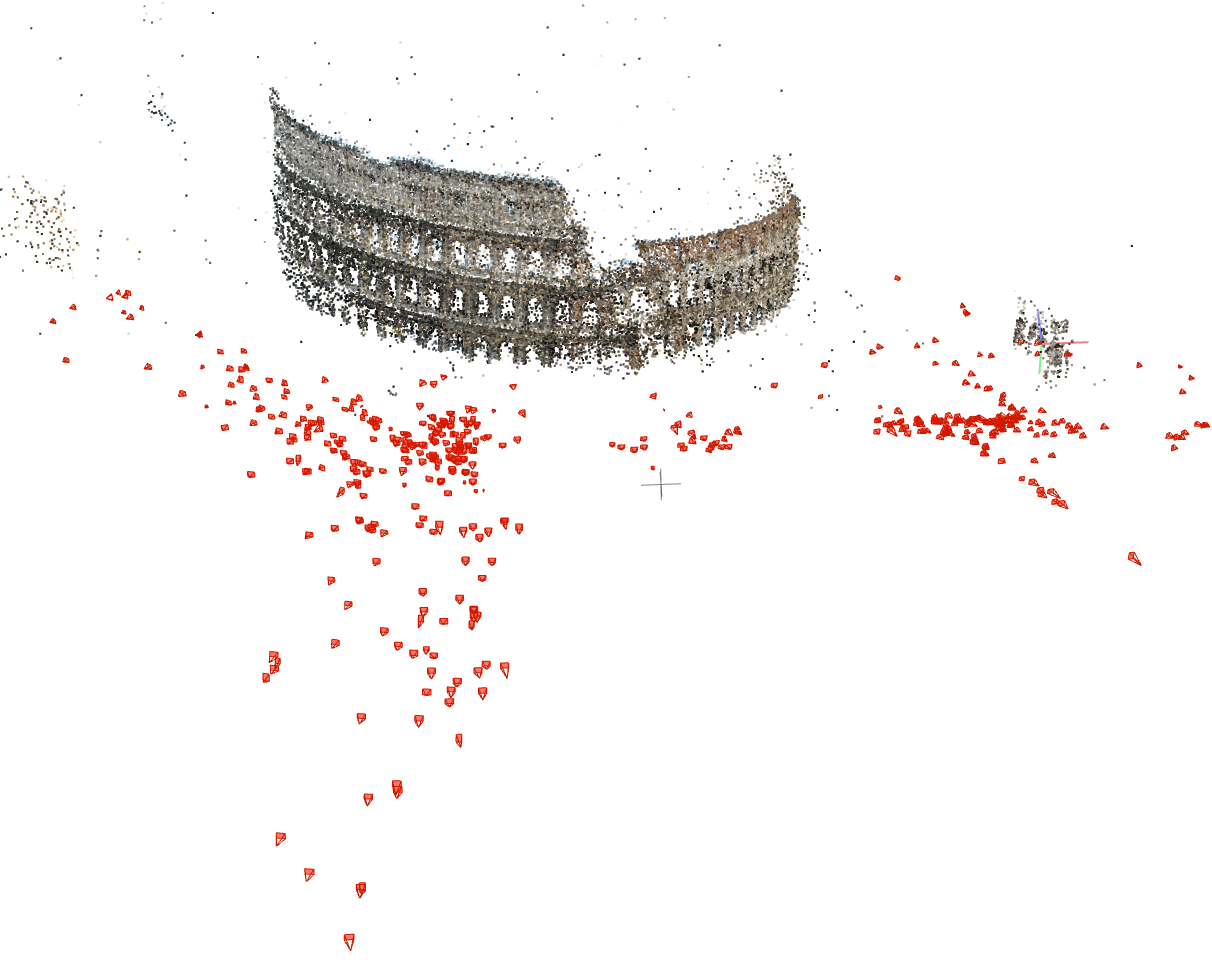}
    \includegraphics[height=0.18\textwidth]{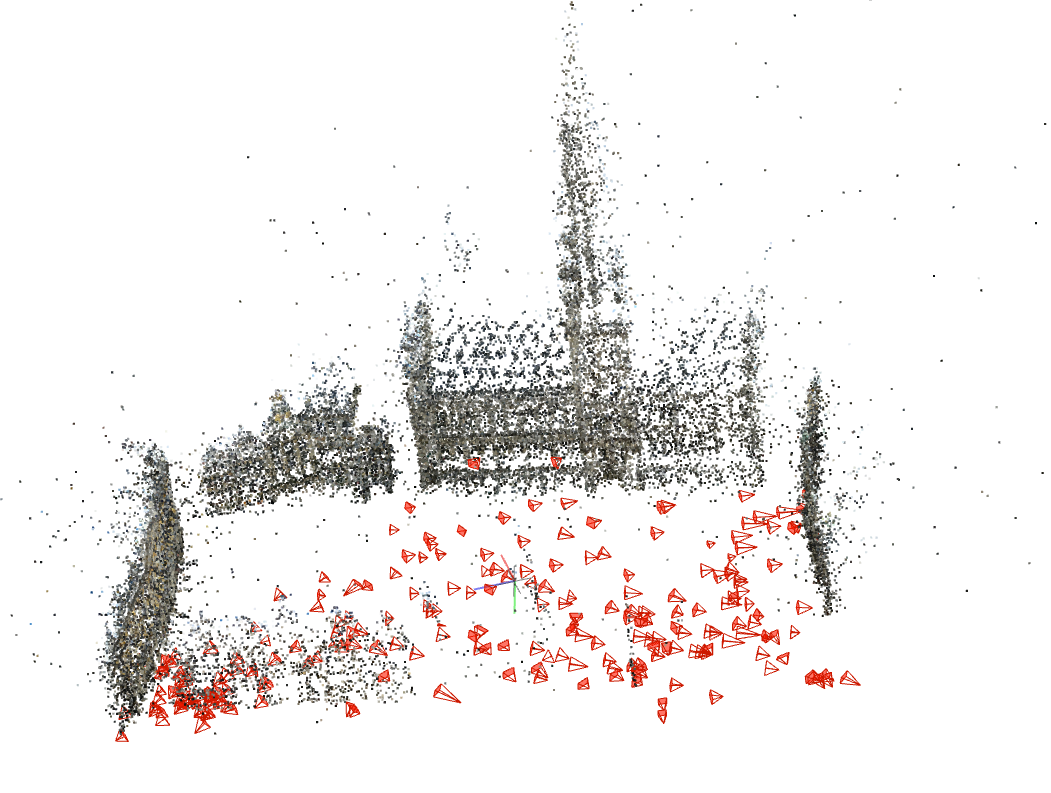}
    \includegraphics[height=0.18\textwidth]{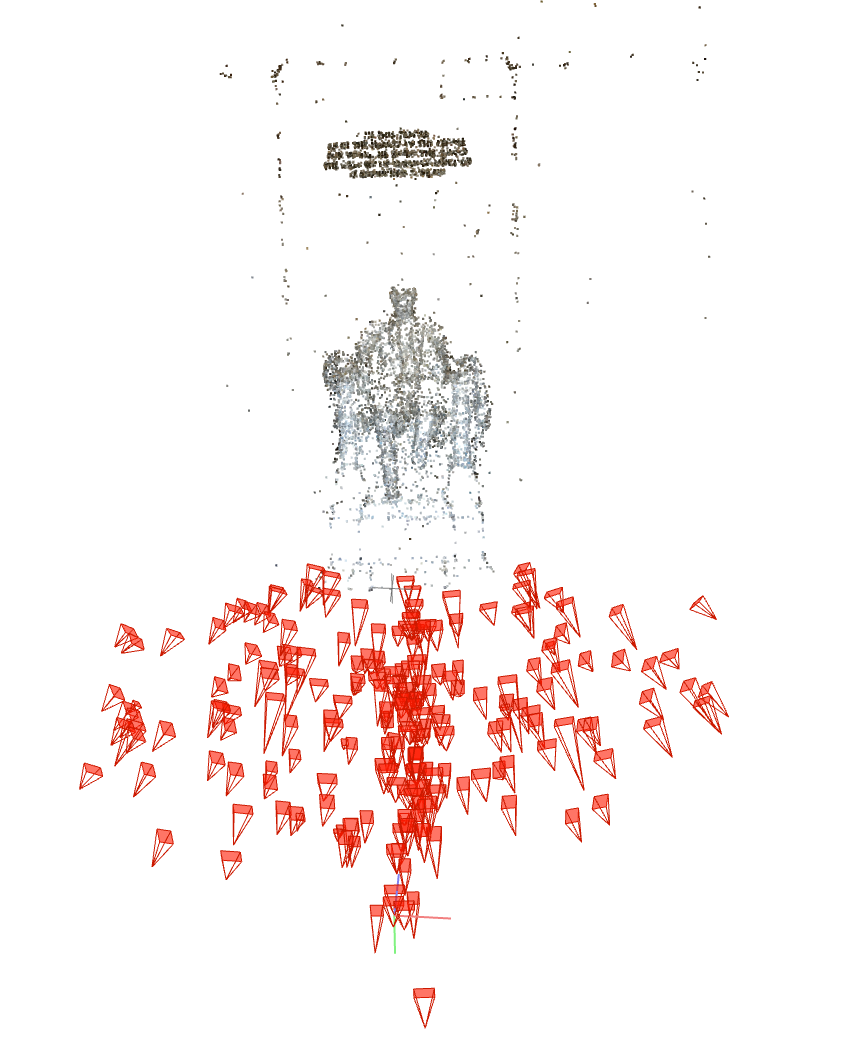}
    \includegraphics[height=0.18\textwidth]{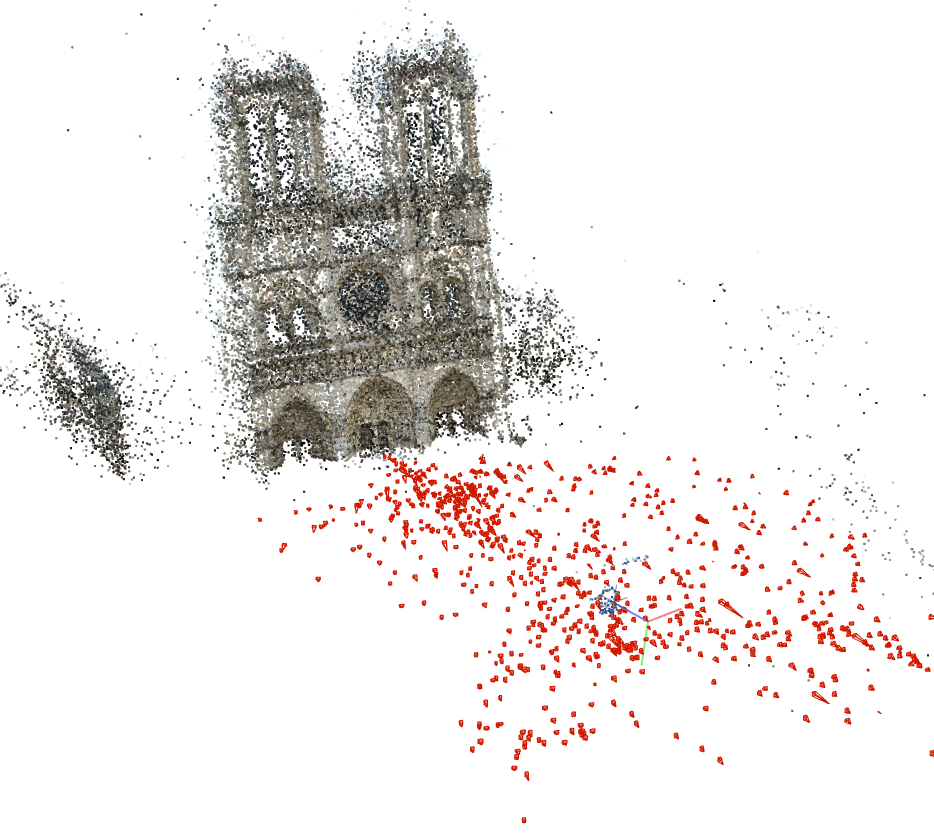}
    \includegraphics[height=0.18\textwidth]{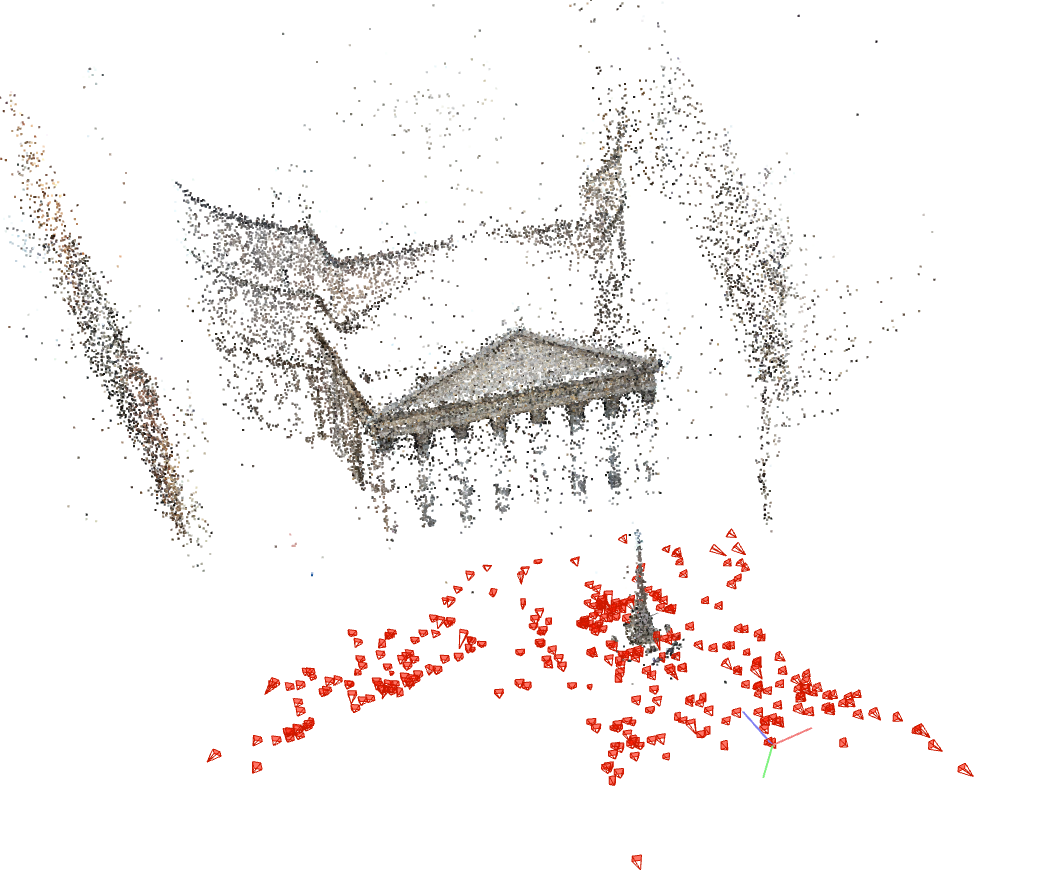}
    \includegraphics[height=0.18\textwidth]{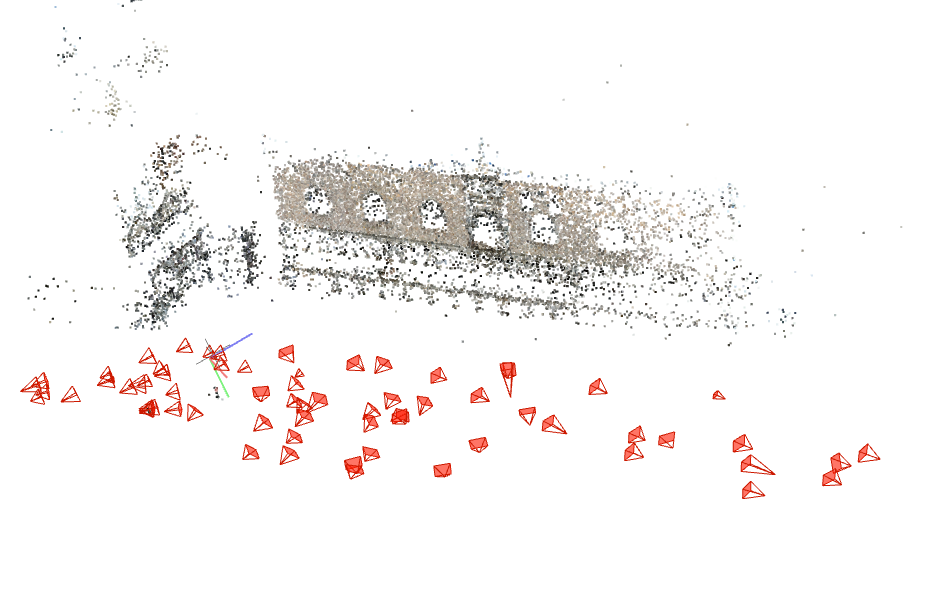}
    \includegraphics[height=0.18\textwidth]{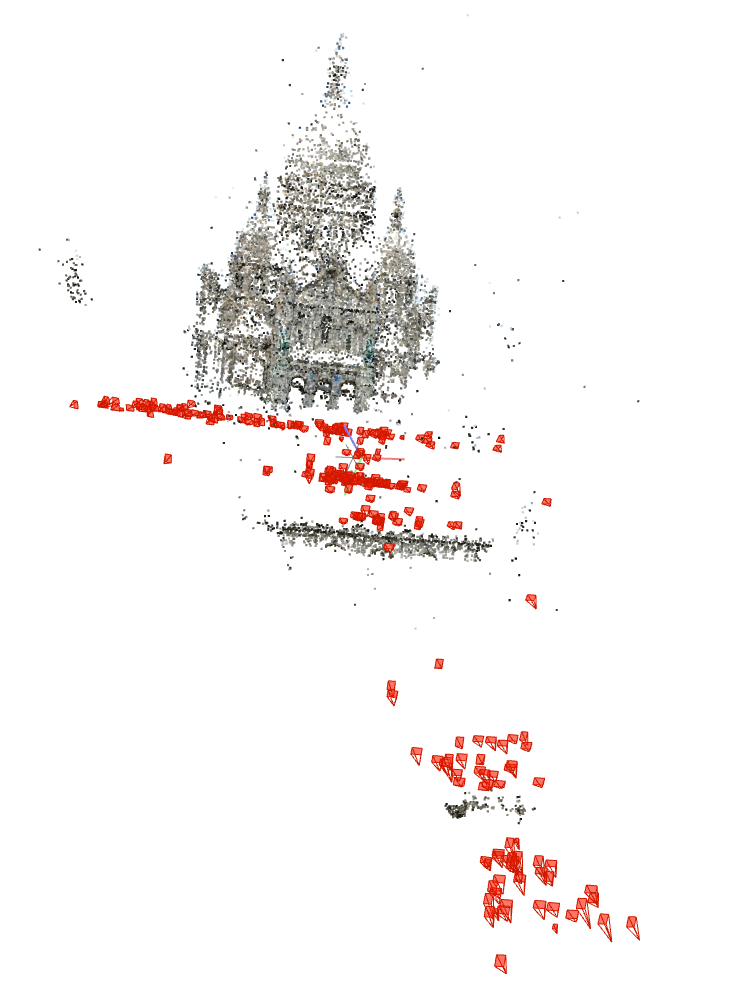}
    \includegraphics[height=0.18\textwidth]{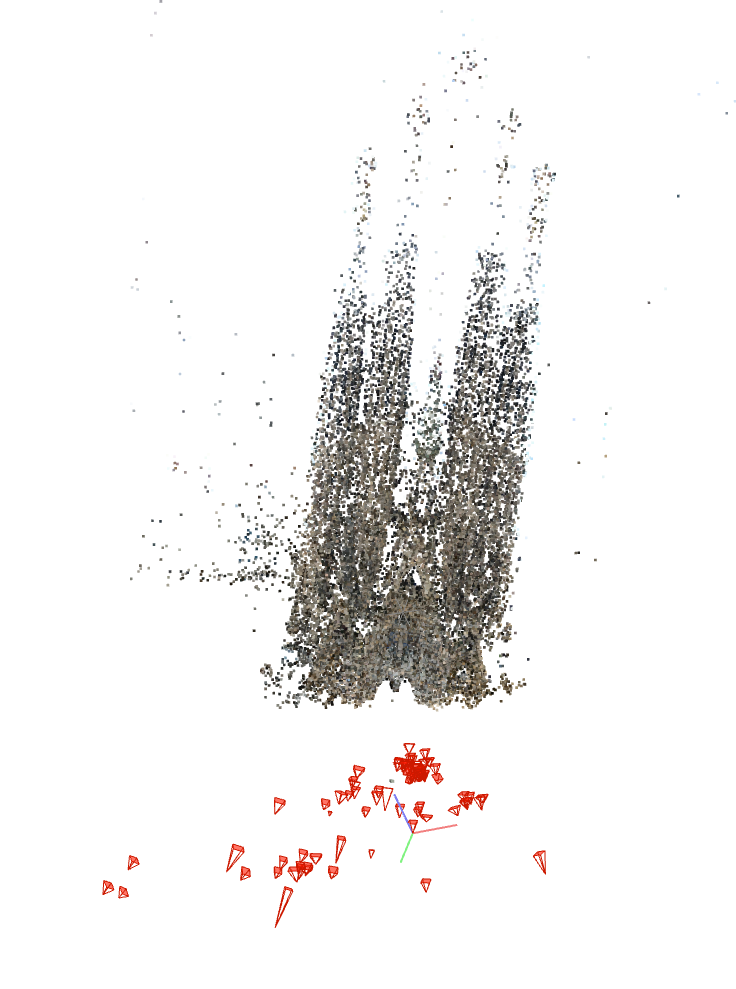}
    \includegraphics[height=0.18\textwidth]{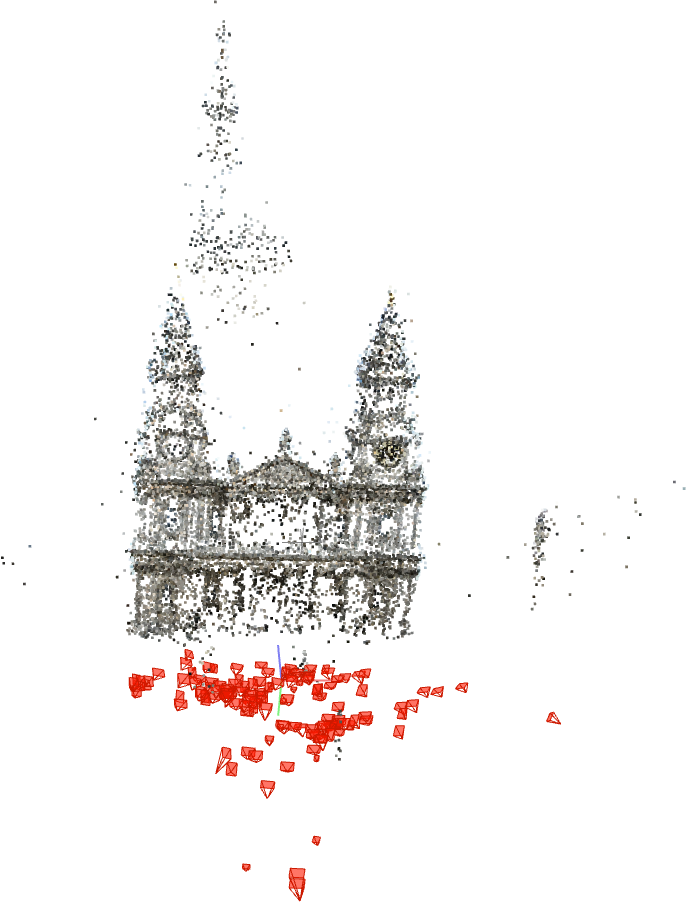}
    \includegraphics[height=0.18\textwidth]{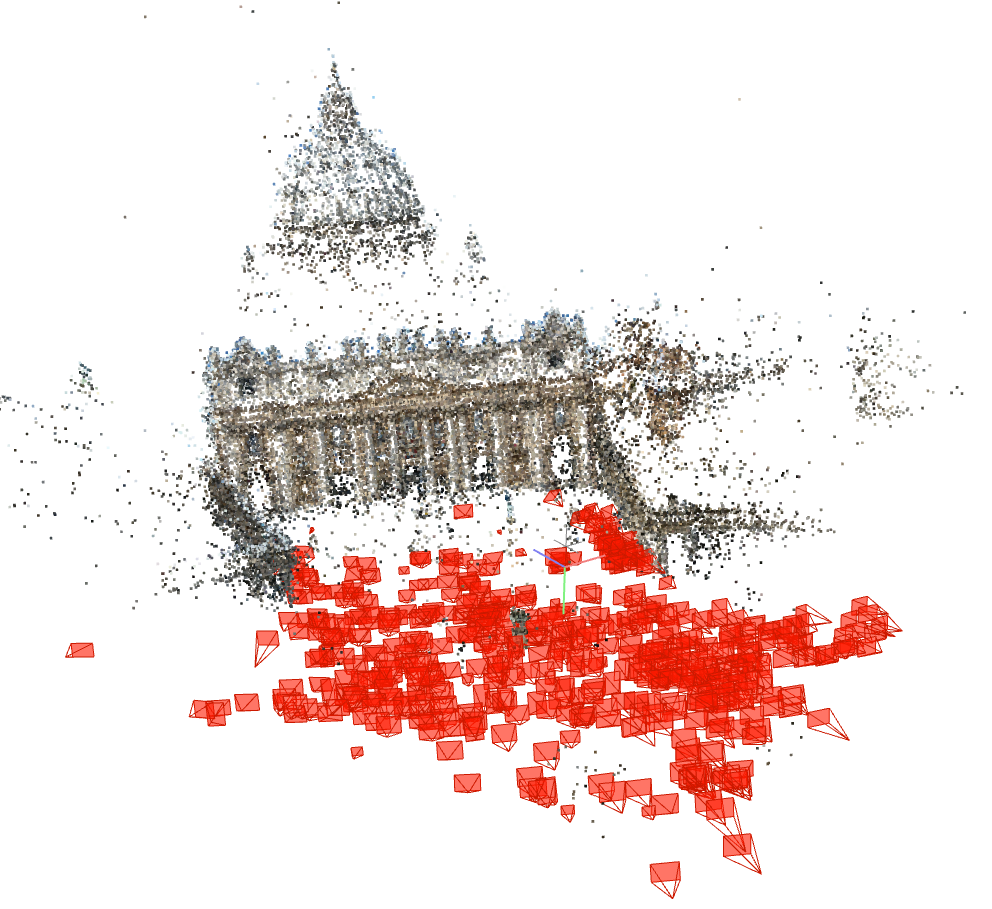}
    \includegraphics[height=0.18\textwidth]{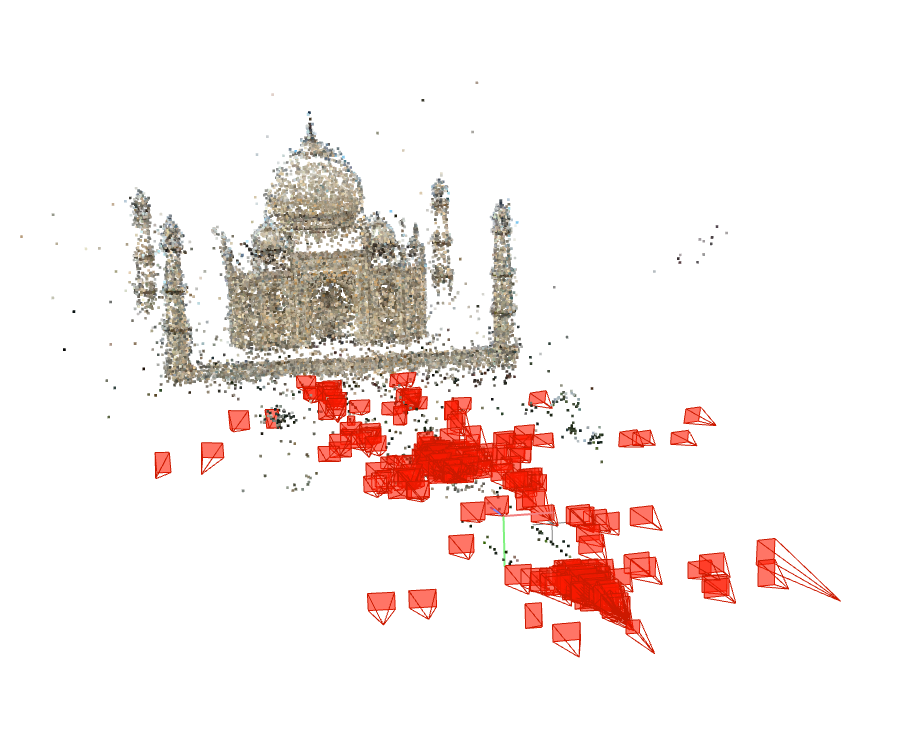}
    \includegraphics[height=0.18\textwidth]{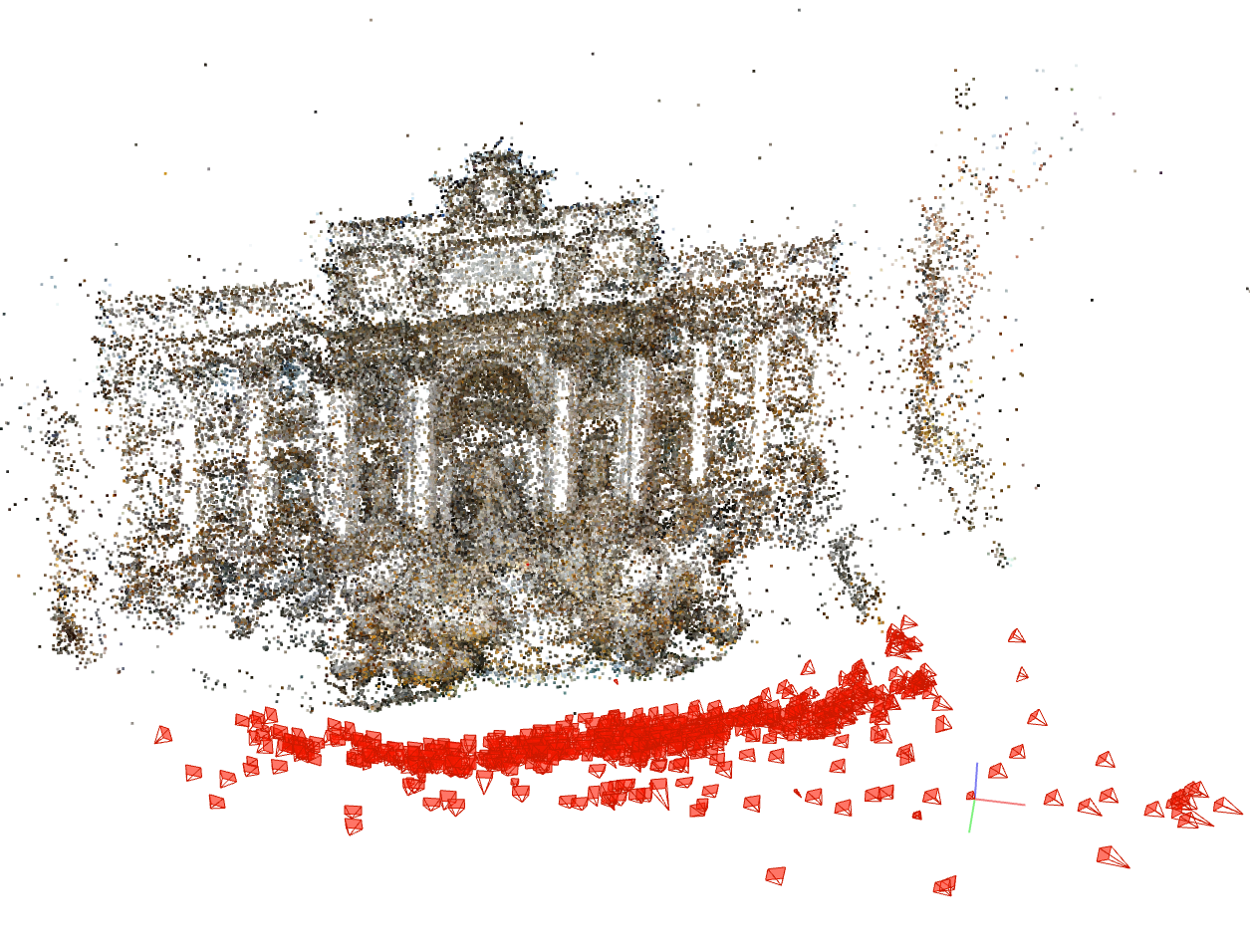}
    \includegraphics[height=0.18\textwidth]{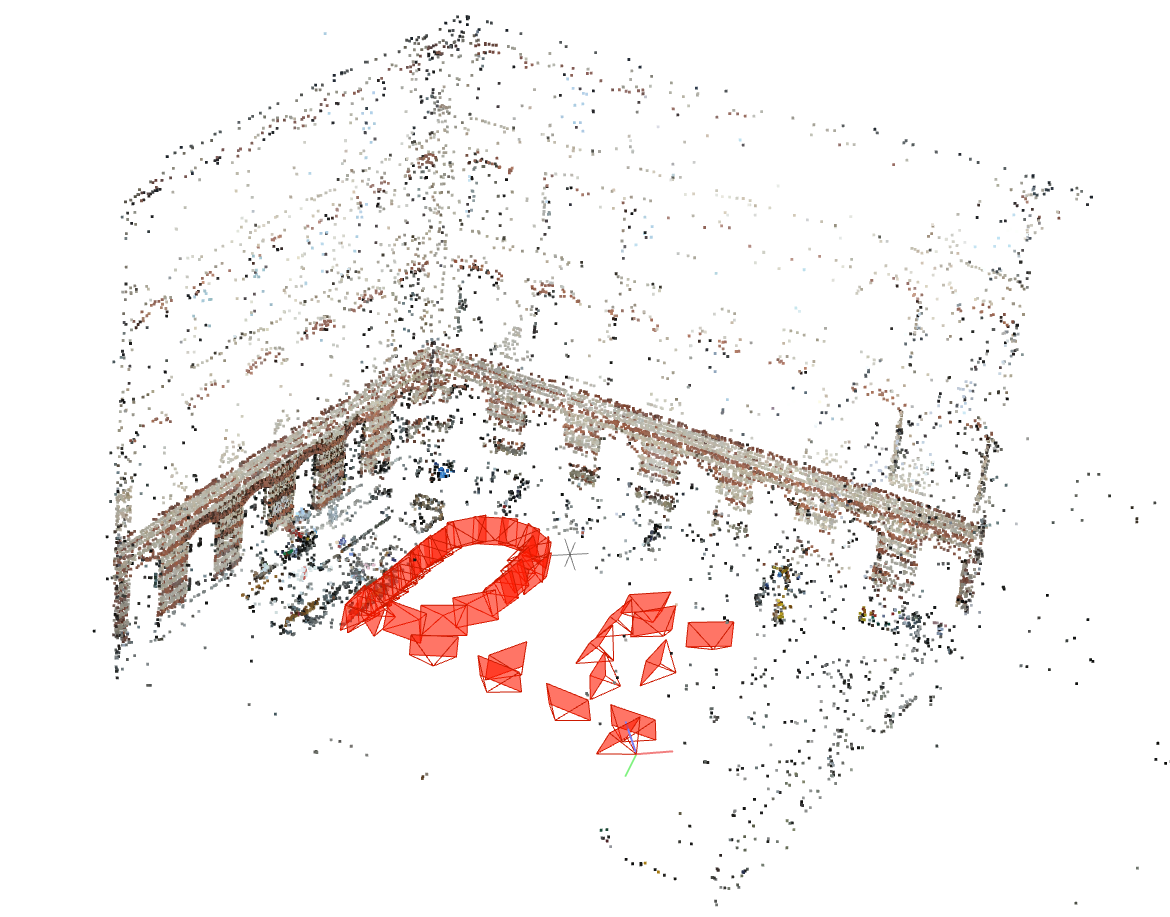}
    \includegraphics[height=0.18\textwidth]{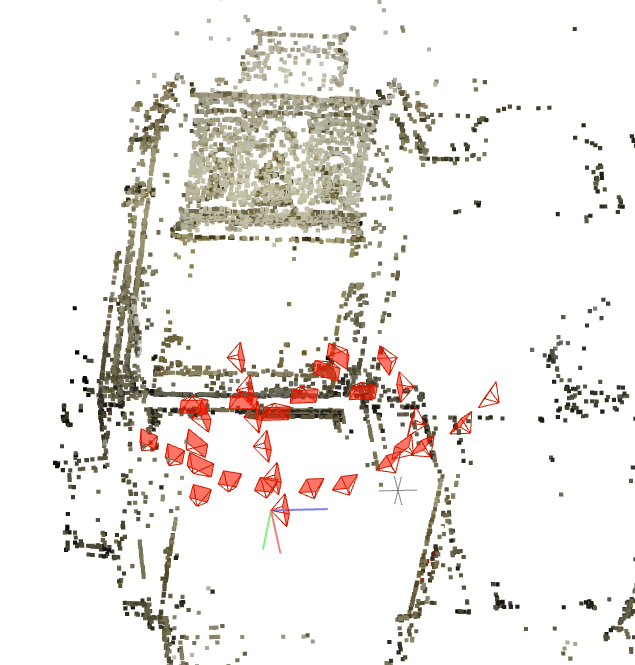}
    \includegraphics[height=0.18\textwidth]{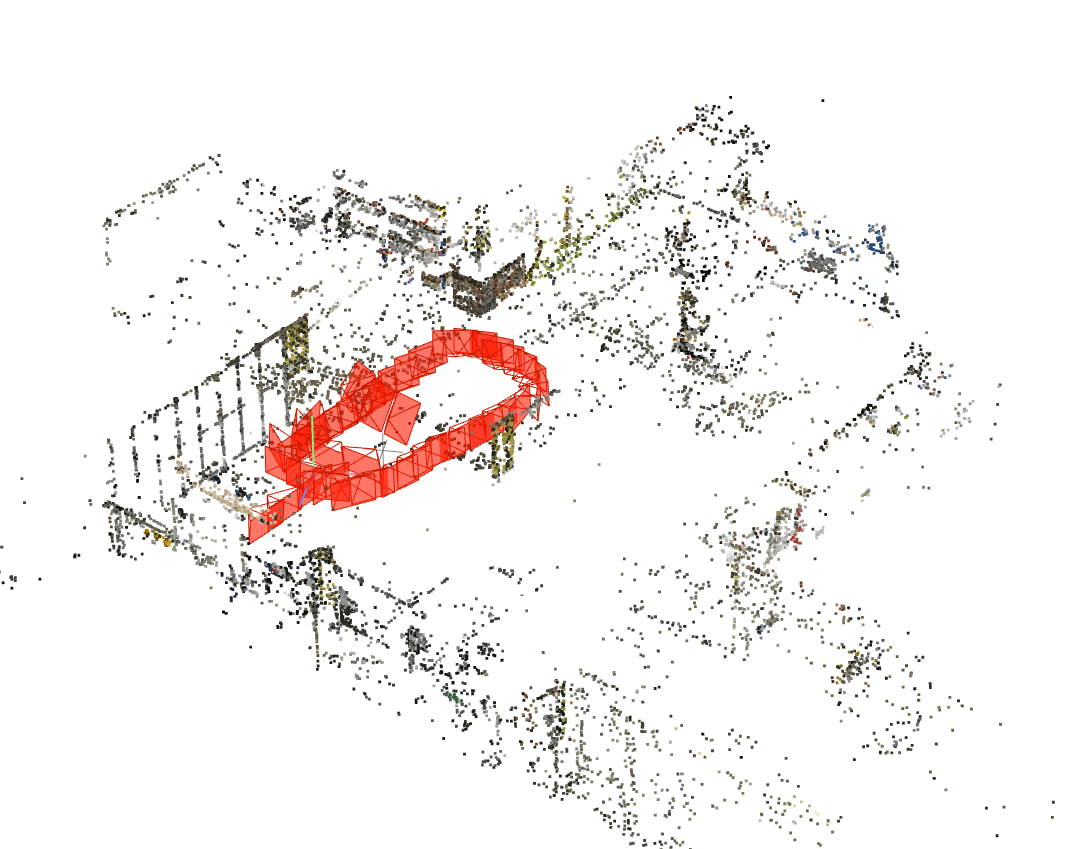}
    \includegraphics[height=0.18\textwidth]{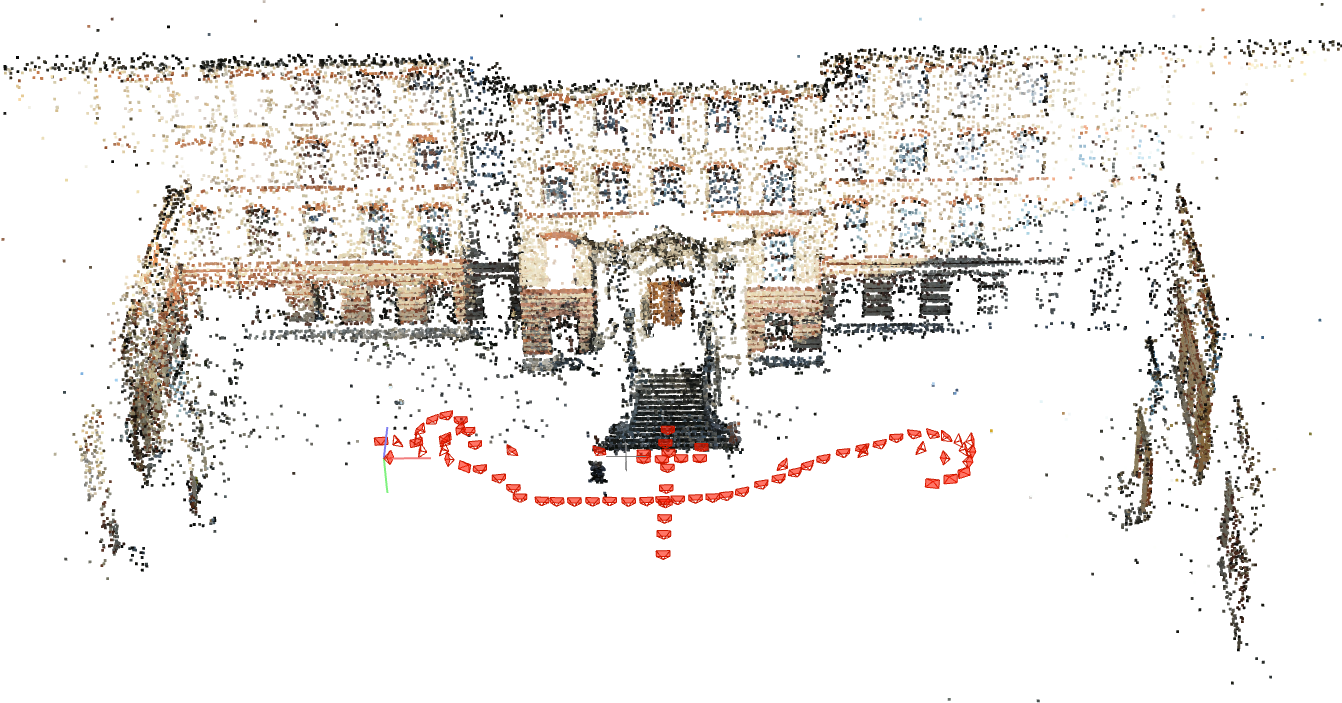}
    \includegraphics[height=0.18\textwidth]{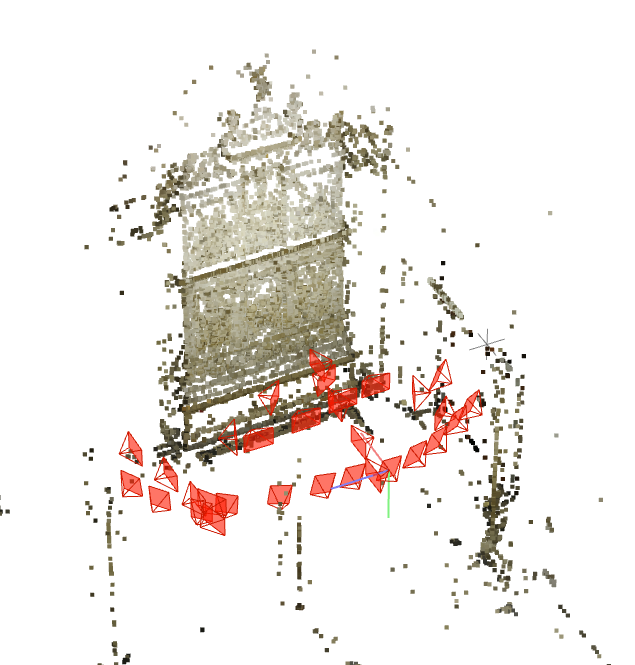}
    \includegraphics[height=0.18\textwidth]{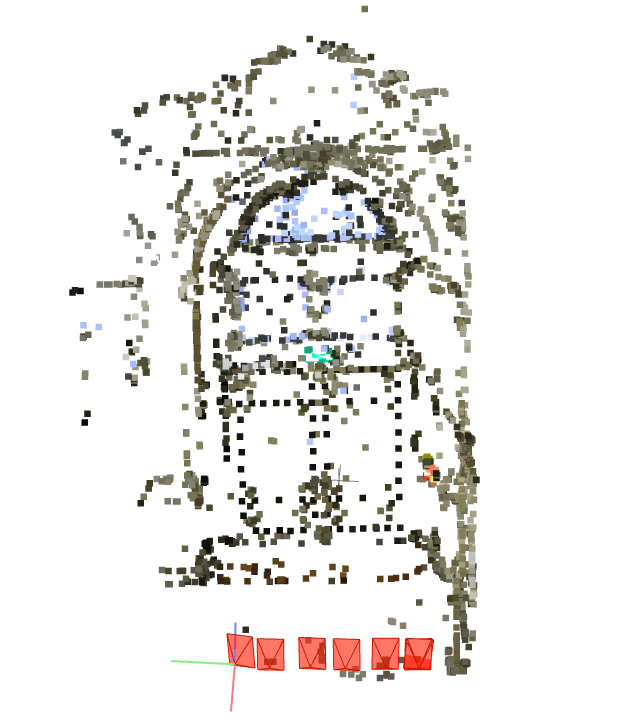}
    \includegraphics[height=0.18\textwidth]{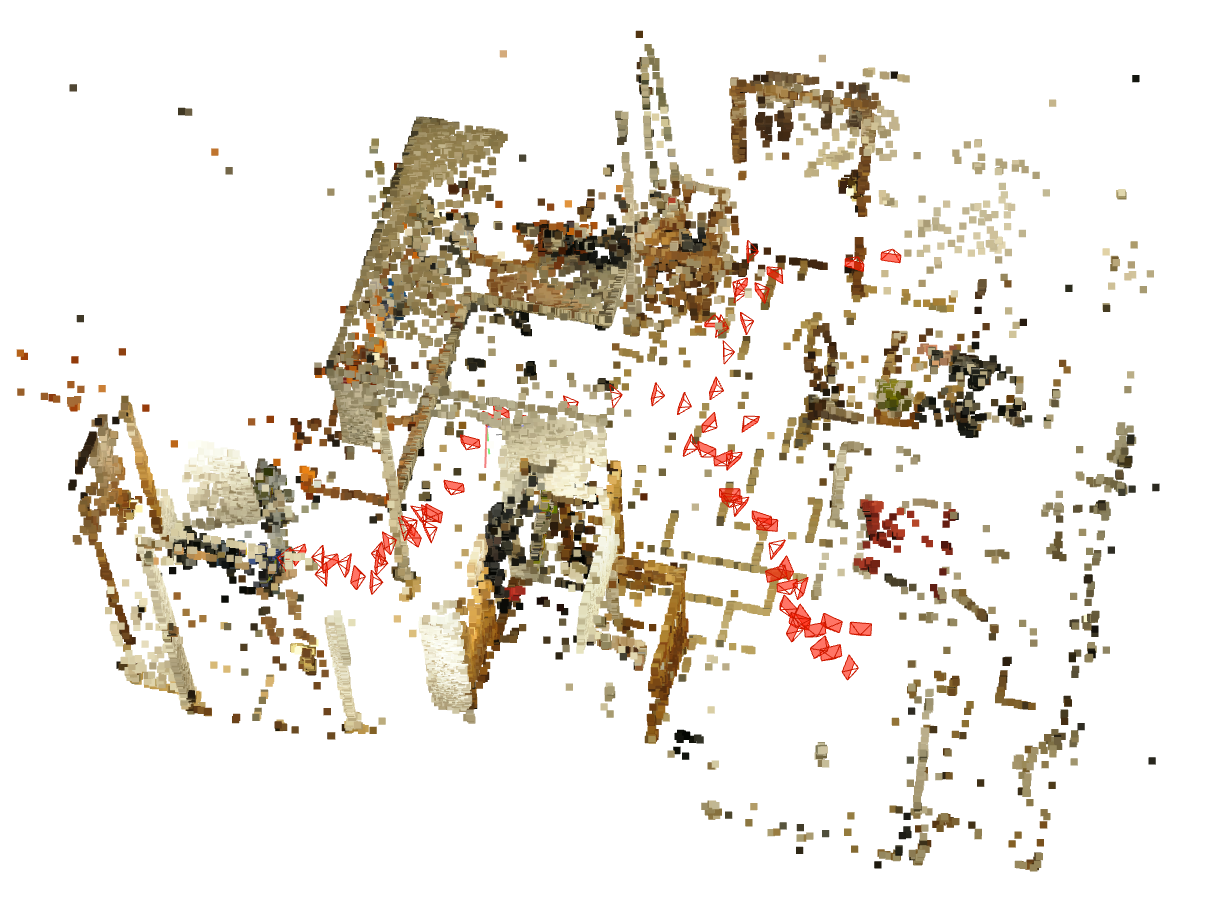}
    \includegraphics[height=0.18\textwidth]{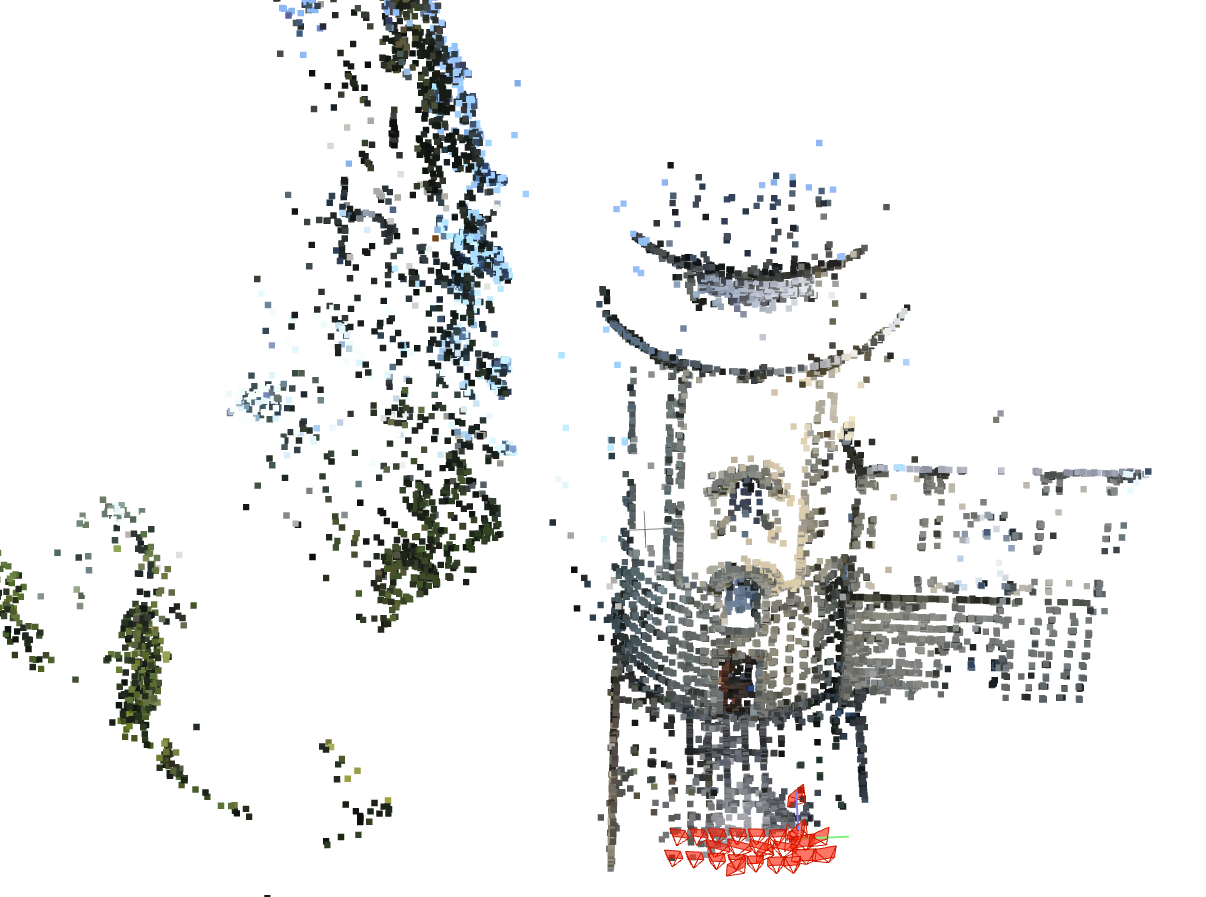}
    \includegraphics[height=0.18\textwidth]{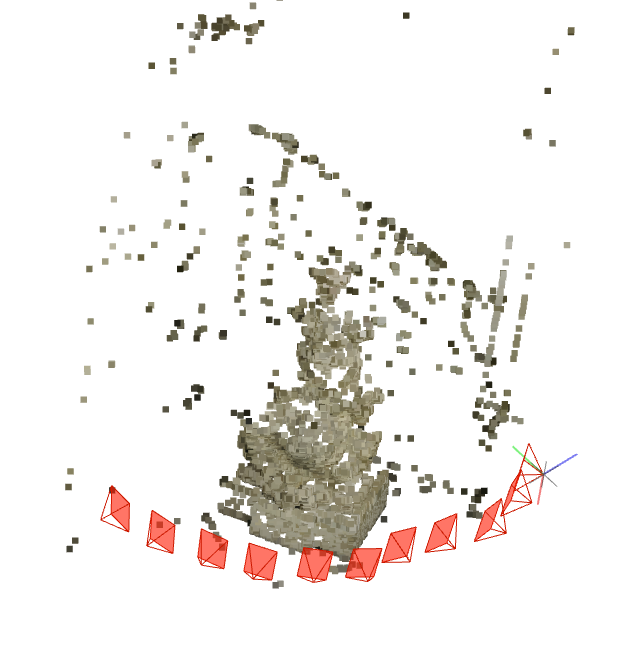}
    \includegraphics[height=0.18\textwidth]{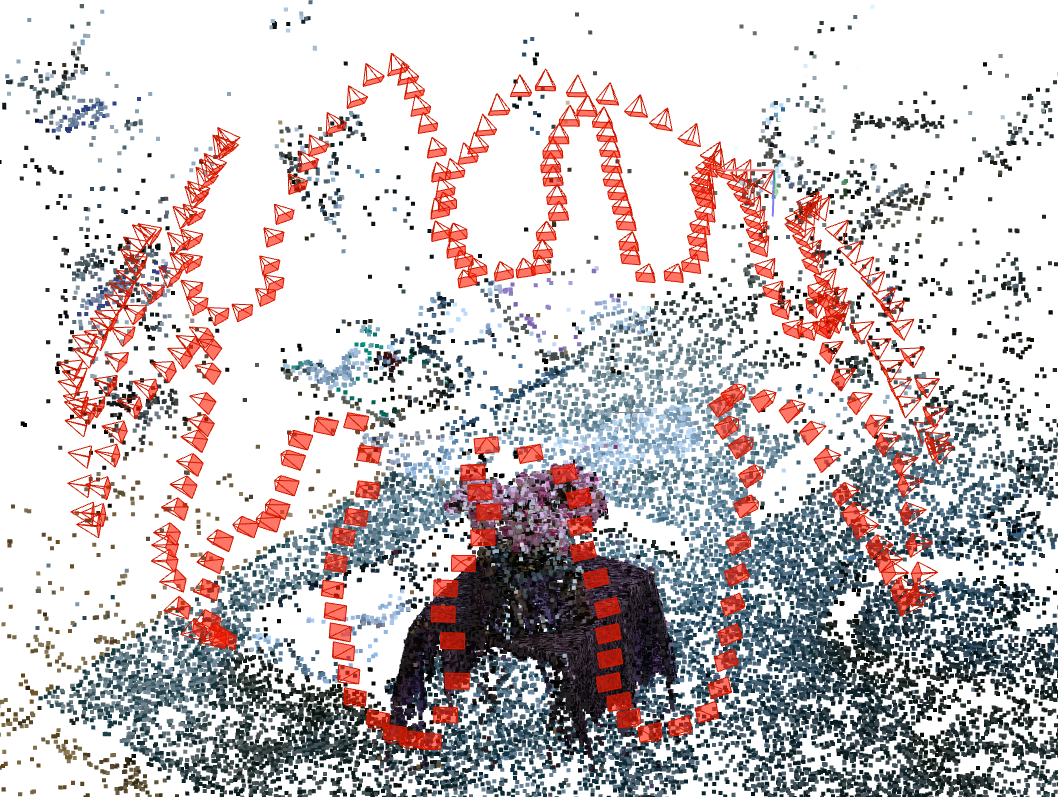}
    \includegraphics[height=0.18\textwidth]{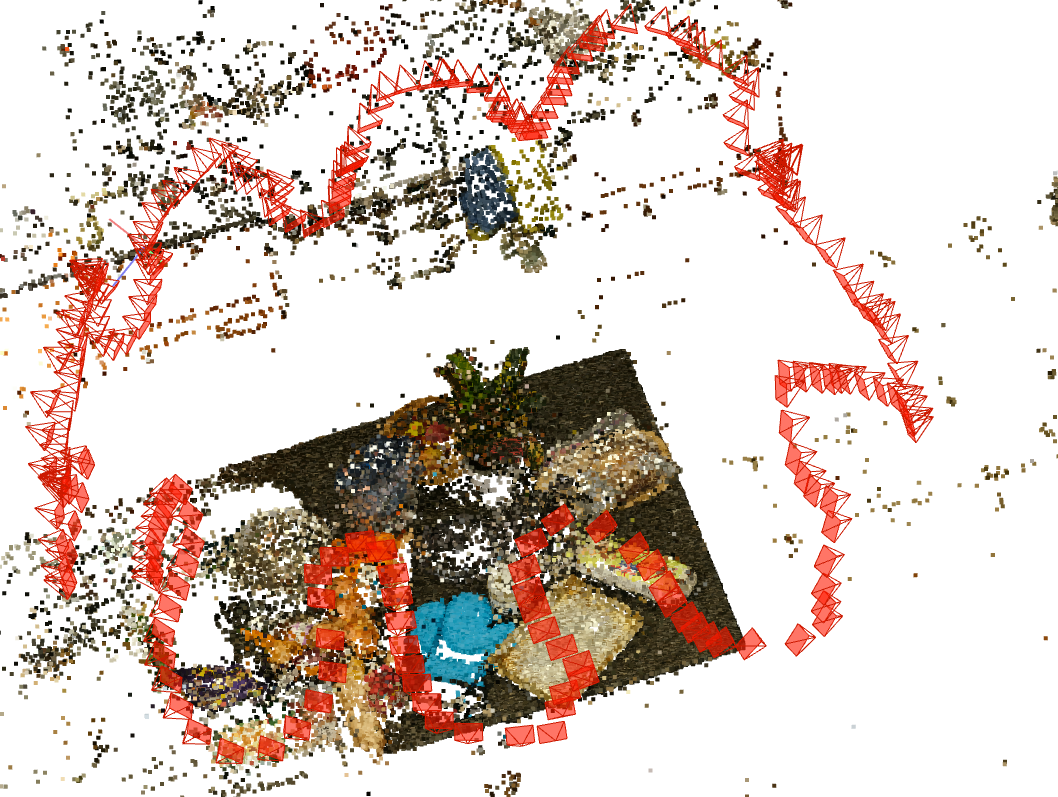}
    \includegraphics[height=0.18\textwidth]{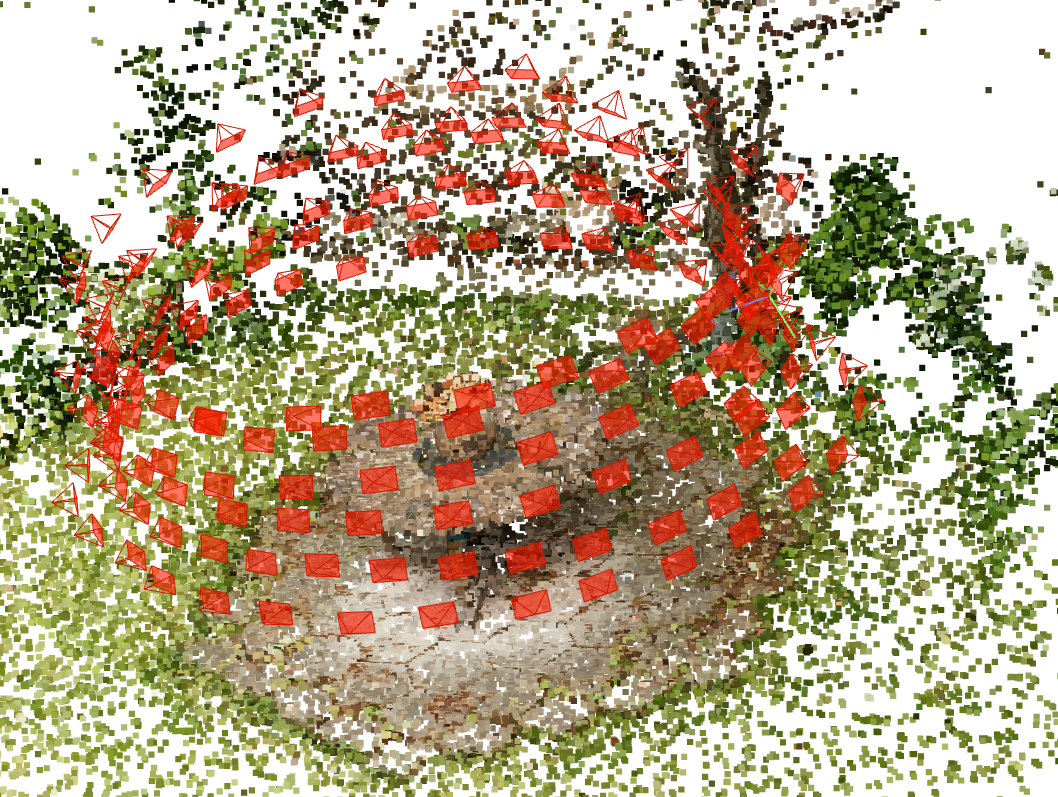}
    \includegraphics[height=0.18\textwidth]{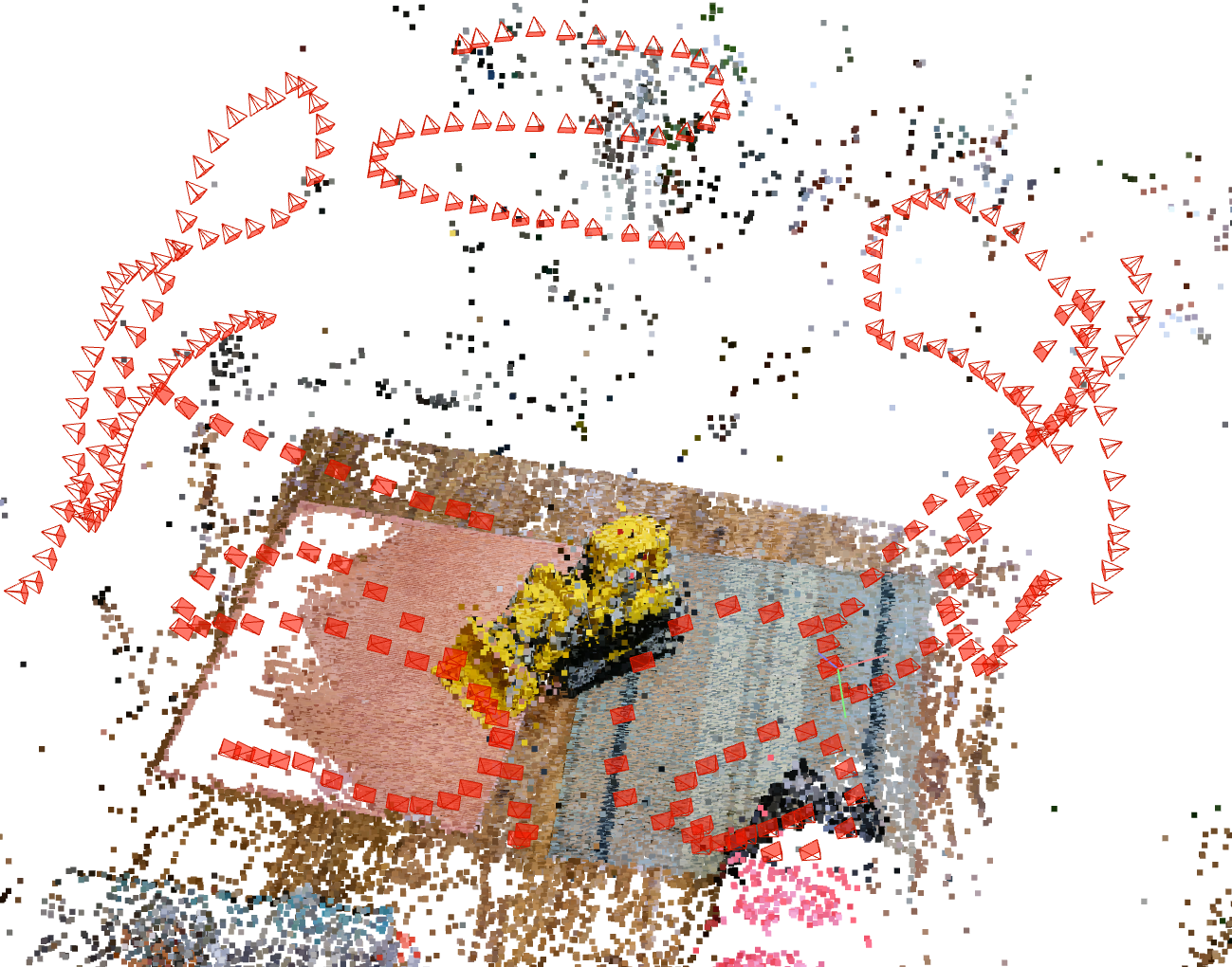}
    \caption{
    Example reconstructions from the proposed GLOMAP on various datasets.
    }
    \label{fig:reconstructions_more}
    % \vspace{-20px}
\end{figure}

More reconstruction results can be found in Fig.~\ref{fig:reconstructions_more}.
From the figure, one can see that the proposed GLOMAP reconstructs the scenes accurately, robustly obtaining the general structure and, also the fine details. 

\begin{table}[h]
    \centering
    \caption{Synthesizing result on MIP360~\cite{barron2022mip} datasets. The proposed method largely outperforms other baselines while obtaining similar results as COLMAP~\cite{schoenberger2016sfm}. 
    For scenes where testing images not all registered (marked \textit{italic}), reference camera pose provided by the dataset is used.
    The differences on \textit{bicycle}, \textit{bonsai}, \textit{garden}, \textit{room} and \textit{stump} are evident. See Fig.~\ref{fig:mip360_instant_ngp} for details.}
    % \vspace{-5px}
    \resizebox{0.6\textwidth}{!}{
    \begin{tabular}{l l c c c a l c c c d
    } \toprule 
        % && \multicolumn{4}{c}{OpenMVG} && \multicolumn{4}{c}{Theia} && \multicolumn{4}{c}{Ours} 
        % && \multicolumn{4}{c}{COLMAP}
        && \multicolumn{4}{c}{PSNR} && \multicolumn{4}{c}{SSIM}
        \\
        \cmidrule{3-6} \cmidrule{8-11}
        && \tiny{OpenMVG} & \tiny{~~Theia~~} & \tiny{GLOMAP} & \tiny{COLMAP} && \tiny{OpenMVG} & \tiny{~~Theia~~} & \tiny{GLOMAP} & \tiny{COLMAP}  \\  \midrule
bicycle && 23.01 & \textit{17.75} & \cellcolor{tabsecond}23.13 & \cellcolor{tabfirst}23.15 && 0.526 & \textit{0.352} & \cellcolor{tabsecond}0.531 & \cellcolor{tabfirst}0.532 \\
bonsai && 23.88 & 28.54 & \cellcolor{tabfirst}30.36 & \cellcolor{tabsecond}29.66 && 0.767 & 0.872 & \cellcolor{tabfirst}0.904 & \cellcolor{tabsecond}0.896 \\
counter && 26.76 & \cellcolor{tabsecond}26.78 & 26.72 & \cellcolor{tabfirst}26.81 && 0.835 & \cellcolor{tabsecond}0.836 & 0.835 & \cellcolor{tabfirst}0.837 \\
garden && \cellcolor{tabsecond}24.97 & \textit{20.19} & \cellcolor{tabsecond}24.97 & \cellcolor{tabfirst}24.98 && \cellcolor{tabsecond}0.653 & \textit{0.456} & \cellcolor{tabfirst}0.655 & \cellcolor{tabsecond}0.653 \\
kitchen && \cellcolor{tabsecond}29.32 & 29.02 & \cellcolor{tabfirst}29.35 & 29.23 && \cellcolor{tabsecond}0.853 & 0.841 & \cellcolor{tabfirst}0.855 & 0.851 \\
room && 19.11 & \textit{17.07} & \cellcolor{tabfirst}29.41 & \cellcolor{tabsecond}29.14 && 0.691 & \textit{0.643} & \cellcolor{tabfirst}0.876 & \cellcolor{tabsecond}0.871 \\
stump && 23.56 & \textit{19.43} & \cellcolor{tabsecond}23.81 & \cellcolor{tabfirst}23.98 && 0.584 & \textit{0.408} & \cellcolor{tabsecond}0.595 & \cellcolor{tabfirst}0.602 \\
\midrule
Average && 24.37 & 22.68 & \cellcolor{tabfirst}26.82 & \cellcolor{tabsecond}26.71 && 0.701 & 0.630 & \cellcolor{tabfirst}0.750 & \cellcolor{tabsecond}0.749 \\
      \bottomrule
    \end{tabular}
    }
    \label{tbl:mip360_instant_ngp}
    % \vspace{-20px}
\end{table}

\begin{figure}[t]
    \centering
    \resizebox{\textwidth}{!}{
    \begin{tabular}{c c c c c 
    } 
    \centering
    \rotatebox{90}{~~~~bicycle} &
    \includegraphics[width=0.23\textwidth]{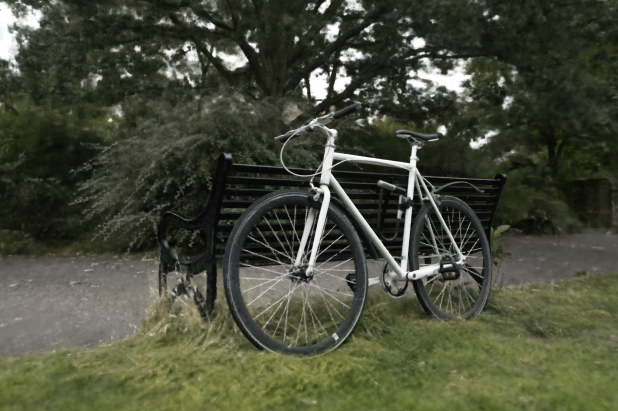} &
    \includegraphics[width=0.23\textwidth]{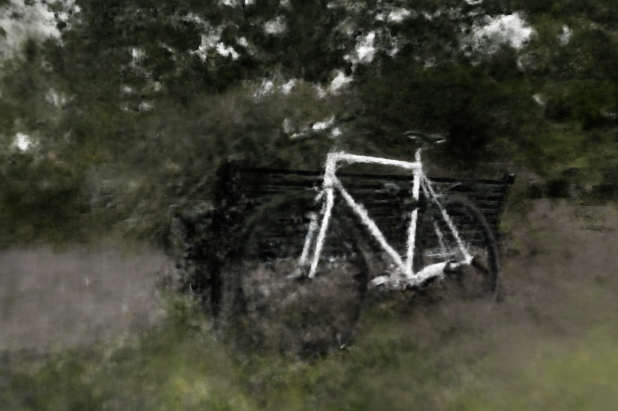} &
    \includegraphics[width=0.23\textwidth]{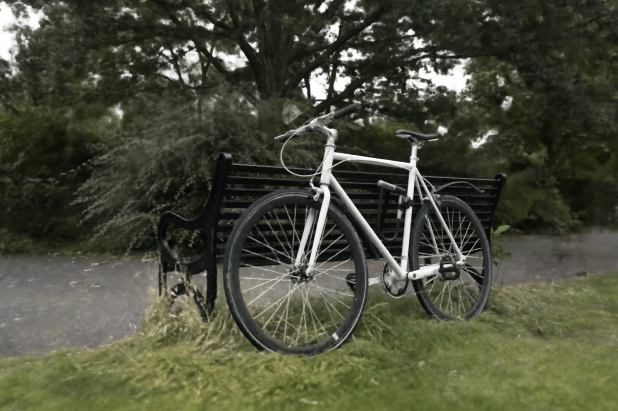} &
    \includegraphics[width=0.23\textwidth]{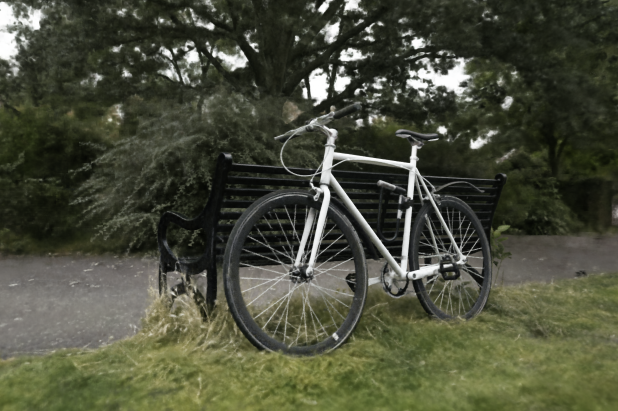} \\
    \rotatebox{90}{~~~~bonsai} &
    \includegraphics[width=0.23\textwidth]{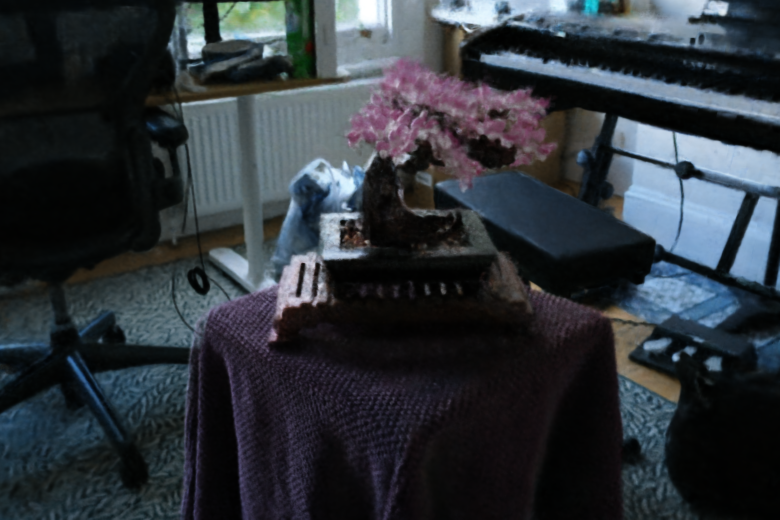} &
    \includegraphics[width=0.23\textwidth]{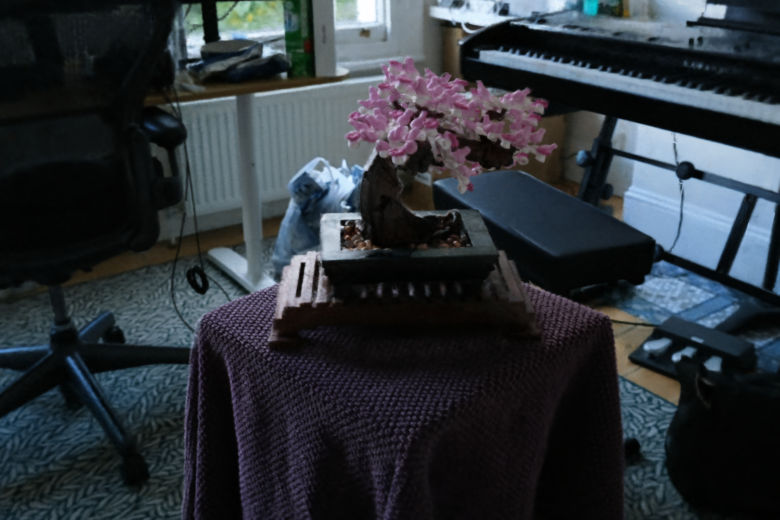} &
    \includegraphics[width=0.23\textwidth]{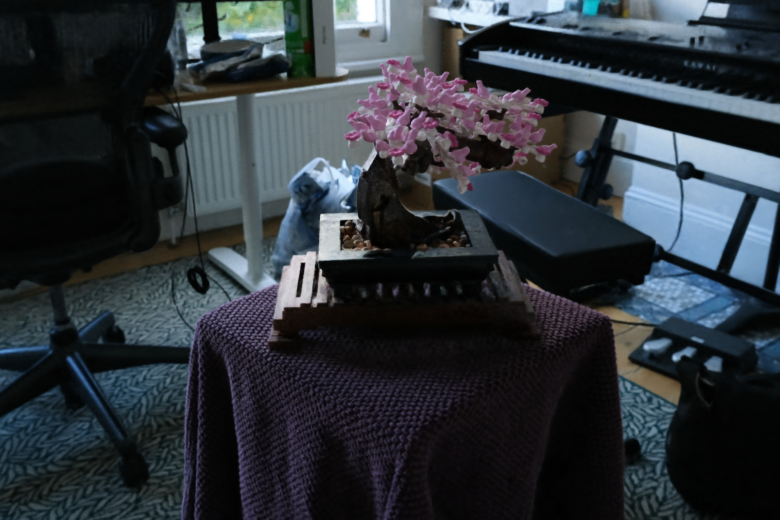} &
    \includegraphics[width=0.23\textwidth]{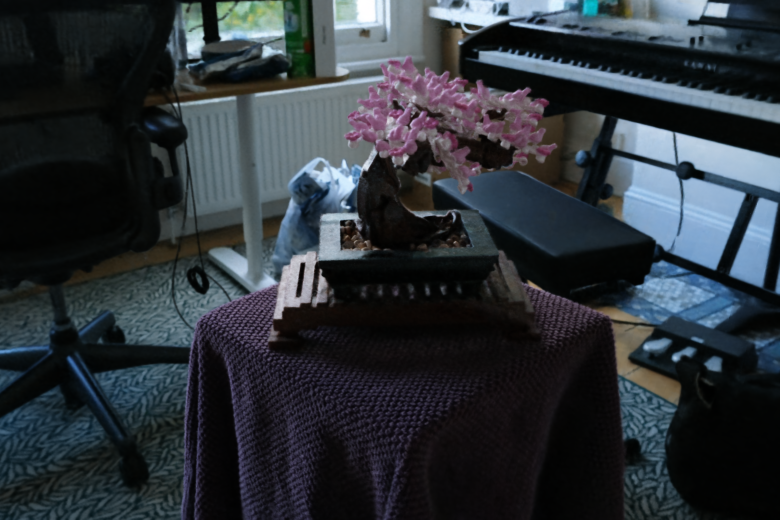} \\
    \rotatebox{90}{~~~~counter} &
    \includegraphics[width=0.23\textwidth]{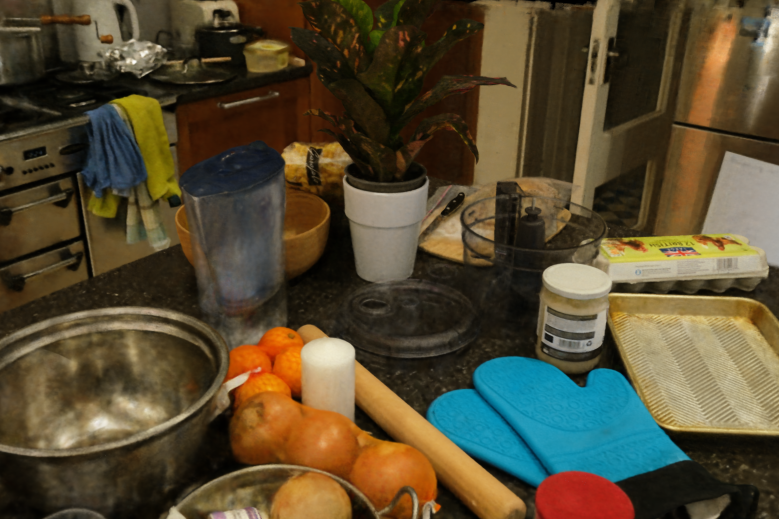} &
    \includegraphics[width=0.23\textwidth]{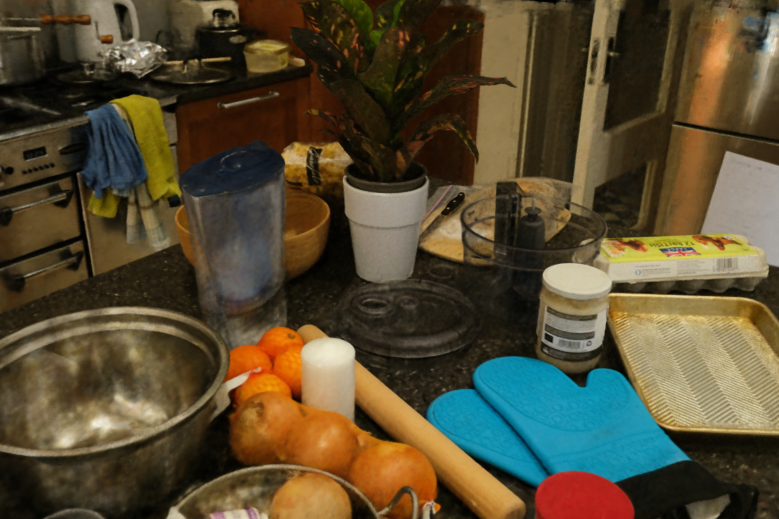} &
    \includegraphics[width=0.23\textwidth]{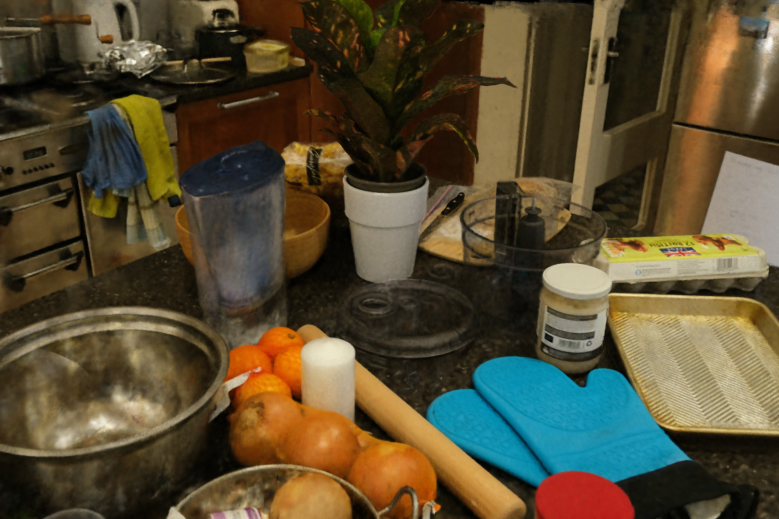} &
    \includegraphics[width=0.23\textwidth]{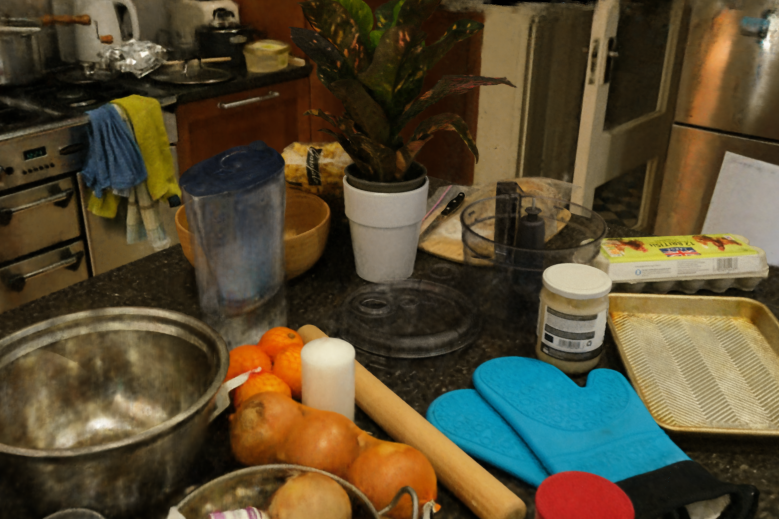} \\
    \rotatebox{90}{~~~~garden} &
    \includegraphics[width=0.23\textwidth]{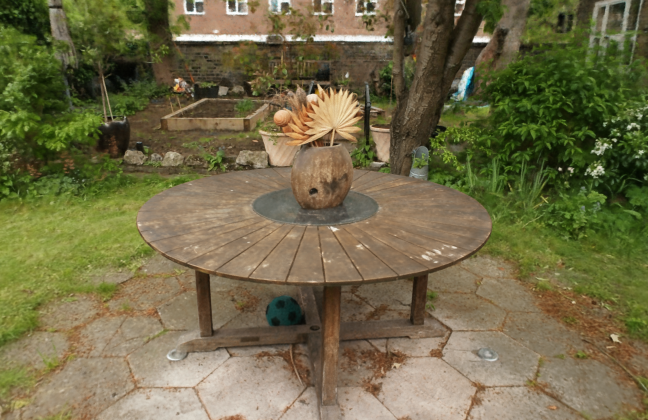} &
    \includegraphics[width=0.23\textwidth]{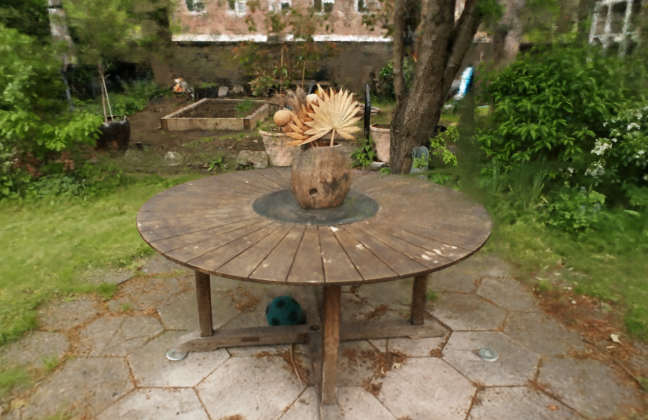} &
    \includegraphics[width=0.23\textwidth]{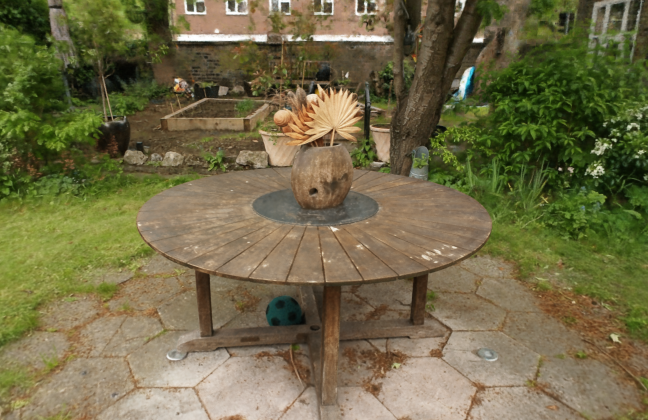} &
    \includegraphics[width=0.23\textwidth]{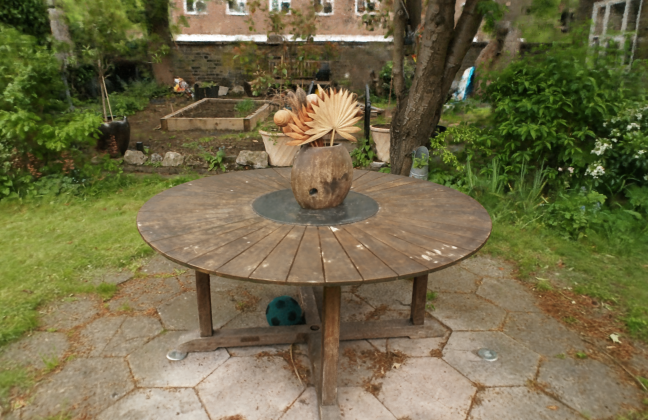} \\
    \rotatebox{90}{~~~~kitchen} &
    \includegraphics[width=0.23\textwidth]{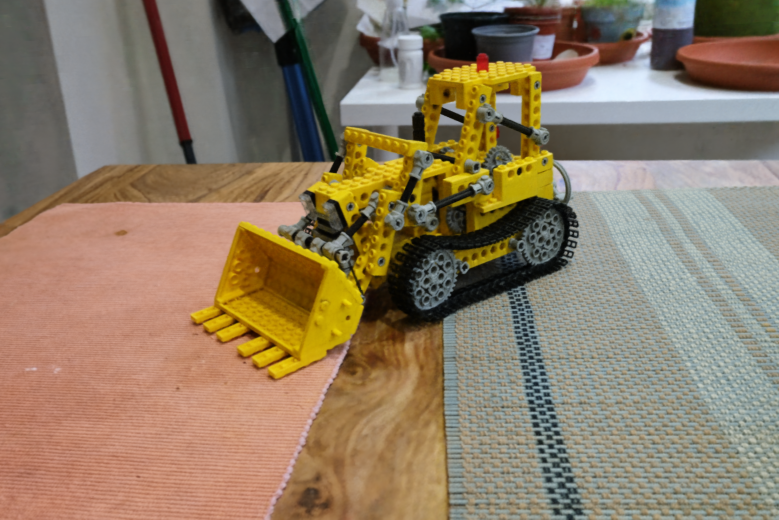} &
    \includegraphics[width=0.23\textwidth]{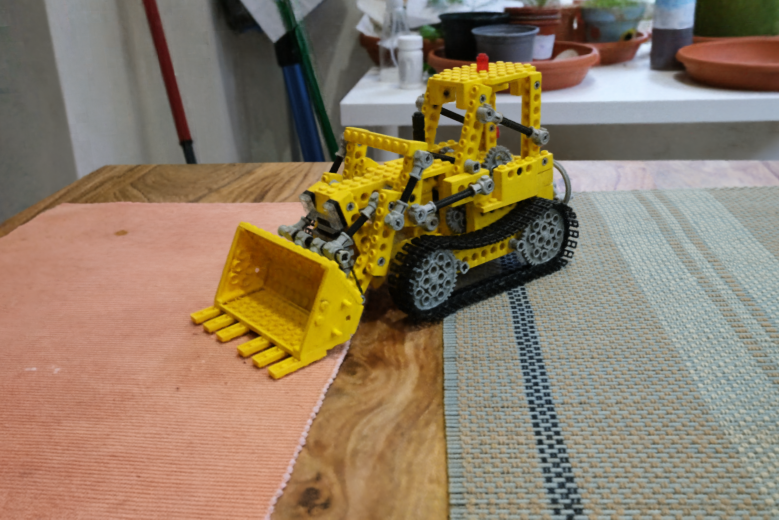} &
    \includegraphics[width=0.23\textwidth]{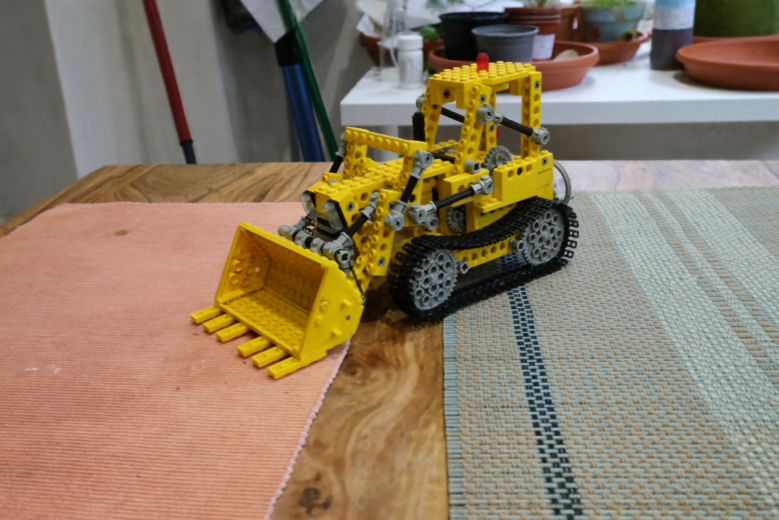} &
    \includegraphics[width=0.23\textwidth]{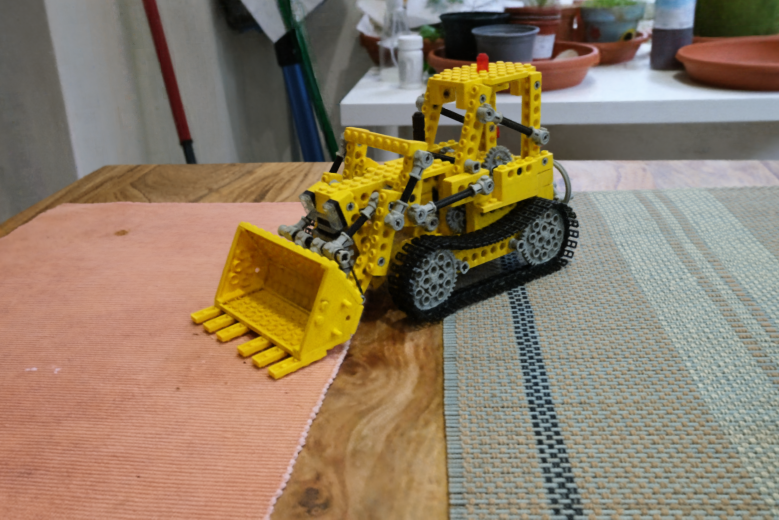} \\
    \rotatebox{90}{~~~~~room} &
    \includegraphics[width=0.23\textwidth]{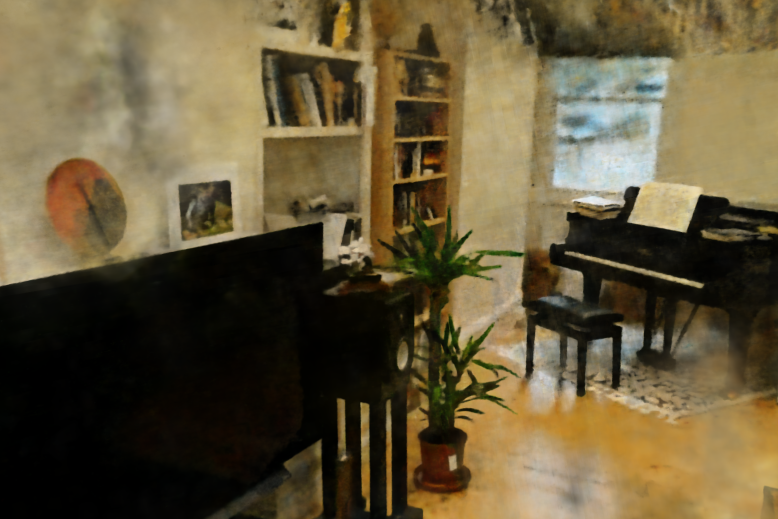} &
    \includegraphics[width=0.23\textwidth]{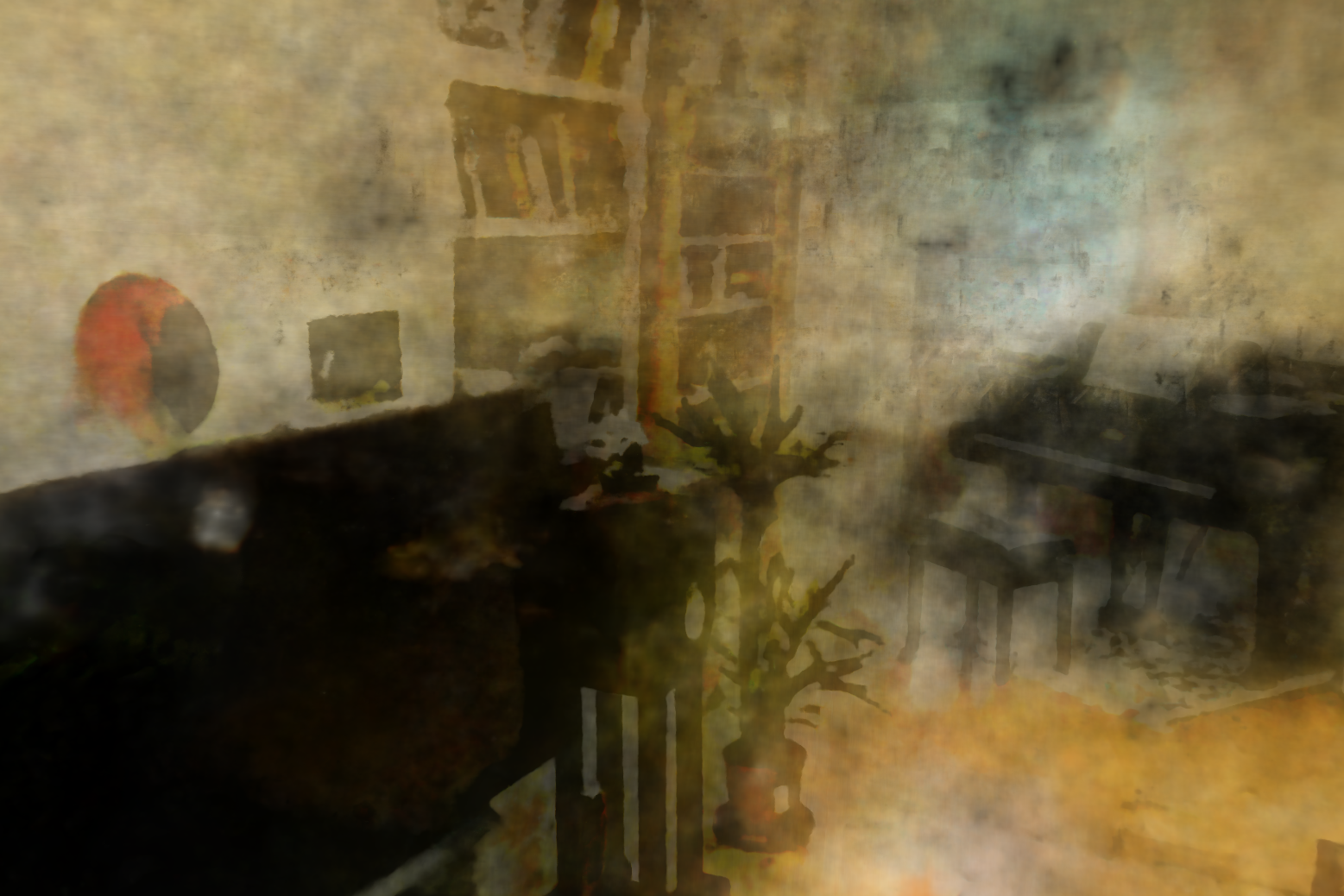} &
    \includegraphics[width=0.23\textwidth]{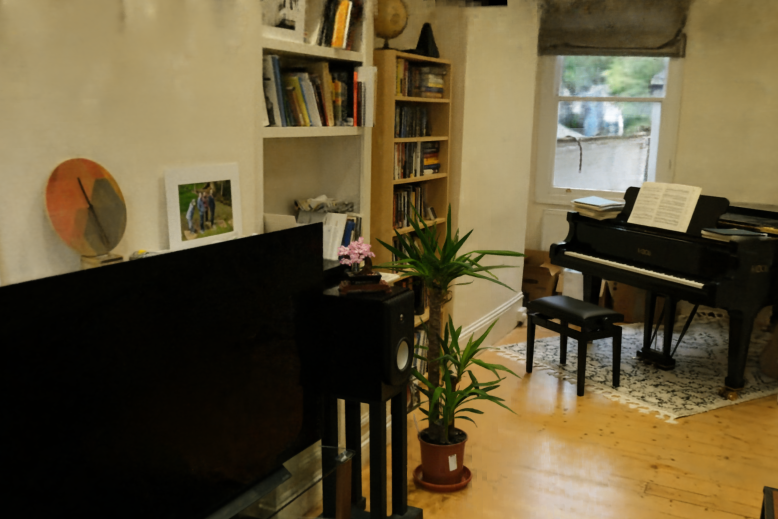} &
    \includegraphics[width=0.23\textwidth]{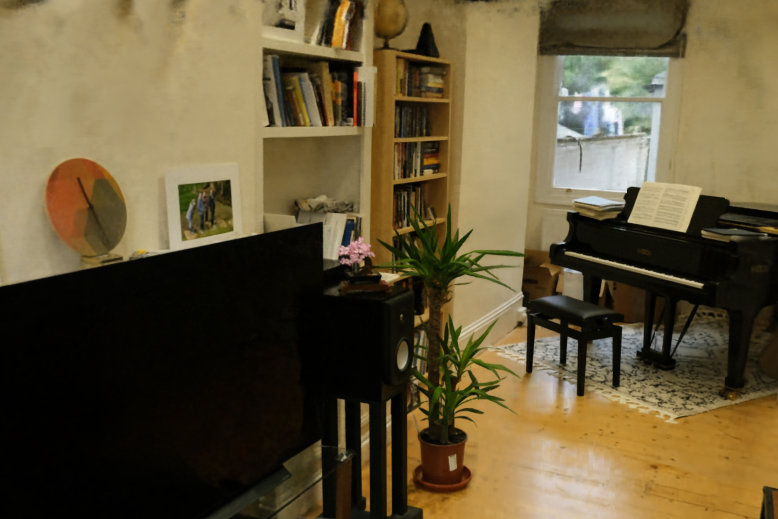} \\
    \rotatebox{90}{~~~~stump} &
    \includegraphics[width=0.23\textwidth]{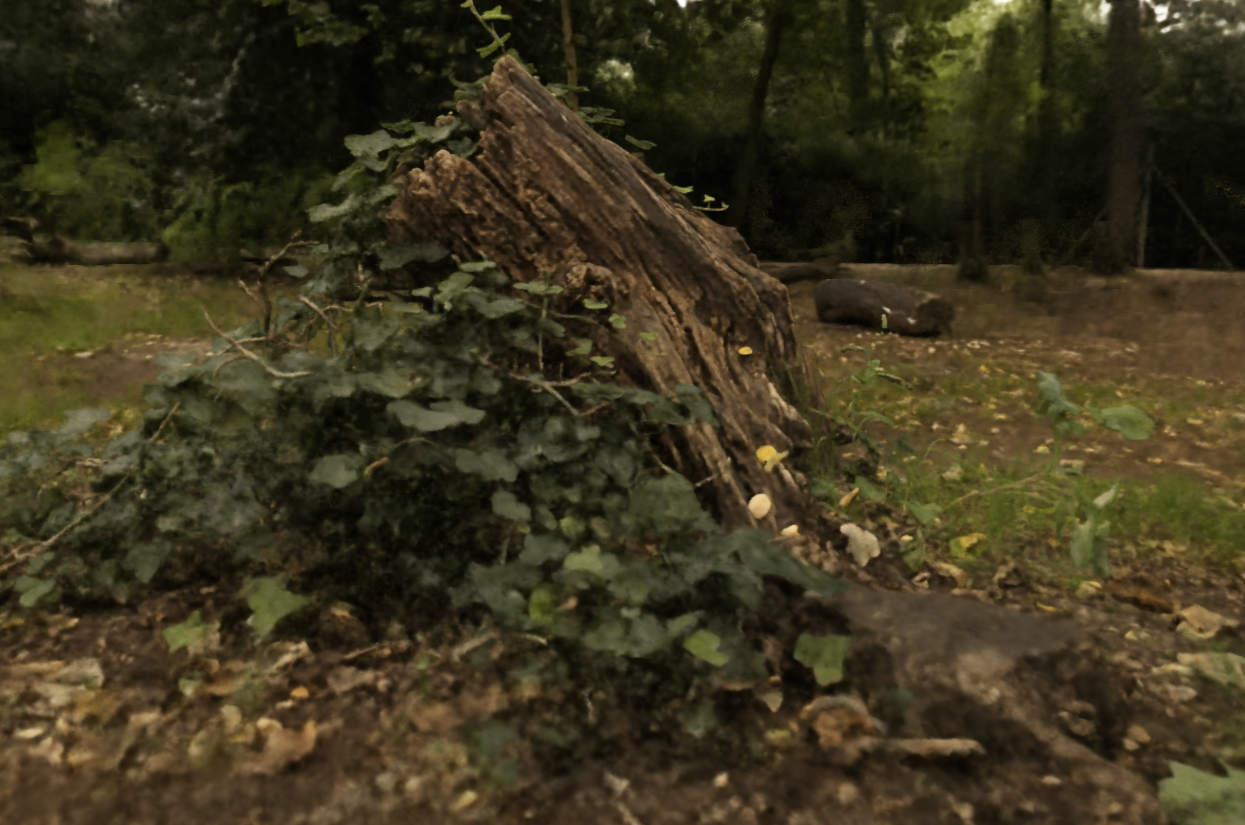} &
    \includegraphics[width=0.23\textwidth]{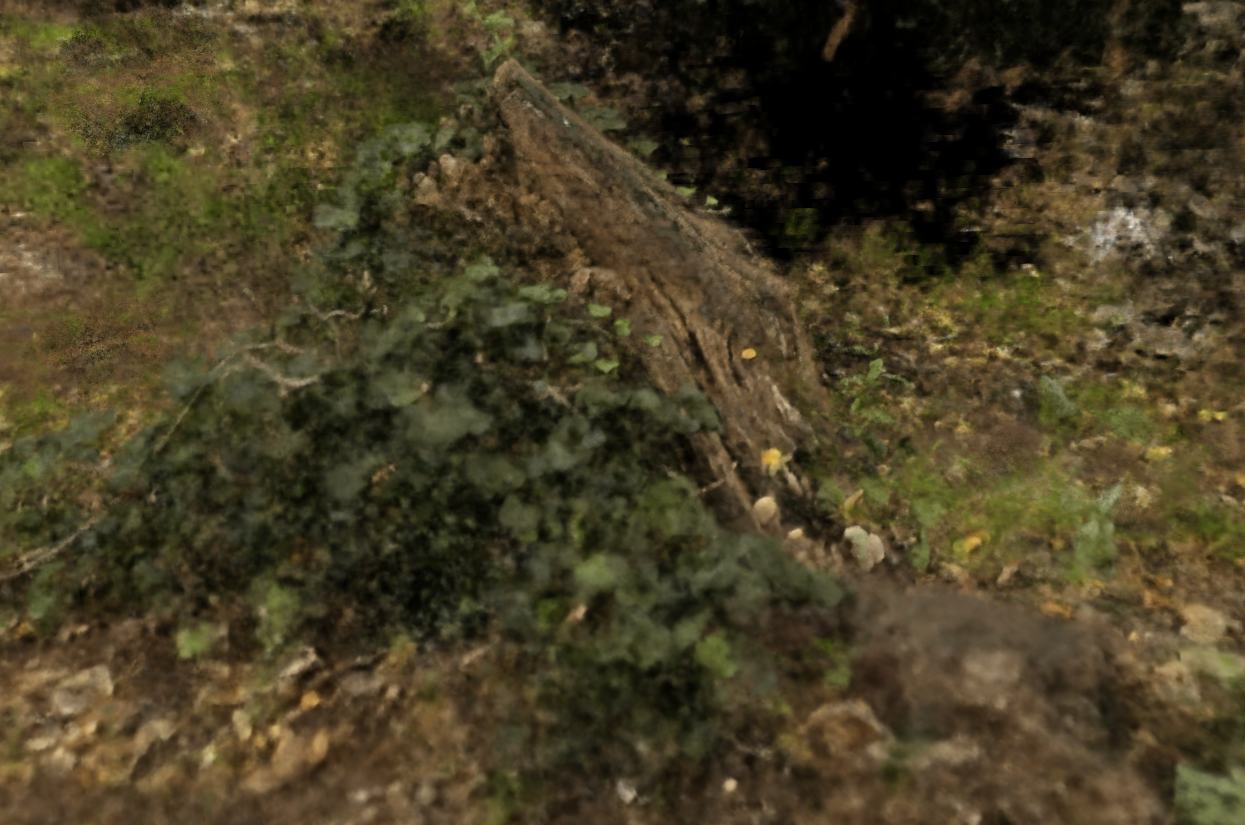} &
    \includegraphics[width=0.23\textwidth]{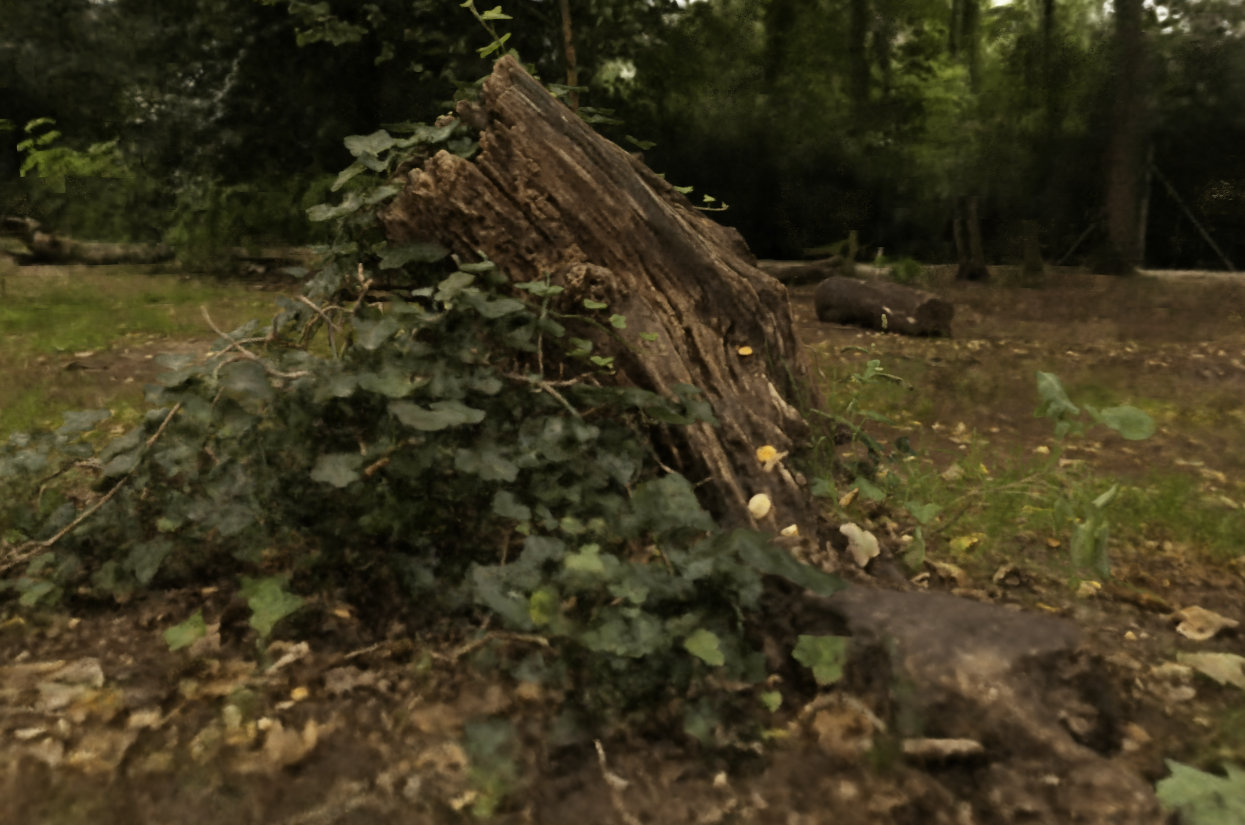} &
    \includegraphics[width=0.23\textwidth]{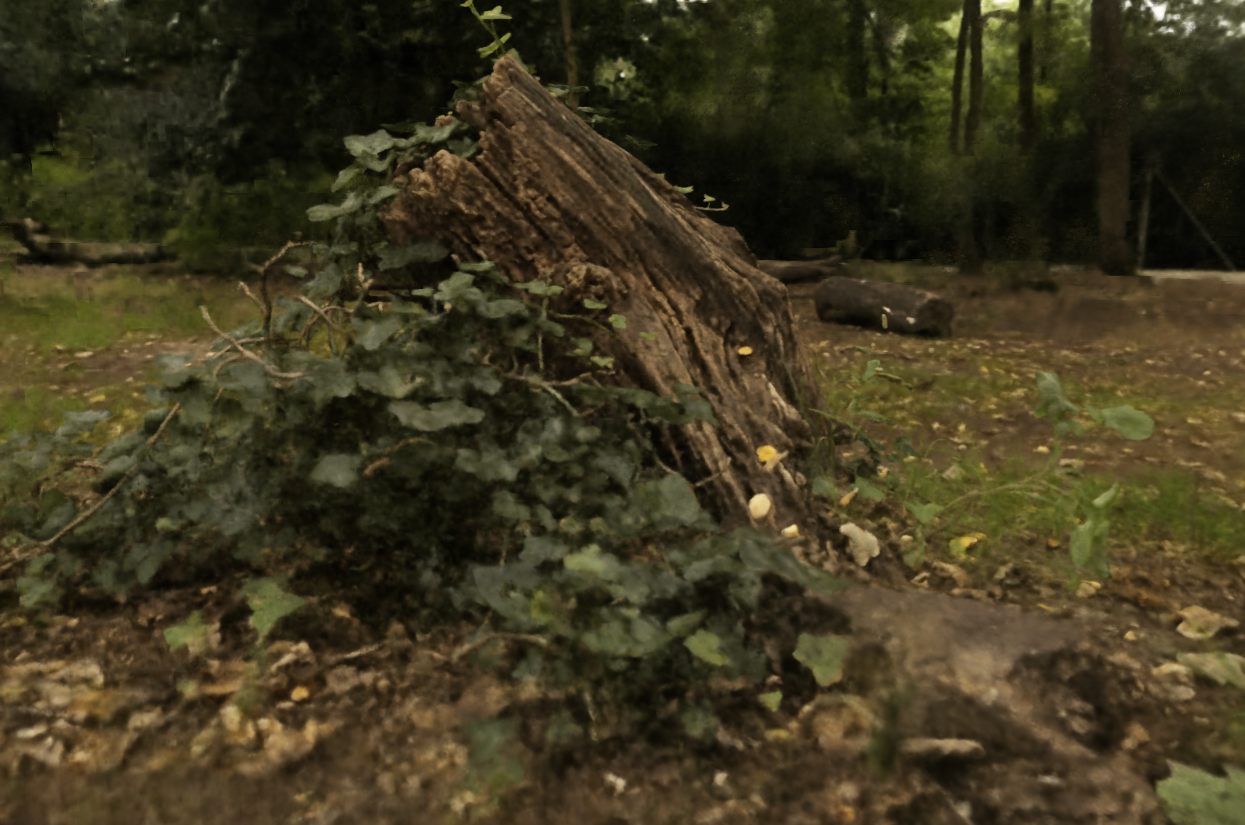} \\
    & (a) OpenMVG~\cite{moulon2016openmvg} & (b) Theia~\cite{theia-manual} & (c) GLOMAP & (d) COLMAP~\cite{schoenberger2016sfm} \\
\end{tabular}
}
    \caption{
    Qualitative results for novel view synthesis with Instant-NGP~\cite{muller2022instant}. The differences on \textit{bicycle}, \textit{bonsai}, \textit{garden}, \textit{room} and \textit{stump} are visually evident.
    }
\label{fig:mip360_instant_ngp}
\end{figure}

\section{Novel View Synthesis}
To examine the impact of reconstruction quality on a downstream task, this section presents results on novel view synthesis.
We conduct experiments with Instant-NGP~\cite{muller2022instant}, a popular method for synthesizing images.
The tested dataset is MIP360~\cite{barron2022mip}, which was originally proposed for this particular application.

Quantitative results can be found in Table~\ref{tbl:mip360_instant_ngp} and qualitative results in Figure~\ref{fig:mip360_instant_ngp}.
We adopt the standard metric, PSNR (peak signal-to-noise ratio) and SSIM (Structural similarity index measure).
From the table, it can be observed that synthesis results with GLOMAP and COLMAP~\cite{schoenberger2016sfm} reconstruction achieve similar scores both in PSNR and SSIM.
For OpenMVG~\cite{moulon2016openmvg} and Theia~\cite{theia-manual}, though they achieve similar scores in some scenes as our reconstruction, they \textit{fail} on several scenes.
Qualitatively, the synthesized results for Theia and OpenMVG are more blurred for several scenes, indicating the poor quality of the camera pose.

\section{Effect of Camera Clustering}
\begin{figure}[t]
    \centering
    \resizebox{\textwidth}{!}{
    \begin{tabular}{c l c c 
    } 
    \centering
    \rotatebox{90}{~~~~Vienna Cathedral} & ~~~ &
    \includegraphics[width=0.43\columnwidth]{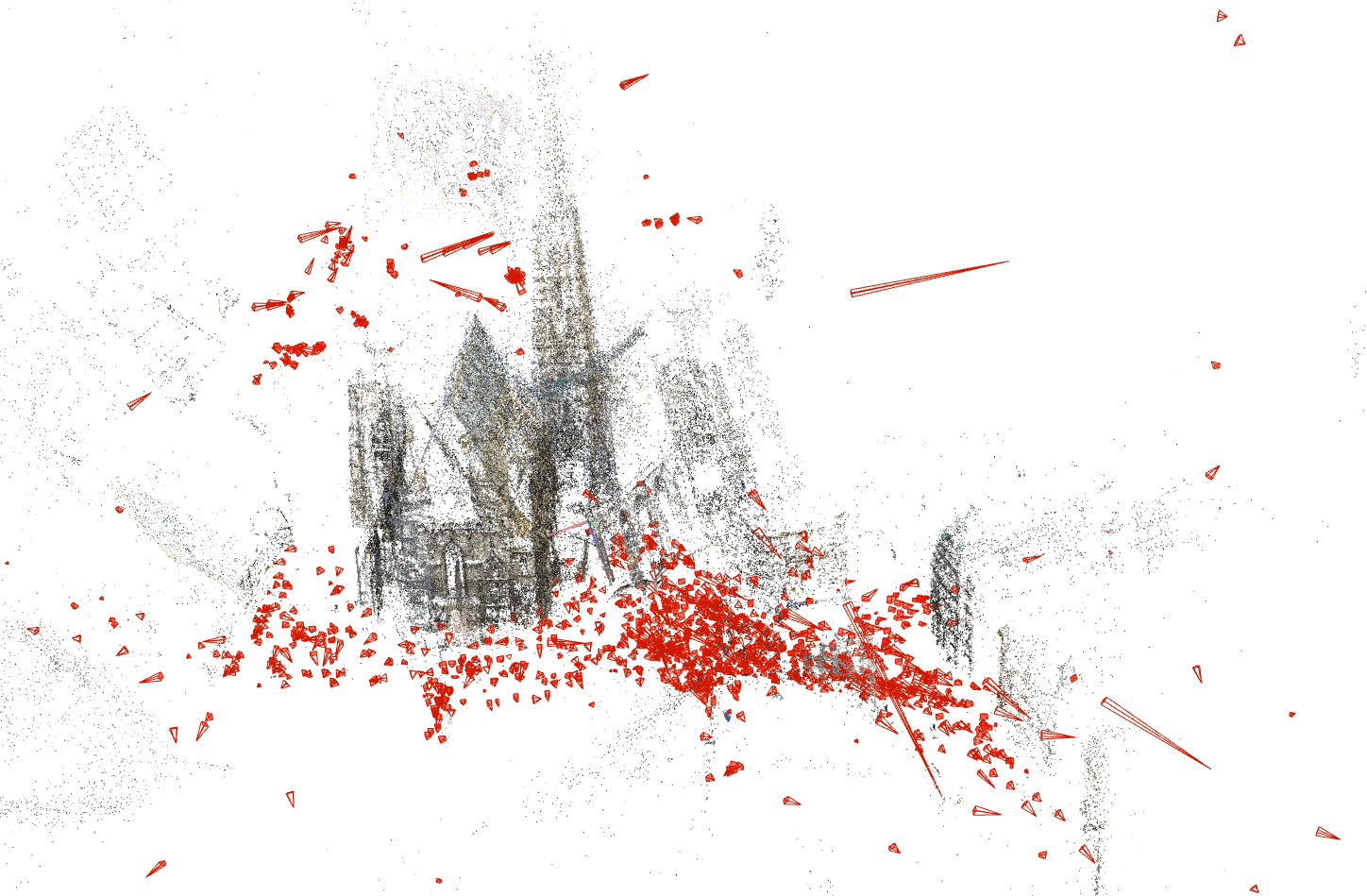} &
    \includegraphics[width=0.43\columnwidth]{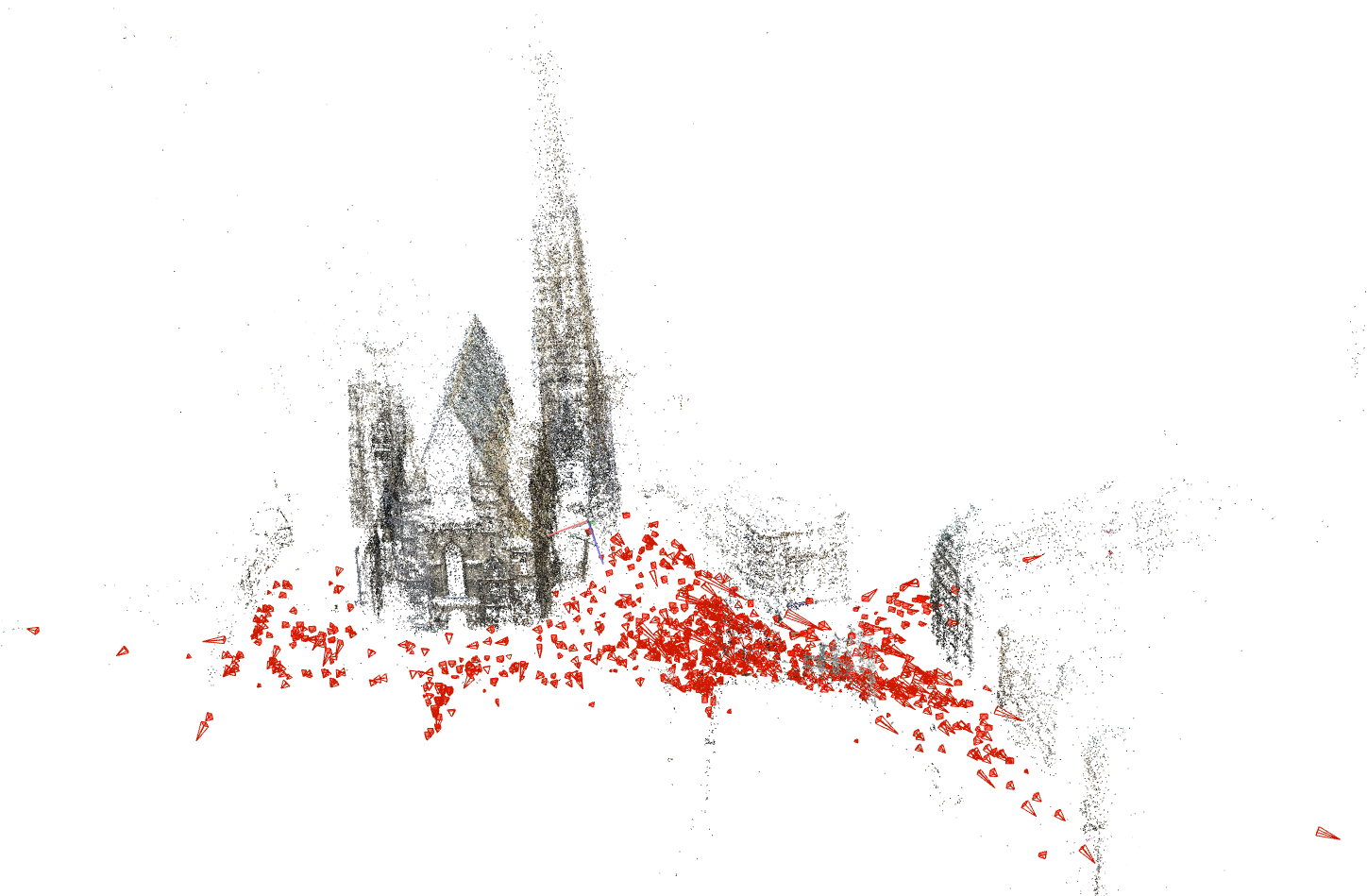} \\
    \rotatebox{90}{~~~~Yorkminster} & ~~~ &
    \includegraphics[width=0.43\columnwidth]{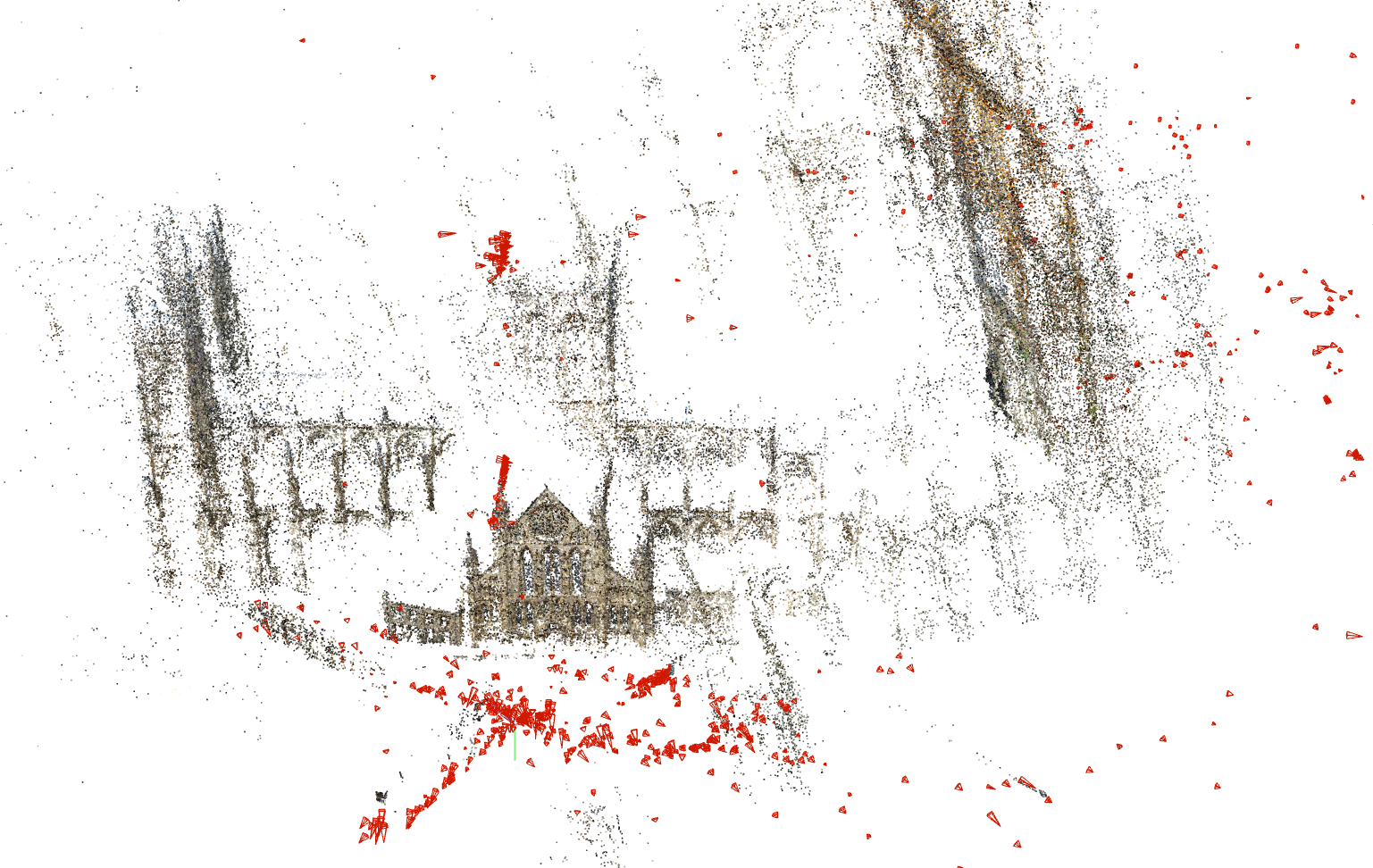} &
    \includegraphics[width=0.43\columnwidth]{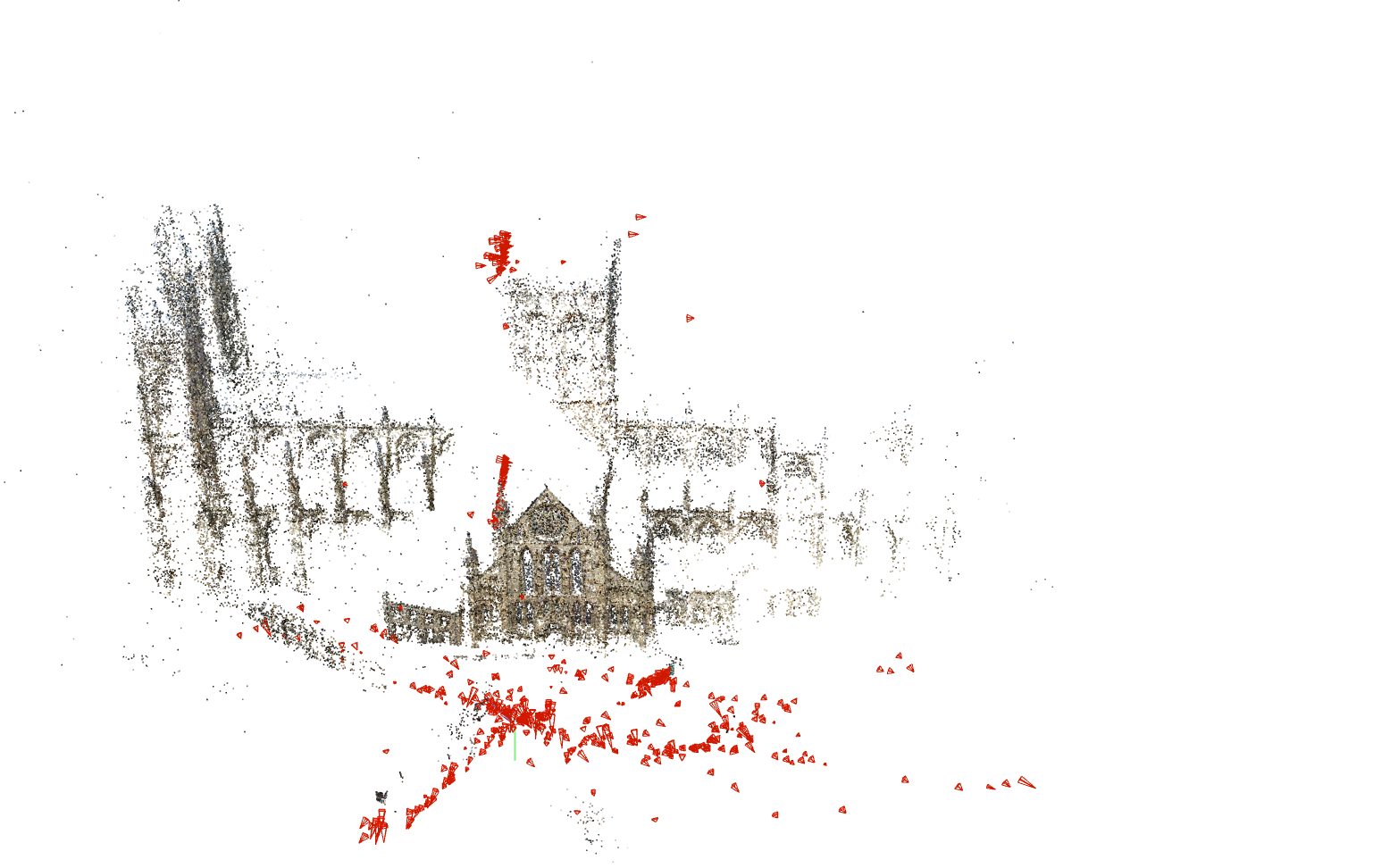} \\
    % & (a) OpenMVG~\cite{moulon2016openmvg} & (b) Theia~\cite{theia-manual} & (c) GLOMAP & (d) COLMAP~\cite{schoenberger2016sfm} \\
\end{tabular}
}
    \caption{
    Qualitative results of camera clustering on 1DSfM~\cite{wilson2014robust} datasets.
    }
\label{fig:camera_clustering}
\end{figure}

For unordered internet image collection, obtaining clean and coherent is not trivial and we propose camera clustering technique for this purpose.
Qualitative results of the mechanism can be found in Figure~\ref{fig:camera_clustering}.
The comparison shows the effectiveness of the proposed mechanism in pruning the floating structures.
% \begin{figure}[h]
%     \centering
%     \subfloat[w/o clustering]{%
%     \includegraphics[width=0.44\columnwidth]{figure/Vienna_Cathedral_all.png}
%     \includegraphics[width=0.44\columnwidth]{figure/Yorkminster_all.png}
%      }
%     \subfloat[w/ clustering]{%
%     \includegraphics[width=0.44\columnwidth]{figure/Vienna_Cathedral.png}
%      }
%     % \vspace{-10px}
%     \caption{
%     Qualitative results of camera clustering on Vienna Cathedral (1DSfM~\cite{wilson2014robust}).
%     % mechanism. This step aims at separating mis-glued clusters.
%     }
%     \label{fig:camera_clustering}
% \end{figure}

\section{Robustness of Global Positioning.}
This experiment validates the robustness in the presence of image noise.
To construct the experiment, we synthesize perfect image observations by projecting COLMAP triangulations to the ground truth cameras, and then, for each observation, we add random Gaussian noise to the reprojections.
We do not run global bundle adjustment in this experiment to isolate the performance of global positioning.
% In theory, a noise level at 0px should make it possible to recover 100\% correct results.
Results can be found in Table~\ref{tbl:ablations}.
% Experimentally, we achieved close to 100 points result with 0px.
For perfect image observations with 0px noise, we reliably converge to the ground truth, underlining the effectiveness of random initialization.
As the noise level increases to extreme values, the AUC scores degrade gradually, indicating a high level of robustness of our proposed global positioning.

\begin{table}[t]
    \centering
    \caption{Ablation on robustness of global positioning to different noise levels of points.}
    % \vspace{-5px}
    % \includegraphics[width=0.5\columnwidth]{fig/ablation.png}
    \resizebox{0.62\textwidth}{!}{
    \begin{tabular}{l l c c c c c c c c
    } \toprule 
    Noise level && ~~0px~~ & ~~1px~~ & ~~2px~~ & ~~4px~~ & ~~8px~~ & ~~16px~~ & ~~32px~~ & ~~64px~~ \\  \midrule
electro && 100.0 & 99.3 & 98.1 & 97.2 & 94.2 & 91.3 & 82.1 & 63.0\\
facade && 98.1 & 97.7 & 97.9 & 98.1 & 97.3 & 95.4 & 94.2 & 90.8\\
kicker && 100.0 & 99.5 & 98.1 & 98.5 & 90.5 & 89.5 & 86.0 & 78.0\\
meadow && 100.0 & 99.3 & 98.0 & 95.8 & 92.6 & 89.2 & 75.3 & 63.8\\
office && 100.0 & 98.6 & 96.6 & 94.2 & 87.3 & 78.5 & 45.5 & 42.5\\
pipes && 100.0 & 98.7 & 97.7 & 97.5 & 94.3 & 88.4 & 82.4 & 56.4\\
playground && 99.7 & 99.5 & 96.4 & 96.6 & 96.0 & 90.4 & 85.8 & 72.2\\
relief && 100.0 & 99.8 & 99.5 & 99.3 & 98.9 & 95.2 & 93.9 & 87.4\\
relief\_2 && 100.0 & 99.8 & 99.5 & 99.1 & 98.4 & 96.7 & 92.5 & 88.3\\
terrace && 100.0 & 99.7 & 99.5 & 99.1 & 98.5 & 94.0 & 91.7 & 80.5\\
terrains && 100.0 & 99.8 & 99.4 & 99.0 & 97.9 & 95.8 & 89.1 & 84.0\\
      \bottomrule
    \end{tabular}
    }
    \label{tbl:ablations}
    % \vspace{-15px}
\end{table}

\section{Detailed Results of ETH3D SLAM}
\begin{table}[]
    \centering
    \caption{Per sequence result for ETH3D SLAM}
    \resizebox{\textwidth}{!}{
    \begin{tabular}{l l c c c a l c c c a l c c c a
    l c c c d
    } \toprule 
        % && \multicolumn{4}{c}{OpenMVG} && \multicolumn{4}{c}{Theia} && \multicolumn{4}{c}{Ours} 
        % && \multicolumn{4}{c}{COLMAP}
        && \multicolumn{4}{c}{Recall @ 0.1} && \multicolumn{4}{c}{AUC @ 0.1} && \multicolumn{4}{c}{AUC @ 0.5} 
        && \multicolumn{4}{c}{Time (s)}
        \\
        \cmidrule{3-6} \cmidrule{8-11} \cmidrule{13-16} \cmidrule{18-21}
        && \tiny{OpenMVG} & \tiny{~~~Theia~~~} & \tiny{GLOMAP} & \tiny{COLMAP} && \tiny{OpenMVG} & \tiny{~~~Theia~~~} & \tiny{GLOMAP} & \tiny{COLMAP} && \tiny{OpenMVG} & \tiny{~~~Theia~~~} & \tiny{GLOMAP} & \tiny{COLMAP} && \tiny{OpenMVG} & \tiny{~~~Theia~~~} & \tiny{GLOMAP} & \tiny{COLMAP} \\  \midrule
cables\_1 && \cellcolor{tabsecond}66.5 & \cellcolor{tabfirst}100.0 & \cellcolor{tabfirst}100.0 & \cellcolor{tabfirst}100.0 && 32.5 & 51.9 & \cellcolor{tabsecond}80.5 & \cellcolor{tabfirst}93.8 && 72.3 & 90.4 & \cellcolor{tabsecond}96.1 & \cellcolor{tabfirst}98.8 && \cellcolor{tabfirst}289.0 & 1001.2 & \cellcolor{tabsecond}549.8 & 7501.8 \\
cables\_2 && \cellcolor{tabfirst}100.0 & \cellcolor{tabfirst}100.0 & \cellcolor{tabfirst}100.0 & \cellcolor{tabsecond}40.2 && \cellcolor{tabfirst}90.9 & 74.6 & \cellcolor{tabsecond}88.0 & 16.7 && \cellcolor{tabfirst}98.2 & 94.9 & \cellcolor{tabsecond}97.6 & 49.4 && \cellcolor{tabsecond}5.2 & \cellcolor{tabfirst}3.8 & 11.6 & 96.8 \\
cables\_3 && \cellcolor{tabfirst}63.9 & \cellcolor{tabfirst}63.9 & \cellcolor{tabfirst}63.9 & \cellcolor{tabfirst}63.9 && 53.9 & 54.1 & \cellcolor{tabfirst}60.1 & \cellcolor{tabsecond}58.9 && 61.9 & 62.0 & \cellcolor{tabfirst}63.2 & \cellcolor{tabsecond}62.9 && \cellcolor{tabsecond}15.0 & \cellcolor{tabfirst}13.5 & 25.5 & 61.5 \\
camera\_shake\_1 && 6.6 & 36.2 & \cellcolor{tabsecond}44.7 & \cellcolor{tabfirst}46.5 && 6.1 & \cellcolor{tabsecond}18.2 & 17.4 & \cellcolor{tabfirst}22.7 && 6.5 & 36.4 & \cellcolor{tabsecond}40.1 & \cellcolor{tabfirst}43.0 && \cellcolor{tabfirst}1.5 & \cellcolor{tabsecond}3.9 & 6.7 & 74.5 \\
camera\_shake\_2 && - & 20.9 & \cellcolor{tabsecond}24.3 & \cellcolor{tabfirst}51.1 && - & 8.6 & \cellcolor{tabsecond}22.1 & \cellcolor{tabfirst}45.0 && - & 19.0 & \cellcolor{tabsecond}24.0 & \cellcolor{tabfirst}49.9 && - & \cellcolor{tabfirst}9.8 & \cellcolor{tabsecond}19.3 & 511.3 \\
camera\_shake\_3 && - & \cellcolor{tabsecond}22.8 & \cellcolor{tabfirst}28.8 & 6.7 && - & \cellcolor{tabsecond}12.0 & \cellcolor{tabfirst}26.2 & 6.4 && - & \cellcolor{tabsecond}23.2 & \cellcolor{tabfirst}28.2 & 6.6 && - & \cellcolor{tabfirst}2.1 & 3.9 & \cellcolor{tabsecond}2.9 \\
ceiling\_1 && - & \cellcolor{tabfirst}9.3 & \cellcolor{tabfirst}9.3 & \cellcolor{tabfirst}9.3 && - & 8.4 & \cellcolor{tabfirst}8.8 & \cellcolor{tabsecond}8.7 && - & \cellcolor{tabsecond}9.1 & \cellcolor{tabfirst}9.2 & \cellcolor{tabfirst}9.2 && - & \cellcolor{tabfirst}17.2 & \cellcolor{tabsecond}40.5 & 116.9 \\
ceiling\_2 && 12.7 & 36.4 & \cellcolor{tabfirst}48.0 & \cellcolor{tabsecond}47.9 && 5.9 & \cellcolor{tabsecond}25.6 & \cellcolor{tabfirst}35.8 & 21.9 && 16.3 & 34.4 & \cellcolor{tabfirst}45.7 & \cellcolor{tabsecond}43.3 && \cellcolor{tabfirst}78.3 & \cellcolor{tabsecond}87.9 & 181.4 & 1998.6 \\
desk\_3 && \cellcolor{tabsecond}37.9 & \cellcolor{tabfirst}46.3 & \cellcolor{tabfirst}46.3 & 28.2 && 16.7 & \cellcolor{tabfirst}42.0 & \cellcolor{tabsecond}39.8 & 27.6 && 39.8 & \cellcolor{tabfirst}45.4 & \cellcolor{tabsecond}45.0 & 28.1 && \cellcolor{tabsecond}126.5 & 255.1 & \cellcolor{tabfirst}124.3 & 767.5 \\
desk\_changing\_1 && 18.1 & 18.2 & \cellcolor{tabsecond}18.4 & \cellcolor{tabfirst}20.6 && 15.2 & \cellcolor{tabsecond}16.4 & \cellcolor{tabfirst}17.2 & 14.7 && 17.5 & 17.8 & \cellcolor{tabsecond}18.1 & \cellcolor{tabfirst}19.4 && 625.9 & \cellcolor{tabfirst}135.4 & \cellcolor{tabsecond}175.6 & 1462.6 \\
einstein\_1 && \cellcolor{tabfirst}90.4 & 85.8 & \cellcolor{tabsecond}87.6 & 45.4 && \cellcolor{tabfirst}46.9 & 35.1 & \cellcolor{tabsecond}44.4 & 18.0 && \cellcolor{tabfirst}88.5 & 85.8 & \cellcolor{tabsecond}88.0 & 56.8 && \cellcolor{tabsecond}122.6 & \cellcolor{tabfirst}120.2 & 139.0 & 1920.6 \\
einstein\_2 && 39.8 & \cellcolor{tabsecond}77.5 & \cellcolor{tabfirst}78.4 & 23.7 && 18.8 & \cellcolor{tabfirst}51.0 & \cellcolor{tabsecond}49.0 & 11.0 && 52.4 & \cellcolor{tabfirst}75.0 & \cellcolor{tabsecond}74.7 & 34.3 && \cellcolor{tabsecond}263.4 & \cellcolor{tabfirst}180.9 & 472.3 & 4765.7 \\
einstein\_dark && - & \cellcolor{tabsecond}5.6 & \cellcolor{tabfirst}7.0 & 5.0 && - & \cellcolor{tabfirst}3.2 & \cellcolor{tabsecond}2.8 & 2.0 && - & \cellcolor{tabsecond}6.3 & \cellcolor{tabfirst}7.4 & 6.1 && - & \cellcolor{tabfirst}20.4 & \cellcolor{tabsecond}31.2 & 267.8 \\
einstein\_flashlight && - & \cellcolor{tabsecond}17.4 & \cellcolor{tabfirst}19.1 & 17.3 && - & \cellcolor{tabsecond}11.2 & \cellcolor{tabfirst}16.3 & 9.4 && - & 16.2 & \cellcolor{tabfirst}18.5 & \cellcolor{tabsecond}17.2 && - & \cellcolor{tabfirst}36.4 & \cellcolor{tabsecond}84.8 & 739.1 \\
einstein\_..change\_1 && - & \cellcolor{tabfirst}17.8 & \cellcolor{tabfirst}17.8 & \cellcolor{tabsecond}11.6 && - & \cellcolor{tabsecond}16.6 & \cellcolor{tabfirst}17.1 & 11.2 && - & \cellcolor{tabsecond}17.6 & \cellcolor{tabfirst}17.7 & 11.5 && - & \cellcolor{tabfirst}8.3 & \cellcolor{tabsecond}19.4 & 21.0 \\
einstein\_..change\_2 && \cellcolor{tabfirst}100.0 & \cellcolor{tabfirst}100.0 & \cellcolor{tabfirst}100.0 & \cellcolor{tabfirst}100.0 && 93.2 & 89.3 & \cellcolor{tabfirst}96.9 & \cellcolor{tabsecond}96.0 && 98.6 & 97.9 & \cellcolor{tabfirst}99.4 & \cellcolor{tabsecond}99.2 && \cellcolor{tabsecond}64.9 & \cellcolor{tabfirst}27.6 & 86.3 & 388.1 \\
einstein\_..change\_3 && - & \cellcolor{tabsecond}30.3 & 30.0 & \cellcolor{tabfirst}32.0 && - & 18.6 & \cellcolor{tabsecond}29.2 & \cellcolor{tabfirst}30.5 && - & 28.0 & \cellcolor{tabsecond}29.9 & \cellcolor{tabfirst}31.7 && - & \cellcolor{tabfirst}100.1 & \cellcolor{tabsecond}161.5 & 513.7 \\
kidnap\_1 && \cellcolor{tabsecond}73.1 & \cellcolor{tabfirst}73.3 & \cellcolor{tabfirst}73.3 & \cellcolor{tabfirst}73.3 && 63.4 & 62.3 & \cellcolor{tabfirst}70.3 & \cellcolor{tabsecond}68.5 && 71.2 & 71.1 & \cellcolor{tabfirst}72.7 & \cellcolor{tabsecond}72.3 && \cellcolor{tabfirst}114.4 & 356.7 & \cellcolor{tabsecond}144.3 & 731.2 \\
large\_loop\_1 && 35.4 & \cellcolor{tabsecond}48.6 & \cellcolor{tabfirst}49.0 & 44.5 && 18.1 & \cellcolor{tabsecond}37.8 & \cellcolor{tabfirst}45.8 & 20.7 && 33.7 & \cellcolor{tabsecond}46.6 & \cellcolor{tabfirst}48.4 & 43.4 && 91.9 & \cellcolor{tabfirst}60.2 & \cellcolor{tabsecond}77.6 & 983.8 \\
mannequin\_1 && 26.8 & \cellcolor{tabsecond}43.1 & \cellcolor{tabfirst}53.2 & 18.8 && 17.5 & 16.3 & \cellcolor{tabfirst}45.8 & \cellcolor{tabsecond}17.7 && 31.2 & \cellcolor{tabsecond}38.0 & \cellcolor{tabfirst}51.7 & 18.6 && \cellcolor{tabsecond}18.6 & 30.0 & \cellcolor{tabfirst}17.7 & 45.4 \\
mannequin\_3 && 23.5 & 34.2 & \cellcolor{tabsecond}35.0 & \cellcolor{tabfirst}41.4 && 21.2 & 13.3 & \cellcolor{tabsecond}32.5 & \cellcolor{tabfirst}38.7 && 23.0 & 30.7 & \cellcolor{tabsecond}34.5 & \cellcolor{tabfirst}40.8 && \cellcolor{tabfirst}10.6 & \cellcolor{tabsecond}13.9 & 16.8 & 72.3 \\
mannequin\_4 && 75.3 & 96.3 & \cellcolor{tabsecond}96.7 & \cellcolor{tabfirst}96.9 && 49.7 & 68.6 & \cellcolor{tabfirst}89.8 & \cellcolor{tabsecond}85.6 && 70.2 & 90.9 & \cellcolor{tabfirst}95.3 & \cellcolor{tabsecond}94.6 && \cellcolor{tabsecond}40.2 & \cellcolor{tabfirst}30.0 & 50.2 & 524.0 \\
mannequin\_5 && 43.4 & 68.9 & \cellcolor{tabsecond}77.3 & \cellcolor{tabfirst}80.2 && 15.9 & 34.1 & \cellcolor{tabfirst}63.9 & \cellcolor{tabsecond}59.8 && 52.5 & 66.4 & \cellcolor{tabsecond}74.6 & \cellcolor{tabfirst}78.0 && \cellcolor{tabsecond}88.0 & \cellcolor{tabfirst}82.4 & 94.2 & 1261.3 \\
mannequin\_7 && 13.9 & 15.2 & \cellcolor{tabfirst}19.6 & \cellcolor{tabsecond}18.3 && 9.0 & 14.6 & \cellcolor{tabfirst}18.0 & \cellcolor{tabsecond}17.8 && 13.8 & 15.1 & \cellcolor{tabfirst}19.3 & \cellcolor{tabsecond}18.2 && \cellcolor{tabsecond}6.8 & \cellcolor{tabfirst}6.2 & 12.5 & 19.7 \\
mannequin\_face\_1 && \cellcolor{tabfirst}100.0 & \cellcolor{tabfirst}100.0 & \cellcolor{tabfirst}100.0 & \cellcolor{tabfirst}100.0 && 92.6 & 97.2 & \cellcolor{tabfirst}98.6 & \cellcolor{tabsecond}98.1 && 98.5 & 99.4 & \cellcolor{tabfirst}99.7 & \cellcolor{tabsecond}99.6 && \cellcolor{tabsecond}40.8 & \cellcolor{tabfirst}17.3 & 49.0 & 248.2 \\
mannequin\_face\_2 && \cellcolor{tabfirst}100.0 & \cellcolor{tabfirst}100.0 & \cellcolor{tabfirst}100.0 & \cellcolor{tabfirst}100.0 && \cellcolor{tabsecond}98.1 & \cellcolor{tabsecond}98.1 & \cellcolor{tabfirst}99.0 & \cellcolor{tabfirst}99.0 && \cellcolor{tabsecond}99.6 & \cellcolor{tabsecond}99.6 & \cellcolor{tabfirst}99.8 & \cellcolor{tabfirst}99.8 && \cellcolor{tabsecond}53.2 & \cellcolor{tabfirst}17.6 & 76.3 & 244.4 \\
mannequin\_face\_3 && 17.4 & 45.2 & \cellcolor{tabfirst}69.5 & \cellcolor{tabsecond}64.3 && 14.9 & 32.5 & \cellcolor{tabfirst}51.9 & \cellcolor{tabsecond}45.1 && 16.9 & 42.7 & \cellcolor{tabfirst}66.1 & \cellcolor{tabsecond}62.3 && \cellcolor{tabfirst}16.8 & \cellcolor{tabsecond}21.7 & 52.4 & 285.6 \\
mannequin\_head && 41.9 & \cellcolor{tabfirst}56.8 & \cellcolor{tabsecond}55.7 & 14.0 && 11.2 & \cellcolor{tabsecond}43.5 & \cellcolor{tabfirst}52.6 & 13.1 && 36.1 & \cellcolor{tabsecond}54.1 & \cellcolor{tabfirst}56.8 & 13.8 && \cellcolor{tabsecond}26.8 & 47.8 & 29.9 & \cellcolor{tabfirst}10.2 \\
motion\_1 && \cellcolor{tabsecond}18.8 & 16.9 & \cellcolor{tabfirst}39.8 & 17.7 && 11.0 & 11.9 & \cellcolor{tabfirst}22.5 & \cellcolor{tabsecond}12.9 && \cellcolor{tabsecond}23.3 & 19.2 & \cellcolor{tabfirst}45.9 & 19.7 && 859.7 & \cellcolor{tabfirst}109.0 & \cellcolor{tabsecond}788.9 & 9995.1 \\
planar\_2 && \cellcolor{tabsecond}24.0 & \cellcolor{tabfirst}100.0 & \cellcolor{tabfirst}100.0 & \cellcolor{tabfirst}100.0 && 9.2 & \cellcolor{tabsecond}99.0 & \cellcolor{tabfirst}99.1 & \cellcolor{tabfirst}99.1 && \cellcolor{tabsecond}31.8 & \cellcolor{tabfirst}99.8 & \cellcolor{tabfirst}99.8 & \cellcolor{tabfirst}99.8 && \cellcolor{tabsecond}330.5 & \cellcolor{tabfirst}149.9 & 540.3 & 1220.9 \\
planar\_3 && \cellcolor{tabsecond}37.2 & \cellcolor{tabfirst}100.0 & \cellcolor{tabfirst}100.0 & \cellcolor{tabfirst}100.0 && 15.7 & 96.7 & \cellcolor{tabfirst}98.3 & \cellcolor{tabsecond}97.5 && 44.2 & 99.3 & \cellcolor{tabfirst}99.7 & \cellcolor{tabsecond}99.5 && \cellcolor{tabsecond}297.1 & \cellcolor{tabfirst}185.1 & 526.3 & 3478.5 \\
plant\_1 && \cellcolor{tabfirst}100.0 & \cellcolor{tabfirst}100.0 & \cellcolor{tabfirst}100.0 & \cellcolor{tabfirst}100.0 && 90.1 & 97.8 & \cellcolor{tabfirst}98.6 & \cellcolor{tabsecond}98.5 && 98.0 & \cellcolor{tabsecond}99.6 & \cellcolor{tabfirst}99.7 & \cellcolor{tabfirst}99.7 && \cellcolor{tabsecond}3.9 & \cellcolor{tabfirst}1.9 & \cellcolor{tabsecond}3.9 & 10.7 \\
plant\_2 && \cellcolor{tabfirst}100.0 & \cellcolor{tabfirst}100.0 & \cellcolor{tabfirst}100.0 & \cellcolor{tabfirst}100.0 && 98.1 & 98.5 & \cellcolor{tabfirst}98.8 & \cellcolor{tabsecond}98.6 && 99.6 & \cellcolor{tabsecond}99.7 & \cellcolor{tabfirst}99.8 & \cellcolor{tabsecond}99.7 && \cellcolor{tabfirst}7.4 & \cellcolor{tabsecond}7.6 & 20.4 & 45.3 \\
plant\_3 && \cellcolor{tabfirst}100.0 & \cellcolor{tabfirst}100.0 & \cellcolor{tabfirst}100.0 & \cellcolor{tabfirst}100.0 && 54.4 & \cellcolor{tabfirst}96.2 & 93.2 & \cellcolor{tabsecond}93.8 && 90.9 & \cellcolor{tabfirst}99.2 & 98.6 & \cellcolor{tabsecond}98.8 && 15.0 & \cellcolor{tabfirst}7.7 & \cellcolor{tabsecond}14.6 & 45.9 \\
plant\_4 && \cellcolor{tabfirst}100.0 & \cellcolor{tabfirst}100.0 & \cellcolor{tabfirst}100.0 & \cellcolor{tabfirst}100.0 && 97.9 & \cellcolor{tabsecond}98.8 & 98.7 & \cellcolor{tabfirst}98.9 && 99.6 & \cellcolor{tabfirst}99.8 & \cellcolor{tabsecond}99.7 & \cellcolor{tabfirst}99.8 && \cellcolor{tabsecond}3.8 & \cellcolor{tabfirst}3.6 & 16.0 & 19.1 \\
plant\_5 && \cellcolor{tabfirst}100.0 & \cellcolor{tabfirst}100.0 & \cellcolor{tabfirst}100.0 & \cellcolor{tabfirst}100.0 && 95.7 & 96.3 & \cellcolor{tabfirst}98.3 & \cellcolor{tabsecond}97.0 && 99.1 & 99.3 & \cellcolor{tabfirst}99.7 & \cellcolor{tabsecond}99.4 && \cellcolor{tabsecond}7.3 & \cellcolor{tabfirst}6.9 & 18.5 & 36.3 \\
plant\_scene\_1 && 43.9 & \cellcolor{tabsecond}77.8 & \cellcolor{tabsecond}77.8 & \cellcolor{tabfirst}98.5 && 36.1 & 52.2 & \cellcolor{tabsecond}71.8 & \cellcolor{tabfirst}85.0 && 42.3 & 72.7 & \cellcolor{tabsecond}76.6 & \cellcolor{tabfirst}95.8 && \cellcolor{tabsecond}35.4 & \cellcolor{tabfirst}25.0 & 39.6 & 441.4 \\
plant\_scene\_2 && 41.6 & \cellcolor{tabsecond}84.4 & \cellcolor{tabfirst}99.0 & 76.1 && 15.8 & 27.8 & \cellcolor{tabfirst}58.3 & \cellcolor{tabsecond}35.5 && 59.2 & \cellcolor{tabsecond}83.4 & \cellcolor{tabfirst}91.3 & 82.3 && 80.4 & \cellcolor{tabfirst}54.2 & \cellcolor{tabsecond}64.8 & 644.6 \\
plant\_scene\_3 && 30.4 & 50.3 & \cellcolor{tabfirst}69.6 & \cellcolor{tabsecond}67.8 && 15.1 & 34.6 & \cellcolor{tabsecond}38.6 & \cellcolor{tabfirst}50.8 && 27.7 & 52.3 & \cellcolor{tabfirst}81.9 & \cellcolor{tabsecond}64.4 && \cellcolor{tabfirst}20.2 & 179.1 & \cellcolor{tabsecond}51.5 & 378.5 \\
reflective\_1 && 12.6 & 16.1 & \cellcolor{tabsecond}22.0 & \cellcolor{tabfirst}26.2 && 6.7 & 9.0 & \cellcolor{tabfirst}12.1 & \cellcolor{tabsecond}9.2 && 16.5 & 23.0 & \cellcolor{tabsecond}31.3 & \cellcolor{tabfirst}33.5 && 721.3 & \cellcolor{tabfirst}118.3 & \cellcolor{tabsecond}434.4 & 6573.9 \\
repetitive && 26.3 & \cellcolor{tabsecond}28.5 & \cellcolor{tabfirst}32.7 & \cellcolor{tabsecond}28.5 && 23.9 & 15.2 & \cellcolor{tabfirst}29.2 & \cellcolor{tabsecond}27.2 && 25.8 & 27.0 & \cellcolor{tabfirst}32.0 & \cellcolor{tabsecond}28.3 && \cellcolor{tabfirst}63.2 & 136.8 & \cellcolor{tabsecond}74.5 & 561.1 \\
sfm\_bench && 72.0 & 96.6 & \cellcolor{tabsecond}98.3 & \cellcolor{tabfirst}100.0 && 61.0 & 88.6 & \cellcolor{tabsecond}92.7 & \cellcolor{tabfirst}94.1 && 69.8 & 95.0 & \cellcolor{tabsecond}97.2 & \cellcolor{tabfirst}98.8 && \cellcolor{tabsecond}73.9 & \cellcolor{tabfirst}41.3 & 103.6 & 461.4 \\
sfm\_garden && 76.4 & 80.0 & \cellcolor{tabfirst}87.8 & \cellcolor{tabsecond}84.7 && \cellcolor{tabsecond}32.5 & 31.1 & \cellcolor{tabfirst}57.2 & 29.4 && 76.6 & 79.7 & \cellcolor{tabfirst}90.3 & \cellcolor{tabsecond}82.3 && \cellcolor{tabfirst}272.8 & \cellcolor{tabsecond}698.3 & 835.2 & 798.1 \\
sfm\_house\_loop && \cellcolor{tabsecond}70.5 & \cellcolor{tabfirst}100.0 & \cellcolor{tabfirst}100.0 & 42.4 && 53.0 & \cellcolor{tabsecond}76.6 & \cellcolor{tabfirst}86.3 & 31.2 && 67.0 & \cellcolor{tabsecond}95.3 & \cellcolor{tabfirst}97.3 & 40.8 && \cellcolor{tabfirst}95.6 & \cellcolor{tabsecond}96.5 & 222.8 & 1030.1 \\
sfm\_lab\_room\_1 && \cellcolor{tabfirst}99.6 & \cellcolor{tabfirst}99.6 & \cellcolor{tabfirst}99.6 & \cellcolor{tabsecond}20.9 && \cellcolor{tabsecond}75.7 & 44.0 & \cellcolor{tabfirst}77.9 & 11.7 && \cellcolor{tabsecond}94.8 & 88.5 & \cellcolor{tabfirst}95.3 & 23.5 && \cellcolor{tabfirst}12.8 & \cellcolor{tabsecond}14.3 & 32.1 & 37.9 \\
sfm\_lab\_room\_2 && \cellcolor{tabsecond}97.6 & 94.4 & \cellcolor{tabfirst}99.2 & 31.2 && \cellcolor{tabsecond}67.2 & 28.2 & \cellcolor{tabfirst}84.0 & 12.3 && \cellcolor{tabsecond}91.5 & 84.9 & \cellcolor{tabfirst}96.2 & 33.5 && \cellcolor{tabsecond}3.3 & \cellcolor{tabfirst}2.3 & 5.0 & 20.7 \\
sofa\_1 && 10.6 & 21.3 & \cellcolor{tabsecond}25.9 & \cellcolor{tabfirst}27.0 && 5.7 & 11.7 & \cellcolor{tabsecond}22.9 & \cellcolor{tabfirst}24.1 && 9.6 & 22.1 & \cellcolor{tabsecond}25.3 & \cellcolor{tabfirst}26.4 && \cellcolor{tabfirst}13.3 & 21.1 & \cellcolor{tabsecond}14.1 & 252.8 \\
sofa\_2 && 18.6 & \cellcolor{tabsecond}21.8 & \cellcolor{tabsecond}21.8 & \cellcolor{tabfirst}43.1 && 8.3 & 11.3 & \cellcolor{tabsecond}20.4 & \cellcolor{tabfirst}38.7 && 16.6 & 19.7 & \cellcolor{tabsecond}21.5 & \cellcolor{tabfirst}42.4 && \cellcolor{tabsecond}9.2 & \cellcolor{tabfirst}6.1 & 9.8 & 137.2 \\
sofa\_3 && 15.5 & 19.5 & \cellcolor{tabsecond}22.5 & \cellcolor{tabfirst}28.9 && 8.1 & 15.4 & \cellcolor{tabsecond}21.1 & \cellcolor{tabfirst}25.0 && 14.9 & 20.4 & \cellcolor{tabsecond}22.2 & \cellcolor{tabfirst}28.1 && \cellcolor{tabsecond}4.9 & \cellcolor{tabfirst}2.9 & 5.1 & 50.4 \\
sofa\_4 && - & \cellcolor{tabsecond}25.3 & \cellcolor{tabsecond}25.3 & \cellcolor{tabfirst}29.9 && - & 14.0 & \cellcolor{tabsecond}23.8 & \cellcolor{tabfirst}26.8 && - & 23.2 & \cellcolor{tabsecond}25.1 & \cellcolor{tabfirst}29.3 && - & \cellcolor{tabfirst}5.5 & \cellcolor{tabsecond}11.5 & 188.8 \\
table\_3 && \cellcolor{tabfirst}100.0 & \cellcolor{tabfirst}100.0 & \cellcolor{tabfirst}100.0 & \cellcolor{tabfirst}100.0 && 85.9 & 90.3 & \cellcolor{tabfirst}97.7 & \cellcolor{tabsecond}96.8 && 97.2 & 98.1 & \cellcolor{tabfirst}99.5 & \cellcolor{tabsecond}99.4 && 299.0 & \cellcolor{tabfirst}83.0 & \cellcolor{tabsecond}281.2 & 2116.8 \\
table\_4 && \cellcolor{tabfirst}100.0 & \cellcolor{tabfirst}100.0 & \cellcolor{tabfirst}100.0 & \cellcolor{tabfirst}100.0 && 89.1 & 88.2 & \cellcolor{tabsecond}95.9 & \cellcolor{tabfirst}96.0 && \cellcolor{tabsecond}97.8 & 97.6 & \cellcolor{tabfirst}99.2 & \cellcolor{tabfirst}99.2 && \cellcolor{tabsecond}142.0 & \cellcolor{tabfirst}68.8 & 182.8 & 2817.7 \\
table\_7 && 37.2 & 81.0 & \cellcolor{tabsecond}83.0 & \cellcolor{tabfirst}99.7 && 29.6 & 29.0 & \cellcolor{tabsecond}59.2 & \cellcolor{tabfirst}94.1 && 35.7 & 72.0 & \cellcolor{tabsecond}78.2 & \cellcolor{tabfirst}98.6 && \cellcolor{tabfirst}104.9 & \cellcolor{tabsecond}141.7 & 200.5 & 3398.7 \\
vicon\_light\_1 && 47.2 & 68.5 & \cellcolor{tabsecond}94.1 & \cellcolor{tabfirst}97.8 && 14.4 & 25.6 & \cellcolor{tabfirst}65.8 & \cellcolor{tabsecond}55.4 && 60.8 & 77.5 & \cellcolor{tabsecond}88.4 & \cellcolor{tabfirst}89.9 && \cellcolor{tabsecond}67.9 & 128.7 & \cellcolor{tabfirst}53.3 & 345.1 \\
vicon\_light\_2 && \cellcolor{tabsecond}82.0 & \cellcolor{tabfirst}100.0 & \cellcolor{tabfirst}100.0 & 64.3 && 26.4 & \cellcolor{tabsecond}88.6 & \cellcolor{tabfirst}95.3 & 22.3 && 82.7 & \cellcolor{tabsecond}97.7 & \cellcolor{tabfirst}99.1 & 60.1 && \cellcolor{tabfirst}33.8 & 47.8 & \cellcolor{tabsecond}39.0 & 604.6 \\
\midrule
\textit{Average} && 48.2 & \cellcolor{tabsecond}62.8 & \cellcolor{tabfirst}66.4 & 57.9 && 34.9 & 46.0 & \cellcolor{tabfirst}57.0 & \cellcolor{tabsecond}47.6 && 48.6 & \cellcolor{tabsecond}61.1 & \cellcolor{tabfirst}65.7 & 57.9 && \cellcolor{tabsecond}120.8 & \cellcolor{tabfirst}91.8 & 133.5 & 1115.4 \\

      \bottomrule
    \end{tabular}
    }
    \label{tbl:eth3d_slam_full}
    % \vspace{-3px}
\end{table}

Per-sequence result for ETH3D SLAM can be found in Table~\ref{tbl:eth3d_slam_full}.
From the table, one can see that the relative performance is consistent with the averaged results across sequences sharing the same prefix.
\end{document}